\documentclass[10pt]{article} 
\usepackage[preprint]{tmlr}

\usepackage{amsmath,amsfonts,bm,amssymb,dsfont}

\usepackage{amsthm}

\def\eqref#1{equation~\ref{#1}}

\def\1{\bm{1}}

\def\rmL{{\mathbf{L}}}

\def\rmS{{\mathbf{S}}}
\def\rmT{{\mathbf{T}}}
\def\rmU{{\mathbf{U}}}
\def\rmV{{\mathbf{V}}}
\def\rmW{{\mathbf{W}}}
\def\rmX{{\mathbf{X}}}
\def\rmY{{\mathbf{Y}}}
\def\rmZ{{\mathbf{Z}}}

\def\mS{{\bm{S}}}

\def\mU{{\bm{U}}}
\def\mV{{\bm{V}}}
\def\mW{{\bm{W}}}
\def\mX{{\bm{X}}}
\def\mY{{\bm{Y}}}
\def\mZ{{\bm{Z}}}

\DeclareMathAlphabet{\mathsfit}{\encodingdefault}{\sfdefault}{m}{sl}
\SetMathAlphabet{\mathsfit}{bold}{\encodingdefault}{\sfdefault}{bx}{n}

\def\gA{{\mathcal{A}}}

\def\gT{{\mathcal{T}}}

\def\gW{{\mathcal{W}}}

\def\gY{{\mathcal{Y}}}
\def\gZ{{\mathcal{Z}}}

\def\sR{{\mathbb{R}}}

\newcommand{\E}{\mathbb{E}}

\newtheorem{theorem}{Theorem} 
\newtheorem{corollary}{Corollary} 
\newtheorem{lemma}{Lemma} 
\newtheorem{definition}{Definition} 

\newtheorem{assumption}{Assumption}
\newtheorem{remark}{Remark}

\usepackage{hyperref}
\usepackage{url}
\usepackage{booktabs} 

\usepackage{color,soul}

\usepackage{hyperref}
\usepackage{url}
\usepackage{graphicx}
\usepackage{enumitem}

\title{Class-wise Generalization Error: \\an Information-Theoretic Analysis}

\author{\name Firas Laakom \email firas.laakom@tuni.fi \\
      \addr Department of Computing Sciences\\
      Tampere University, Finland
      \AND
      \name Yuheng Bu\email buyuheng@ufl.edu \\
      \addr Department of Electrical \& Computer Engineering \\
      University of Florida, USA
      \AND
      \name Moncef Gabbouj \email moncef.gabbouj@tuni.fi\\
      \addr Department of Computing Sciences\\
      Tampere University, Finland      \\
      }

\def\month{MM}  
\def\year{YYYY} 
\def\openreview{\url{https://openreview.net/forum?id=XXXX}} 

\begin{document}

\maketitle

\begin{abstract}
Existing generalization theories of supervised learning typically take a holistic approach and provide bounds for the expected generalization over the whole data distribution, which implicitly assumes that the model generalizes similarly for all the classes. In practice, however, there are significant variations in generalization performance among different classes, which cannot be captured by the existing generalization bounds.
In this work, we tackle this problem by theoretically studying the class-generalization error, which quantifies the generalization performance of each individual class. 
We derive a novel information-theoretic bound for class-generalization error using the KL divergence, and we further obtain several tighter bounds using the conditional mutual information (CMI), which are significantly easier to estimate in practice. 
We empirically validate our proposed bounds in different neural networks and show that they accurately capture the complex class-generalization error behavior. Moreover, we show that the theoretical tools developed in this paper can be applied in several applications beyond this context.
\end{abstract}

\section{Introduction} \label{introduction_sec}


Despite the considerable progress towards a theoretical foundation for neural networks~\citep{he2020recent}, a comprehensive understanding of the generalization behavior of deep learning is still elusive~\citep{zhang2016understanding,zhang2021understanding}. Over the past decade, several approaches have been proposed to uncover and provide a theoretic understanding of the different facets of generalization~\citep{kawaguchi2017generalization,he2020recent,roberts_yaida_hanin_2022}. In particular, multiple tools have been used to characterize the expected generalization error of neural networks, such as VC dimension~\citep{sontag1998vc,harvey2017nearly}, algorithmic stability~\citep{bousquet2000algorithmic,hardt2016train}, algorithmic robustness~\citep{xu2012robustness,kawaguchi2022robustness}, and information-theoretic measures~\citep{xu2017information,steinke2020reasoning,wang2023tighter}. However, relying solely on the analysis of the expected generalization over the entire data distribution may not provide a complete picture. One key limitation of the standard expected generalization error is that it does not provide any insight into the class-specific generalization behavior, as it implicitly assumes that the model generalizes similarly for all the classes.

\begin{figure*}[t]
\centering
\includegraphics[width=0.99\linewidth]{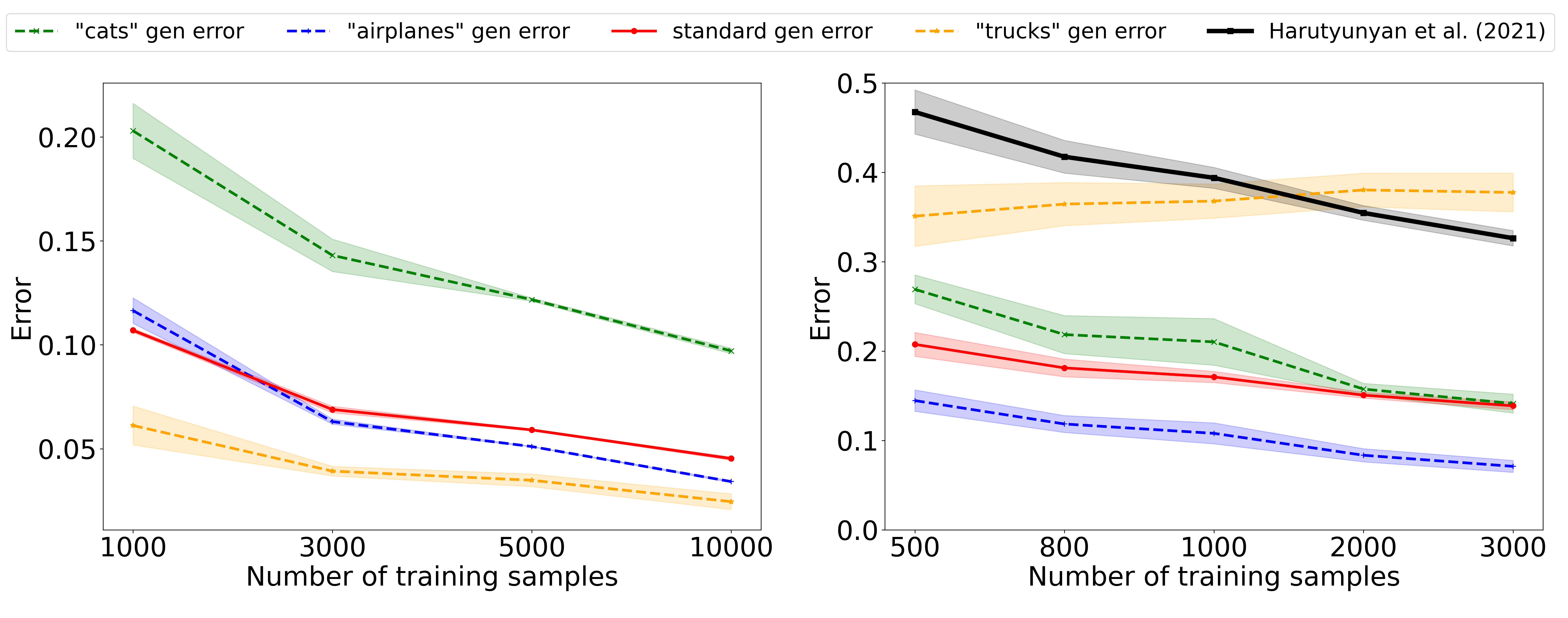}
\caption{Left: The standard generalization error, i.e., test loss - train loss, and the generalization errors for several classes on CIFAR10 as a function of number of training samples. Right: The standard generalization error, bound proposed by~\cite{harutyunyan2021information}, and the generalization errors for several classes on noisy CIFAR10. Experimental details are available in Section~\ref{num_results}.}
\label{motivating_example}
\end{figure*}

\textbf{Does the model generalize equally for all classes?} To answer this question, we conduct an experiment using deep neural networks, namely ResNet50  \citep{he2016deep} on the CIFAR10 dataset \citep{krizhevsky2009learning}. We plot the standard generalization error along with the class-generalization errors, i.e., the gap between the test error of the samples from the selected class and the corresponding training error, for three different classes of CIFAR10 in Figure~\ref{motivating_example} (left).  As can be seen, there are significant variants in generalization performance among different classes. For instance, the model overfits the ``cats'' class, i.e., large generalization error, and generalizes relatively well for the ``trucks'' class, with a generalization error of the former class consistently 4 times worse than the latter. This suggests that \textit{neural networks do not generalize equally for all classes}. However, reasoning only with respect to the standard generalization error (red curve) cannot capture this behavior.


Motivated by these observations, we conduct an additional experiment by introducing label noise ($5\%$) to the CIFAR10 dataset. Results are presented in Figure~\ref{motivating_example} (right). Intriguingly, despite the low level of noise, the disparities between the class-wise generalization errors are aggravated, with some classes generalizing up to 8 times worse than others. Further, as shown in this example, different classes do not exhibit the same trend when the number of training samples increases. For instance, unlike the other classes, the generalization error of the ``trucks'' class increases when more training data is available. 
To further illustrate the issue of standard generalization analysis, we plot the information-theoretic generalization bound proposed in \cite{harutyunyan2021information}. Although the bound captures the behavior of the standard generalization error well and can be used to explain the behavior of some classes (e.g., ``cat''), it becomes an invalid upper bound for the ``trucks'' class\footnote{We note that the bound by \cite{harutyunyan2021information} is proposed for the standard generalization error instead of the class-generalization. Here, we plot it only for illustrative purposes. }. 

When comparing the results on both datasets, it is also worth noting that the generalization error of the same class ``trucks''  behaves significantly differently on the two datasets. This suggests that class-wise generalization highly depends on factors beyond the class itself, including data distribution, learning algorithm, and the number of training samples. Moreover, in alignment with our findings,~\cite{NEURIPS2022_f73c0453,kirichenko2023understanding} showed that standard data augmentation and regularization techniques, e.g., weight decay and dropout~\citep{goodfellow2016deep} improve standard average generalization. However, it is surprising to note that these techniques inadvertently exacerbate the performance variations among different classes.


The main conclusion of all the aforementioned observations is that \textit{neural networks do not generalize equally for all classes, and their class-wise generalization depends on all ingredients of a learning problem}. This paper aims to provide some theoretical understanding of this phenomenon using information-theoretic bounds, as they are both distribution-dependent and algorithm-dependent \citep{xu2017information,neu2021information}. This makes them an ideal tool to characterize the class-generalization properties of a learning algorithm. A detailed related work discussion is presented in Appendix~\ref{related}. Our main contributions are as follows:

\begin{itemize}[leftmargin=*,itemsep=-0.1em]
\item We introduce the concept of ``class-generalization error,'' which quantifies the generalization performance of each individual class. We derive a novel information-theoretic bound for this quantity based on KL divergence (Theorem~\ref{MI_class_bound}). Then, using the super-sample technique by~\cite{steinke2020reasoning}, we derive various tighter bounds that are significantly easier to estimate and do not require access to the model's parameters (Theorems~\ref{CMI_class_bound}, ~\ref{fCMI_class_bound}, and~\ref{deltaLCMI_class_bound}). 

\item We validate our proposed bounds empirically in different neural networks using CIFAR10 and its noisy variant in Section~\ref{num_results}. We show that the proposed bounds can accurately capture the complex behavior of the class-generalization error behavior in different contexts.

\item We show that our novel theoretical tools can be applied to the following cases beyond the class-generalization error: (i) study how the class-generalization error affects the standard generalization (Section~\ref{sec:standard_gen}); (ii) provide tighter bounds for the subtask problem, where the test data only encompasses a specific subset of the classes encountered during training (Section~\ref{section_subtask}); (iii) derive generalization error bounds for learning in the presence of sensitive attributes (Section~\ref{section_fairness}).
\end{itemize} 

\textbf{Notations:} We use upper-case letters to denote random variables, e.g., $\rmZ$, and lower-case letters to denote the realization of random variables. $\E_{\rmZ \sim P}$ denotes the expectation of $\rmZ$ over a distribution $P$. Consider a pair of random variables $\rmW$ and $\rmZ=(\rmX,\rmY)$ with joint distribution $P_{\rmW,\rmZ}$. Let $\overline{\rmW}$ be an independent copy of $\rmW$, and $\overline{\rmZ}=(\overline{\rmX},\overline{\rmY})$ be an independent copy of $\rmZ$, such that $P_{\overline{\rmW},\overline{\rmZ}}= P_{\rmW} \otimes P_{\rmZ}$. For random variables $\rmX$, $\rmY$ and $\rmZ$, $I(\rmX;\rmY) \triangleq D(P_{\rmX,\rmY}\|P_{\rmX}\otimes P_{\rmY})$ denotes the mutual information (MI), and $I_{z}(\rmX;\rmY) \triangleq D(P_{\rmX,\rmY|\rmZ=z}\|P_{\rmX|\rmZ=z}\otimes P_{\rmY|\rmZ=z} )$ denotes disintegrated conditional mutual information (CMI), and $\E_\rmZ [I_{\rmZ}(\rmX;\rmY)] = I(\rmX;\rmY|\rmZ)$ is the standard CMI. We will also use the notation $\rmX,\rmY|z$ to simplify $\rmX,\rmY|\rmZ=z$ when it is clear from the context.

\vspace{-1em}

\section{Class-Generalization Error} \label{section_classgenerror}

\subsection{MI-setting} \label{section_misettings}
Typically, in supervised learning, the training set $\rmS = \{(\rmX_i,\rmY_i)\}_{i=1}^n = \{\rmZ_i\}_{i=1}^n$ contains $n$ i.i.d. samples $\rmZ_i \in \mathcal{Z}$ generated from the distribution $P_{\rmZ}$. Here, we are interested in the performance of a model with weights $w \in \mathcal{W}$ for data coming from a specific class $y \in \mathcal{Y}$. To this end, we define $\rmS_y$ as the subset of $\rmS$ composed of samples only in class $y$. For any model $w\in \mathcal{W}$ and fixed training sets $s$ and $s_y$, the class-wise empirical risk can be defined as follows:
\begin{equation}
    L_{E}(w,s_y) = \frac{1}{n_y} \sum_{(x_i,y) \in s_y} \ell(w,x_i,y),
\end{equation}
where $n_y$ is the size of $s_y$ ($n_y < n$), and $\ell: \mathcal{W} \times \mathcal{X} \times \mathcal{Y} \to \mathbb{R}_0^+$ is a non-negative loss function. In addition, the class-wise population risk that quantifies how well $w$ performs on the conditional data distribution $P_{\rmX|\rmY=y}$ is defined as
\begin{equation}
    L_{P}(w,P_{\rmX|\rmY=y}  ) =  \E_{P_{\rmX|\rmY=y}}[ \ell(w,\rmX,y)].
\end{equation}
A learning algorithm can be characterized by a randomized mapping from the entire training dataset $\rmS$ to model weights $\rmW$ according to a conditional distribution $P_{\rmW|\rmS}$.
The gap between $L_{P}(w,P_{\rmX|\rmY=y})$ and $L_{E}(w,s_y)$ measures how well the trained model $\rmW$ overfits with respect to the data with label $y$,  
and the expected class-generalization error is formally defined as follows.
\begin{definition} (class-generalization error)
 \label{class_gen} Given  $y \in \gY$, the class-generalization error is 
 \begin{equation} \label{eqclassgen}    \overline{\mathrm{gen}_y}(P_{\rmX,\rmY}, P_{\rmW|\rmS}) \triangleq \E_{P_{\rmW}} [  L_{P}(\rmW,P_{\rmX|\rmY=y}  )] -  \E_{P_{\rmW, \mS_y}} [  L_{E}(\rmW, \rmS_y )],
 \end{equation}
where $P_{\rmW}$ and $P_{\rmW,\rmS_y}$ are marginal distributions induced by the learning algorithm $P_{\rmW|\rmS}$ and data generating distribution $P_{\rmS}$.
\end{definition}

\paragraph{KL divergence bound} For most learning algorithms used in practice, the index of training samples $i$ will not affect the learned model. Thus, we assume that the learning algorithm is symmetric with respect to each training sample. Formally, this assumption can be expressed as follows:
\begin{assumption} \label{fixed_joint}
The marginal distribution $P_{\rmW,\rmZ_i}$ obtained from the joint distribution $P_{\rmW,\rmS}$, satisfying 
   $P_{\rmW,\rmZ_i} = P_{\rmW,\rmZ_j}$, $\forall i \neq j$,
\end{assumption}
Assumption~\ref{fixed_joint} states that the learned model $\rmW$ and each sample $\rmZ_i$ has the same joint distribution $P_{\rmW,\rmZ}$. Under this assumption, the following lemma simplifies the class-generalization error.
\begin{lemma}
\label{class_gen_lemma}  Under Assumption~\ref{fixed_joint}, the class-generalization error  in definition~\ref{class_gen} is given by
 \begin{equation}       \overline{\mathrm{gen}_y}(P_{\rmX,\rmY}, P_{\rmW|\rmS}) = \E_{P_{\overline{\rmW}}\otimes P_{\overline{\rmX}|y}} [\ell(\overline{\rmW},\overline{\rmX},y)] - \E_{P_{\rmW,\rmX|y}}[ \ell(\rmW,\rmX,y)].
 \end{equation}
\end{lemma}
Lemma~\ref{class_gen_lemma} shows that, similar to the standard generalization error \citep{xu2017information,bu2020tightening,zhou2022individually}, the class-wise generalization error can be expressed as the difference between the loss evaluated
under the joint distribution and the product-of-marginal distribution. The key difference is that both expectations are taken with respect to conditional distributions ($\rmY=y$). 

The following theorem provides a bound for the class-generalization error in Definition~\ref{class_gen}. 

\begin{theorem} \label{MI_class_bound}
For $y \in \gY$, assume Assumption~\ref{fixed_joint} holds and the loss $\ell(\overline{\rmW},\overline{\rmX},y)$ is $\sigma_y$ sub-gaussian under $P_{\overline{\rmW}} \otimes P_{\overline{\rmX}|\overline{\rmY}=y }$, then  the  class-generalization error of class $y$ in Definition~\ref{class_gen} can be bounded as:
\begin{equation}
   |  \overline{\mathrm{gen}_y}(P_{\rmX,\rmY}, P_{\rmW|\rmS}) | \leq  \sqrt{2\sigma_{y}^2D(P_{\rmW,\rmX|y} ||P_{\rmW}\otimes P_{\rmX|\rmY=y}) }.
\end{equation}  
\end{theorem}
The full proof is given in Appendix~\ref{MI_class_bound_proof}, which utilizes Lemma~\ref{class_gen_lemma} and Donsker-Varadhan’s variational representation of the KL divergence.

Theorem~\ref{MI_class_bound} shows that the class-generalization error can be bounded using a class-dependent conditional KL divergence. Theorem~\ref{MI_class_bound} implies that classes with a lower conditional KL divergence between the conditional joint distribution and the product of the marginal can generalize better. To our best knowledge, the bound in Theorem~\ref{MI_class_bound} is the first label-dependent bound that can explain the variation of generalization errors among the different classes. 

We note that our class-generalization error bound is obtained by considering the generalization gap of each individual sample with label $y$. This approach, as shown in \cite{bu2020tightening,zhou2022individually,harutyunyan2021information}, yields tighter bounds using the mutual information (MI) between an individual sample and the output of the learning algorithm, compared to the conventional bounds relying on the MI of the total training set and the algorithm's output \citep{xu2017information}. 


\subsection{CMI-setting} \label{section_cmisettings}
 
One limitation of the proposed bound in Theorem~\ref{MI_class_bound} is that it can be vacuous or even intractable to estimate in practice, as the bound involves high dimensional entities, i.e., model weights $\rmW$ and dataset $\rmS_y$. 
The conditional mutual information (CMI) framework, as pioneered by \cite{steinke2020reasoning}, has been shown in recent studies \citep{zhou2022individually, wang2023tighter} to offer tighter bounds on generalization error. Remarkably, CMI bounds are always finite even if the weights $\rmW$ are high dimensional and continuous. 

In this section, we extend our analysis using the CMI framework. In particular, we assume that there are $n$ super-samples $\rmZ_{[2n]}= (\rmZ^{\pm}_1, \cdots, \rmZ^{\pm}_n) \in \gZ^{2n} $ i.i.d generated  from $P_{\rmZ}$. The training data $\rmS = (\mZ^{\rmU_1}_1, \rmZ^{\rmU_2}_2, \cdots,\rmZ^{\rmU_n}_n) $ are selected from $\rmZ_{[2n]}$, where $\rmU=(\rmU_1, \cdots, \rmU_n) \in \{-1,1\}^n$ is the selection vector composed of $n$ independent Rademacher random variables.  Intuitively, $\rmU_i$ selects sample ${\rmZ^{\rmU_i}_i}$ from ${\rmZ^{\pm}_i}$ to be used in training, and the remaining  one ${\rmZ^{-\rmU_i}_i}$ is for the test.

For a specific class $y\in \mathcal{Y}$, let $n^y= n P(\rmY=y) $, the number of supersamples n scaled with the probability of class $y$. We define the class-generalization error in the CMI setting as follows:
\begin{definition} (super-sample-based class-generalization error) 
\label{CMI_class_gen}
For any $y \in \gY$, the class-generalization error is defined as
\begin{equation} 
\small
         \overline{\mathrm{gen}_y}(P_{\rmX,\rmY}, P_{\rmW|\rmS}) \triangleq\E_{\rmZ_{[2n]}}  \Big[ \frac{1}{n^y}  \sum_{i=1}^n   \E_{\rmU_i,\mW|  \rmZ_{[2n]}}  \big[  \mathds{1}_{\{Y_i^{-U_i}=y\}} \ell(\mW,  \rmZ^{-\rmU_i}_i)    - \mathds{1}_{\{Y_i^{U_i}=y\}} \ell(\rmW, \rmZ^{\rmU_i}_i) \big]\Big],
 \end{equation}
where $\mathds{1}_{\{a=b\}}$ is the indicator function, returning 1 when $a=b$ and zero otherwise.
\end{definition}

Similar to Definition~\ref{class_gen}, the class-generalization error in Definition~\ref{CMI_class_gen} measures the expected error gap between the training set and the test set relative to one specific class $y$. Compared to the standard generalization error definition typically used in the super-sample setting \citep{steinke2020reasoning,zhou2022individually}, we highlight two key differences: (i) our class-wise generalization error involves indicator functions to consider only samples belonging to a specific class $y$. (ii) Our generalization error is normalized by $\smash{n_{\rmZ_{[2n]}}^y/2}$, which is half of the total samples in $\rmZ_{[2n]}$ with label $y$. In contrast, the risks are averaged over $n$ samples in the standard super-sample setting.


\paragraph{Class-CMI bound.} The following theorem provides a bound for the super-sample-based class-generalization error using the disintegrated conditional mutual information between $\rmW$ and the selection variable $\rmU_i$ conditioned on super-sample $\rmZ_{[2n]}$. 

\begin{theorem}[class-CMI]\label{CMI_class_bound} 
Assume that the loss $\ell(w,x,y) \in [0,1]$ is bounded, then the class-generalization error for class $y$ in Definition~\ref{CMI_class_gen} can be bounded as
\begin{equation}
|   \overline{\mathrm{gen}_y}(P_{\rmX,\rmY}, P_{\rmW|\rmS}) | \leq  \E_{\rmZ_{[2n]}}   
\Big[ \frac{1}{n^y}  \sum_{i=1}^n   \sqrt{2 \max(\mathds{1}_{\{\rmY_i^{-}=y\}}, \mathds{1}_{\{\rmY_i^{+}=y\}}) I_{\rmZ_{[2n]}} (\rmW; \rmU_i )} \Big]. \nonumber
 \end{equation}
\end{theorem}

The proof is based on Donsker-Varadhan’s variational representation of the KL divergence and Hoeffding's Lemma \citep{hoeffding1994probability}. The key idea of the proof is specifying the function in Donsker-Varadhan’s variational representation to match the definition of class-generalization error. Then, using the fact that for a fixed realization $z_{[2n]}$, $ \mathds{1}_{\{y^{\rmU_i}=y\}} \ell(\rmW, z^{\rmU_i}_i) - \mathds{1}_{\{y^{-\rmU_i}=y\}} \ell(\mW,  z^{-\rmU_i}_i) = \rmU_i (\mathds{1}_{\{y_i^{-}=y\}} \ell(\rmW,z^{-}_i) - \mathds{1}_{\{y_i^{+}=y\}} \ell(\rmW,  z^{+}_i)) $ coupled with Hoeffding's Lemma yields the desired result. The full proof is provided in Appendix~\ref{CMI_class_bound_proof}.

Theorem~\ref{CMI_class_bound} provides a bound of the class-generalization error with explicit dependency on the model weights $\rmW$. It implies that the class-generalization error depends on how much information the random selection reveals about the weights when at least one of the two samples of $z_i^{\pm}$ corresponds to the class of interest $y$. Links between overfitting and memorization in the weights have been established in \cite{zhang2016understanding,arpit2017closer,chatterjee2018learning}. Here, we also see that if the model parameters $\rmW$ memorize the random selection $\rmU$, the CMI and the class-generalization error will exhibit high values.


\paragraph{Class-f-CMI bound.} While the bound in Theorem~\ref{CMI_class_bound} is always finite as $\rmU_i$ is binary, evaluating $ I_{\rmZ_{[2n]}} (\rmW; \rmU_i )$ in practice can be challenging, especially when dealing with high-dimensional $\rmW$ as in deep neural networks. One way to overcome this issue is by considering the predictions of the model $f_{\rmW}(\rmX^\pm_i)$ instead of the model weights $\rmW$, as proposed by~\cite{harutyunyan2021information}. Here, we re-express the loss function $\ell$ based on the prediction $\hat{y}=f_w(x)$ as  $\ell(w,x,y) = \ell(\hat{y},y)= \ell(f_w(x),y)$. Throughout the rest of the paper, we use these two expressions of loss functions interchangeably when it is clear from the context. 

In the following theorem, we bound the class-generalization error based on the disintegrated CMI between the model prediction $f_{\rmW}(\rmX^\pm_i)$  and the random selection, i.e., $I_{\rmZ_{[2n]}} (f_{\rmW}(\rmX^\pm_i); \rmU_i )$. 


\begin{theorem} (class-f-CMI)\label{fCMI_class_bound}
Assume that the loss $\ell(\hat{y},y) \in [0,1]$ is bounded, then the class-generalization error for class $y$ in Definition~\ref{CMI_class_gen} can be bounded as
\begin{equation}
|    \overline{\mathrm{gen}_y}(P_{\rmX,\rmY}, P_{\rmW|\rmS}) | \leq \E_{\rmZ_{[2n]}}\Big[     \frac{1}{n^y}  \sum_{i=1}^n   \sqrt{2 \max(\mathds{1}_{\{\rmY_i^{-}=y\}}, \mathds{1}_{\{\rmY_i^{+}=y\}}) I_{\rmZ_{[2n]}} (f_{\rmW}(\rmX^\pm_i); \rmU_i )} \Big]. \nonumber
 \end{equation}
\end{theorem}

The main benefit of the class-f-CMI bound in Theorem~\ref{fCMI_class_bound}, compared to all previously presented bounds, lies in the evaluation of the CMI term involving a typically low-dimensional random variable $f_{\rmW}(\rmX^\pm_i)$ and a binary random variable $\rmU_i$. For example, in the case of binary classification, $f_{\rmW}(\rmX^\pm_i)$ will be a pair of two binary variables, which enables us to estimate the class-f-CMI bound efficiently and
accurately, as will be shown in Section~\ref{num_results}. 
\begin{remark}
In contrast to the bound in Theorem~\ref{CMI_class_bound}, the bound in Theorem~\ref{fCMI_class_bound} does not require access to the model parameters $\rmW$. It only requires the model output $f(\cdot)$, which makes it suitable even for non-parametric approaches and black-box algorithms.
\end{remark}
\begin{remark}
Both our bounds in Theorems~\ref{CMI_class_bound} and~\ref{fCMI_class_bound} involve the term $\max(\mathds{1}_{\{\rmY_i^{-}=y\}},\mathds{1}_{\{\rmY_i^{+}=y\}})$, which filters out the CMI terms where neither sample $Z_i^+$ nor $Z_i^-$ corresponds to the class $y$. However, this term does not require both samples $Z_i^{\pm}$ to belong to class $y$. In the case that one sample in the pair $(Z_i^-,Z_i^+)$ is from class $y$ and the other is from a different class, this term is non-zero and the information from both samples of the pair contributes to the bound ($ I_{\rmZ_{[2n]}} (f_{\rmW}(\rmX^\pm_i); \rmU_i )$). From this perspective, samples from other classes ($\neq y$) can still affect these bounds, leading to loose bounds on the generalization error of class $y$.
\end{remark}

\paragraph{Class-$\Delta_y L$-CMI bound.}
The term $\max(\mathds{1}_{\{\rmY_i^{-}=y\}},\mathds{1}_{\{\rmY_i^{+}=y\}})$ is a proof artifact and makes the bound inconveniently loose. One solution to overcome this is to consider a new random variable $\Delta_y \rmL_i$ based on the indicator function and the loss, i.e., $\Delta_y \rmL_i \triangleq \mathds{1}_{\{y_i^{-}=y\}} \ell(f_{\rmW}(\rmX_i)^{-}, y^{-}_i) - \mathds{1}_{\{y_i^{+}=y\}} \ell(f_{\rmW}(\rmX_i)^+,  y^{+}_i)$. As shown in \cite{wang2023tighter,hellstrom2022new}, using the difference of the loss functions on $\rmZ_i^\pm$ instead of the model output yields tighter generalization bounds for the standard generalization error. We note that $\Delta_y \rmL_i$ is fundamentally different than  $\rmL_i$ introduced in \cite{wang2023tighter}. $\rmL_i$ simply represents the difference in loss functions, while $\Delta_y \rmL_i$ can be interpreted as a weighted sum of class-dependent losses.
The following Theorem provides a bound based on the CMI between this newly introduced variable and the random selection. 

\begin{theorem} (class-$\Delta_y L$-CMI) \label{deltaLCMI_class_bound}
Assume that the loss $\ell(\hat{y},y) \in [0,1]$, then the  class-generalization error of class $y$ defined in~\ref{CMI_class_gen} can be bounded as
    \begin{equation}
|    \overline{\mathrm{gen}_y}(P_{\rmX,\rmY}, P_{\rmW|\rmS}) | \leq   \E_{\rmZ_{[2n]}} \Big[  \frac{1}{n^y}  \sum_{i=1}^n   \sqrt{2 I_{\rmZ_{[2n]}} (\Delta_y \rmL_i; \rmU_i )} \Big].
    \end{equation}
    Moreover, we have the $\Delta_y L$-CMI bound is always tighter than the class-$f$-CMI bound in Theorem~\ref{fCMI_class_bound}, and the latter is always tighter than the class-CMI bound in Theorem~\ref{CMI_class_bound}. 
\end{theorem}
Unlike the bound in Theorem~\ref{fCMI_class_bound}, the bound in Theorem~\ref{deltaLCMI_class_bound} does not directly rely on the model output $f(\cdot)$. Instead, it only requires the loss values for both $\rmZ_i^\pm$  to compute $\Delta_y \rmL_i$. 

 Intuitively, the difference between two weighted loss values, $\Delta_y \rmL_i$, reveals much less information about the selection process $\rmU_i$ compared to the pair $f_{\rmW}(\rmX^\pm_i)$. Another key advantage of the bound in Theorem~\ref{deltaLCMI_class_bound} compared to Theorem~\ref{fCMI_class_bound} is 
that computing the CMI term $I_{\rmZ_{[2n]}} (\Delta_y \rmL_i; \rmU_i )$ is simpler, given that $\Delta_y \rmL_i$ is a one-dimensional scalar, as opposed to the two-dimensional $f_{\rmW}(\rmX^\pm_i)$.


The result of Theorem~\ref{deltaLCMI_class_bound} provides an interesting interpretation of the class-generalization error in a communication setting. In particular, for fixed $z_{[2n]}$, $P_{\Delta_{y} \rmL_i|\rmU_i}$ specifies a memoryless channel with input $\rmU_i$ and output $\Delta_{y} \rmL_i$. Then, the $\Delta_y L$-CMI term can be interpreted as ``the rate of reliable communication''~\citep{shannon1948mathematical,gallager1968information} achievable with the Rademacher random variable $\rmU_i$ over this channel $P_{\Delta_{y} \rmL_i|\rmU_i}$. From this perspective, Theorem~\ref{deltaLCMI_class_bound} suggests that classes with \emph{lower} rates of reliable communication generalize better.

\section{Empirical Evaluations} \label{num_results}
In this section, we conduct empirical experiments to evaluate the effectiveness of our class-wise generalization error bounds. As mentioned earlier, The bounds in Section~\ref{section_cmisettings} are significantly easy to estimate in practical scenarios. Here, we evaluate the error bounds in Theorems~\ref{fCMI_class_bound} and ~\ref{deltaLCMI_class_bound} for deep neural networks.

We follow the same experimental settings in~\cite{harutyunyan2021information}, i.e., we fine-tune a  ResNet-50 \citep{he2016deep} on the CIFAR10 dataset \citep{krizhevsky2009learning} (Pretrained on ImageNet \citep{deng2009imagenet}). Moreover, to understand how well our bounds perform in a more challenging situation and to further highlight their effectiveness, we conduct an additional experiment with a noisy variant (5\% label noise) of CIFAR10. The experimental details are provided in Appendix~\ref{expi_setup}. 

The class-wise generalization error of two classes from CIFAR10 ``trucks'' and ``cats'', along with the bounds in Theorems~\ref{fCMI_class_bound} and ~\ref{deltaLCMI_class_bound} are presented in the first two columns of Figure~\ref{CMIbounds_results}. The results on all the 10 classes for both datasets are presented in Appendix~\ref{classgen_results}.

\begin{figure*}[h!]
\centering
\includegraphics[width=0.345\linewidth]{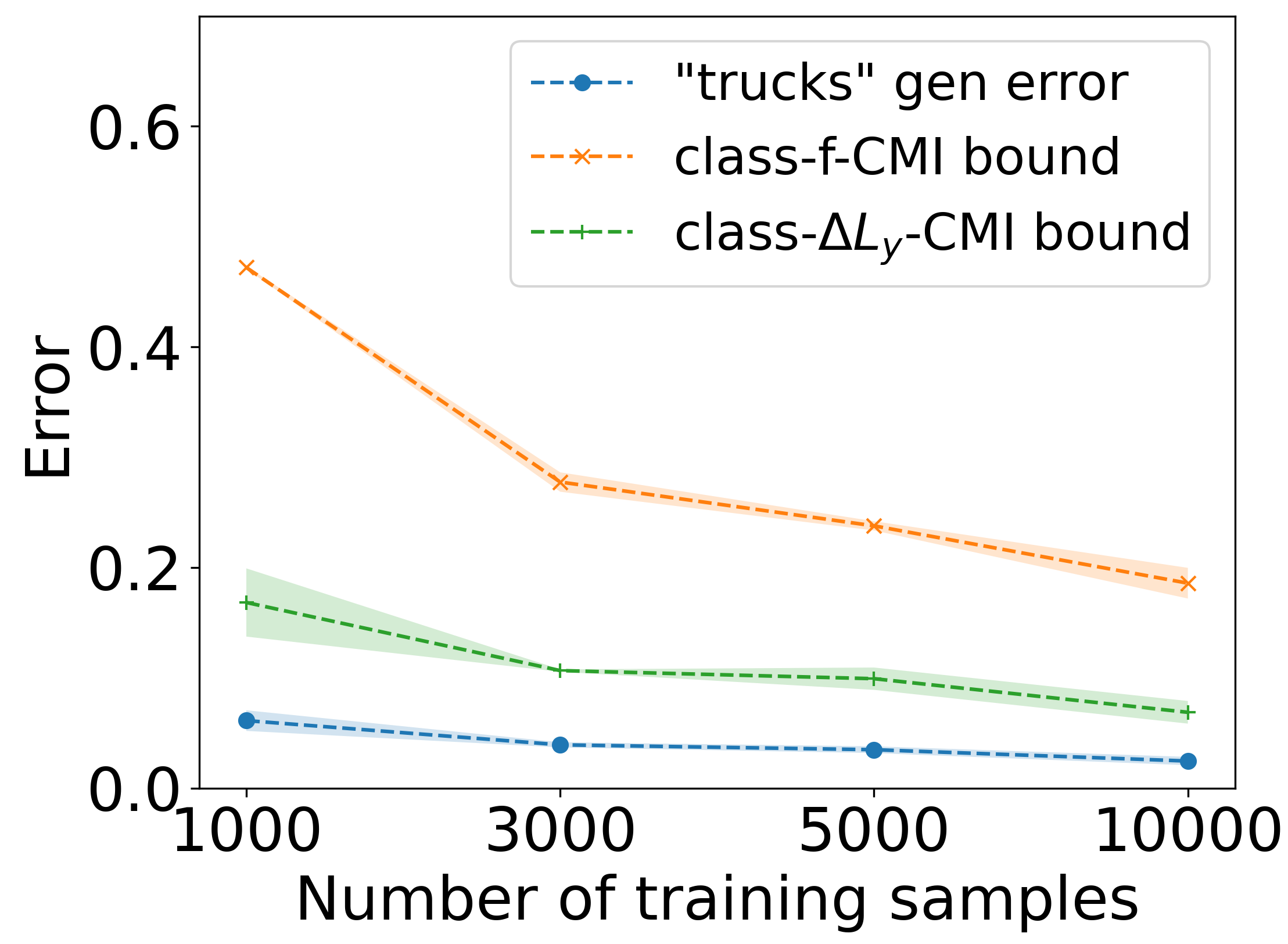}
\includegraphics[width=0.345\linewidth]{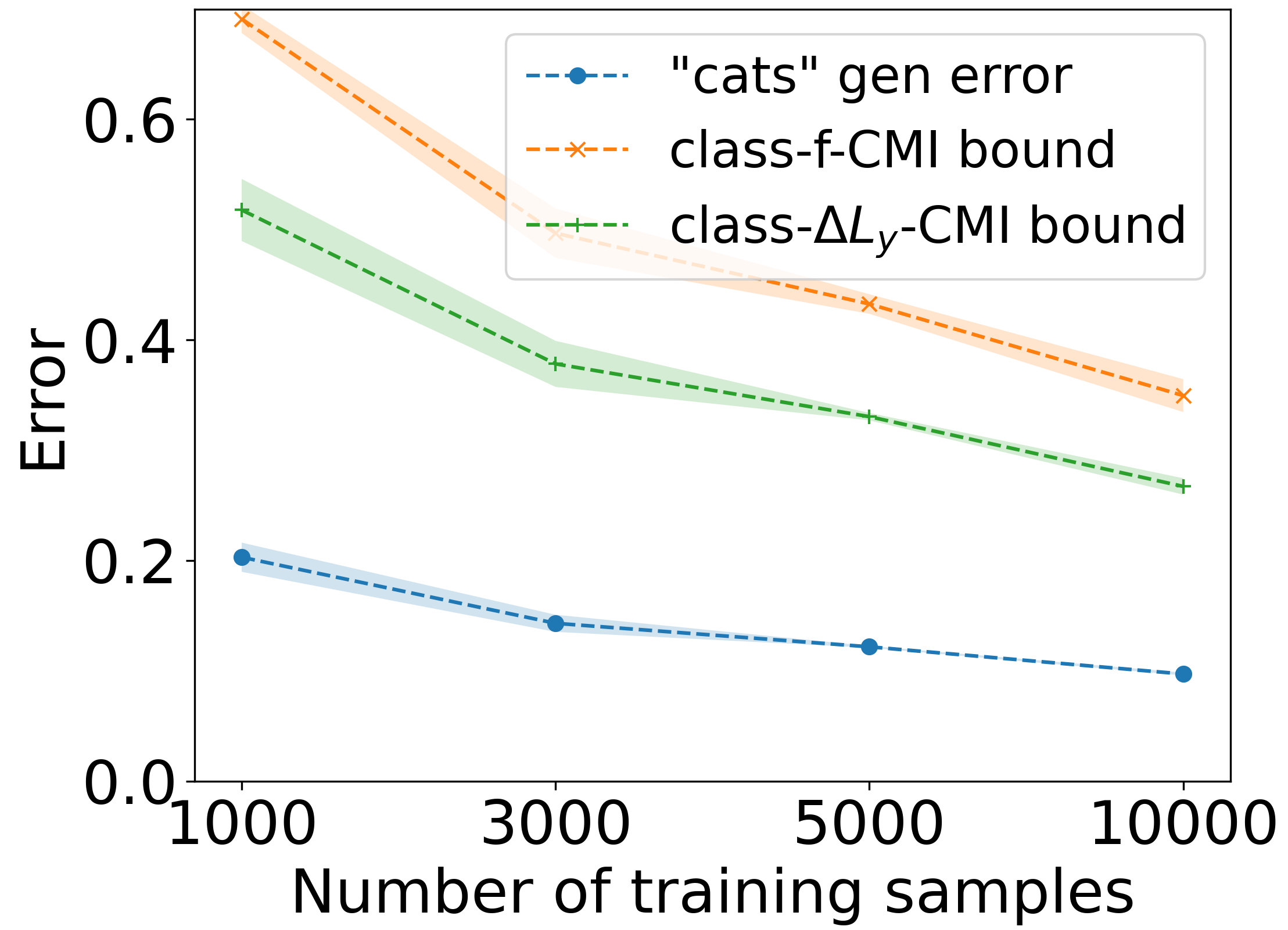}
\includegraphics[width=0.29\linewidth]{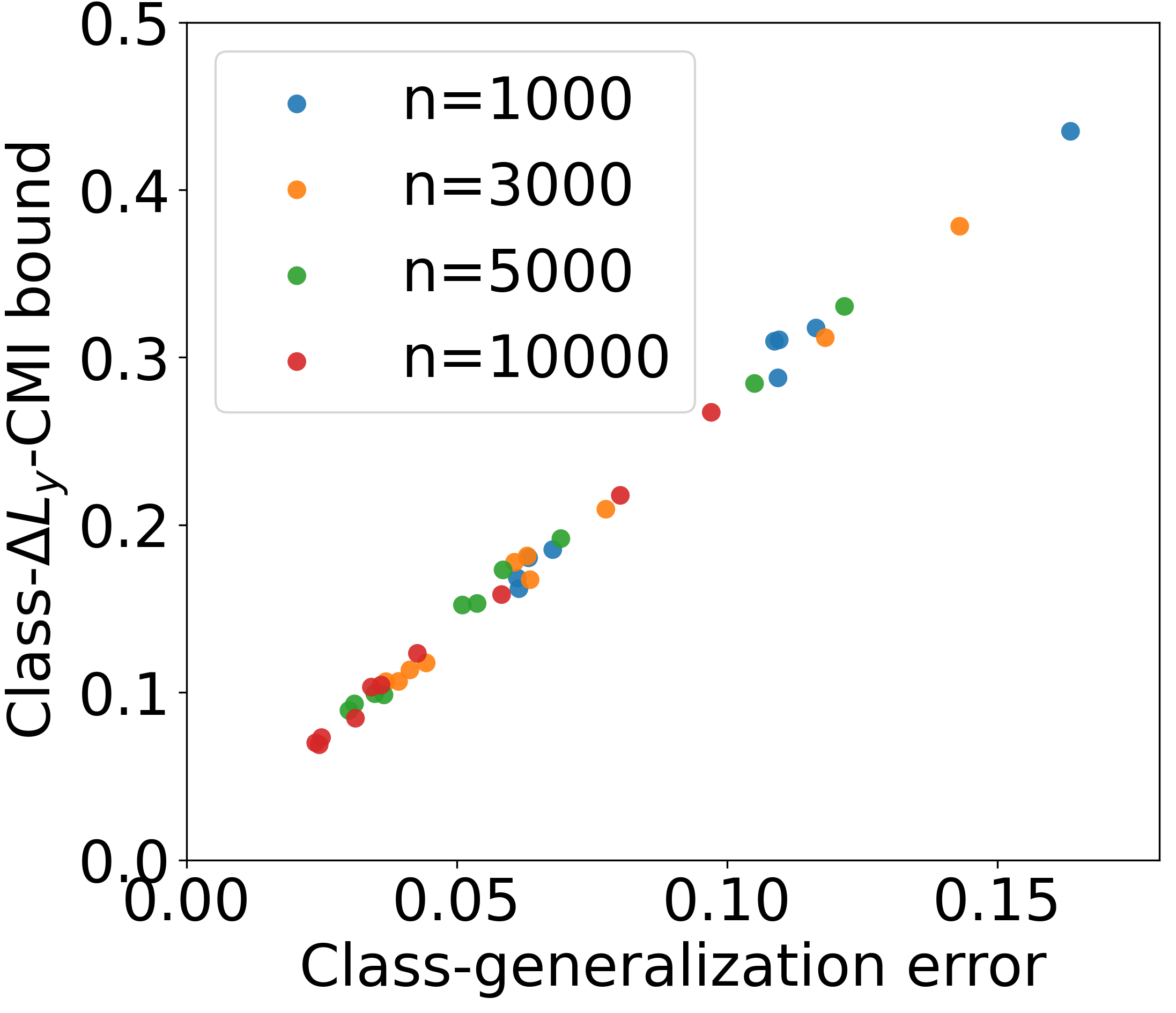}
\includegraphics[width=0.345\linewidth]{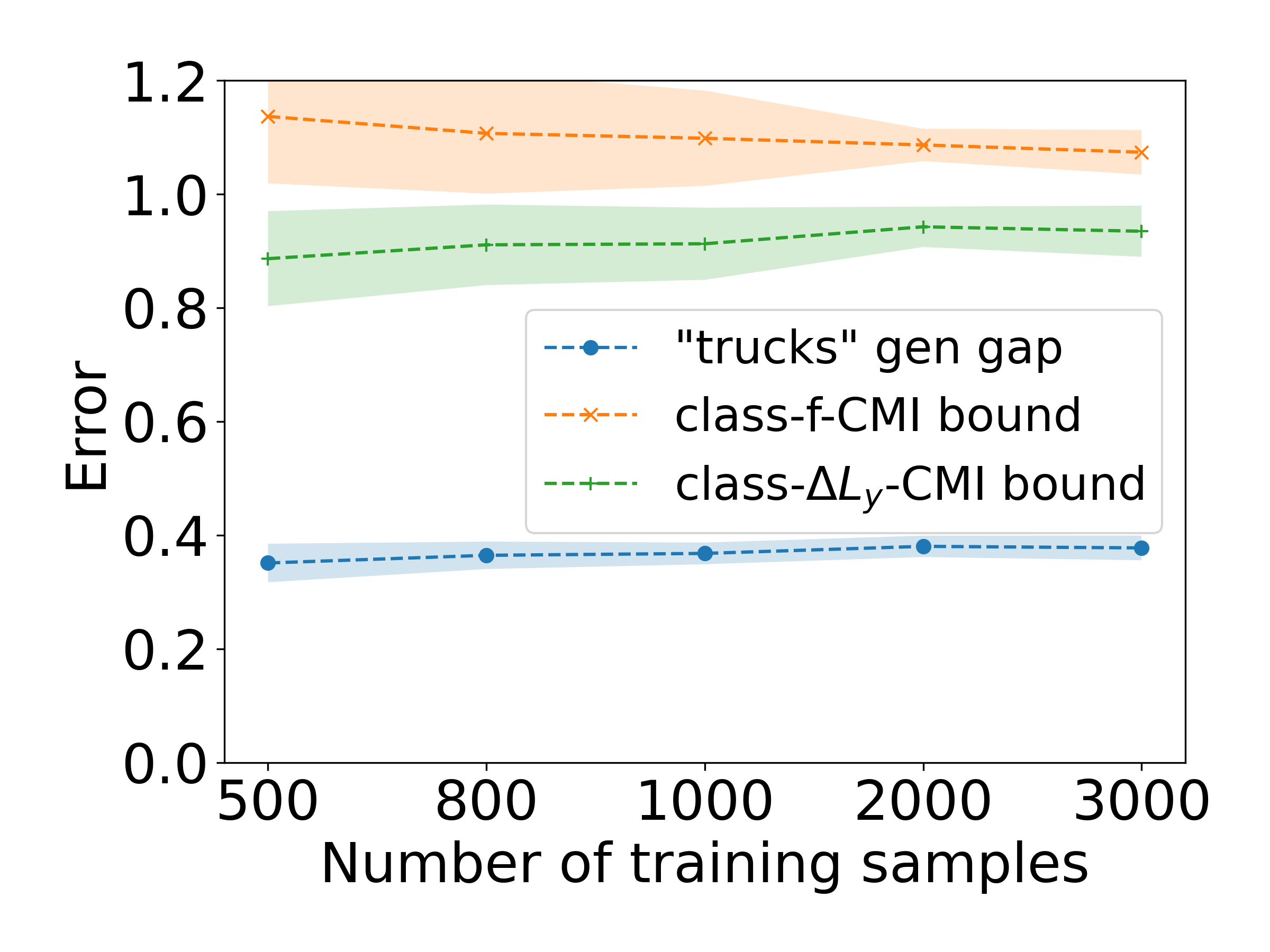}
\includegraphics[width=0.345\linewidth]{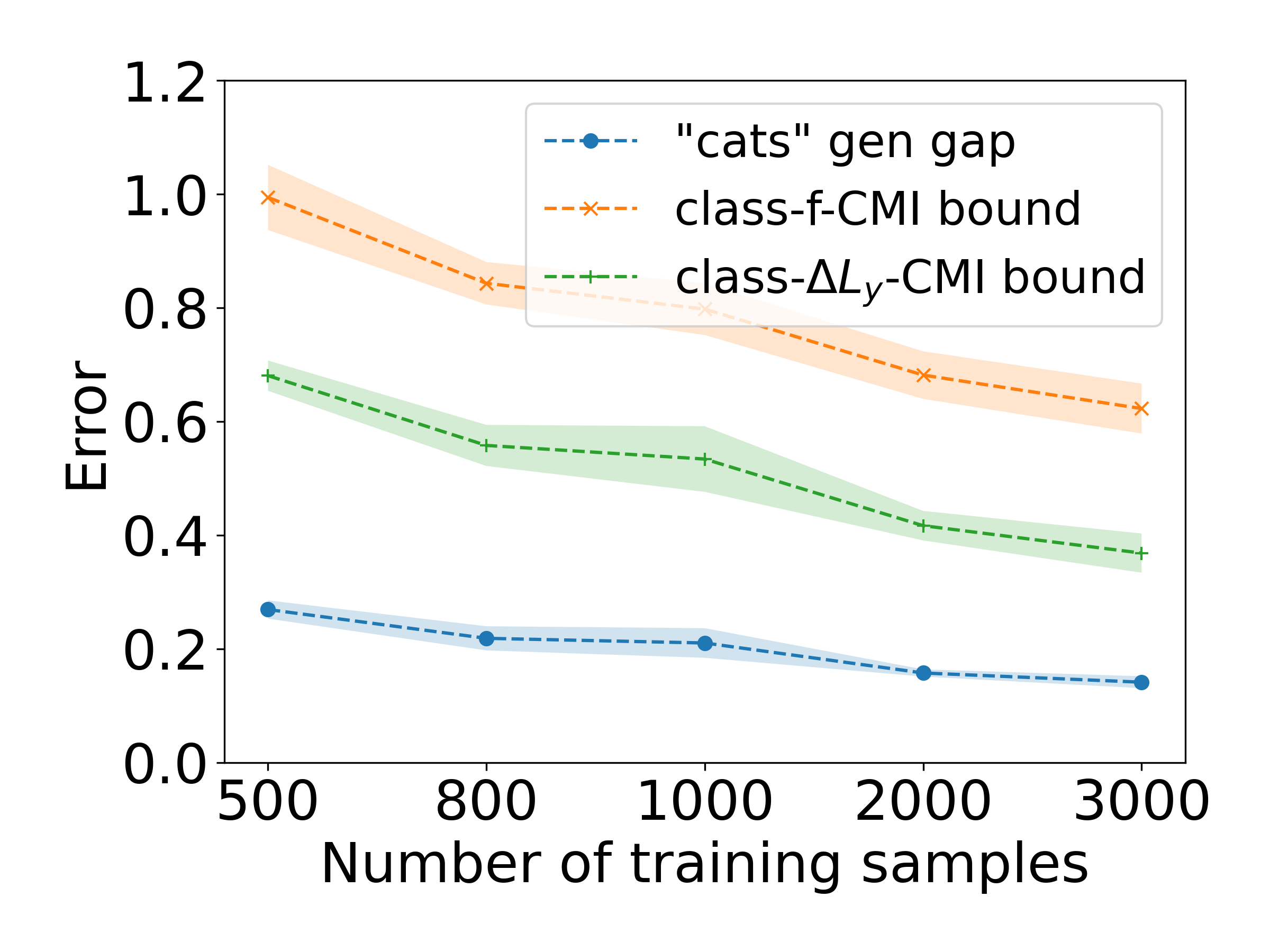}
\includegraphics[width=0.29\linewidth]{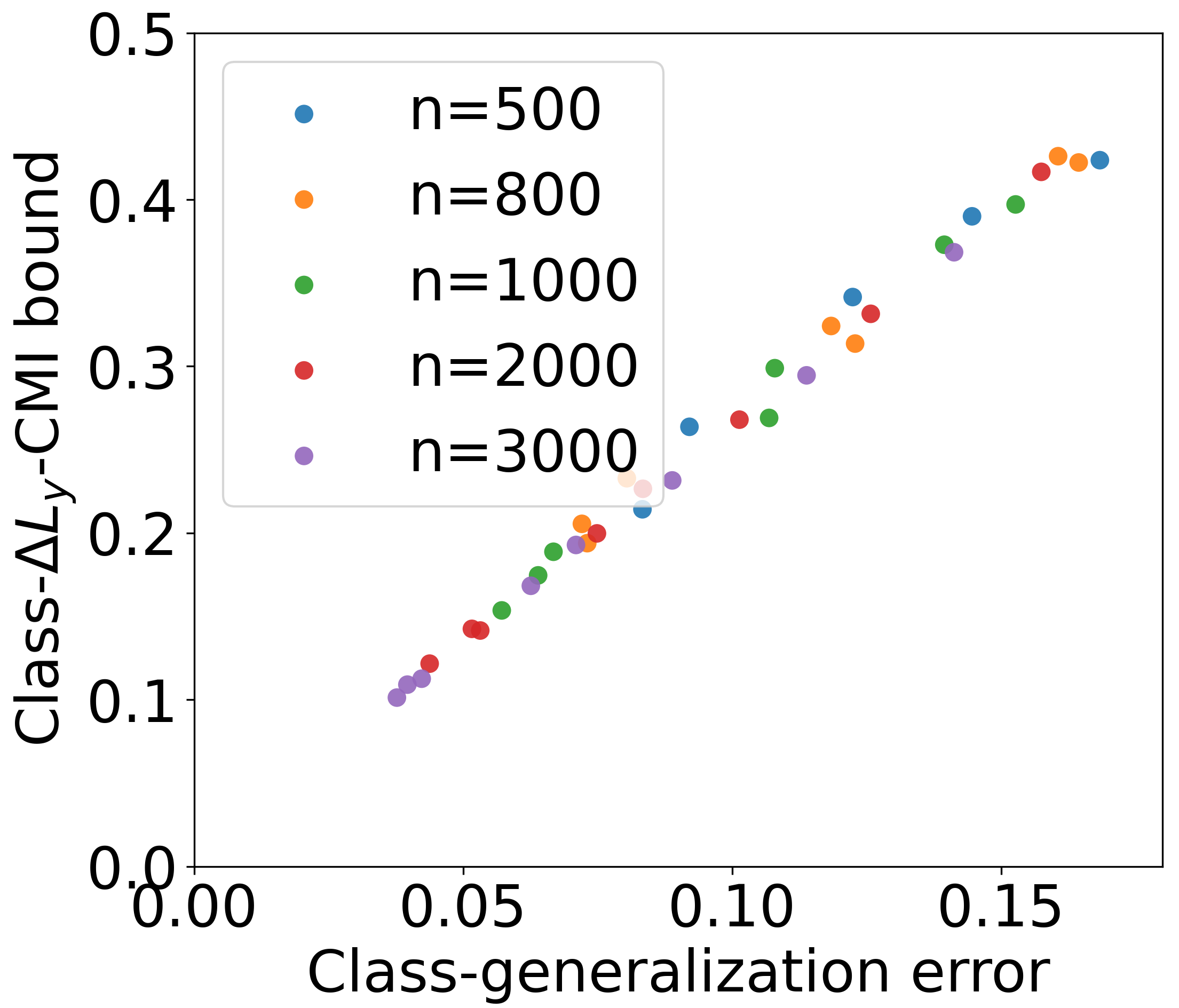}
\caption{Experimental results of class-generalization error and our bounds in Theorems~\ref{fCMI_class_bound} and~\ref{deltaLCMI_class_bound} for the class of ``trucks'' (left) and ``cats'' (middle) in CIFAR10 (top) and noisy CIFAR10 (bottom), as we increase the number of training samples. In the right column, we provide the scatter plots between the bound in Theorem~\ref{deltaLCMI_class_bound} and the class-generalization error of the different classes for CIFAR10 (top) and noisy CIFAR10 (bottom). }
\label{CMIbounds_results}
\end{figure*}

Figure~\ref{CMIbounds_results} shows that both bounds can capture the behavior of the class-generalization error. As expected, the class-$\Delta_y L$-CMI is consistently tighter and more stable compared to the class-$f$-CMI bound for all the different scenarios.  For CIFAR10 in Figure~\ref{CMIbounds_results} (top), as we increase the number of training samples, the ``trucks'' class has a relatively constant class-generalization error, while the ``cats'' class has a large slope at the start and then a steady incremental decrease. 
For both classes, the class-$\Delta_y L$-CMI precisely predicts the behavior of class-generalization error. 

The results on noisy CIFAR10 in Figure~\ref{CMIbounds_results} (bottom) and the results in Appendix~\ref{classgen_results} are consistent with these observations. Notably, the ``trucks'' generalization error decreases for CIFAR10 and increases for noisy CIFAR10 with respect to the number of samples. Moreover, the class-generalization error of ``cat'' is worse than ``trucks'' in CIFAR10, but the opposite is true for the noisy CIFAR10. All these complex behaviors of class-generalization errors are successfully captured by the class-$\Delta_y L$-CMI bound. 

As can be seen from the left and middle plots in Figure~\ref{CMIbounds_results}, the class-$\Delta_y L$-CMI bound scales proportionally to the true class-generalization error, i.e., higher class-$\Delta_y L$-CMI bound value corresponds to higher class-generalization error. To further highlight this dependency, we plot in Figure~\ref{CMIbounds_results} (right) the scatter plot between the different class-generalization errors and their corresponding class-$\Delta_y L$-CMI bound values for the different classes in CIFAR10 (top) and Noisy CIFAR10 (bottom) under different number of samples. We observe that our bound is linearly correlated with the class-generalization error and can efficiently predict its behavior. We also note that $f$-CMI bound exhibits a similar behavior as shown in Appendix~\ref{classgen_results}. This suggests that one can use our bound to predict which classes will generalize better than others. Additional results are provided in Appendix~\ref{app_recall}.

\section{Other applications} \label{otherapp_section}

Besides enabling us to study class-wise generalization errors, the tools developed in this work can also be used to provide theoretical insights into several other applications. In this Section, we explore several use cases for the developed tools.

\subsection{From class-generalization error to standard generalization error} \label{sec:standard_gen}
In this subsection, we study the connection between the standard expected generalization error and the class-generalization error. We extend the bounds presented in Section~\ref{section_classgenerror} into class-dependent expected generalization error bounds.

First, we notice that taking the expectation over $P_\rmY$ for the class-generalization error defined in~\ref{class_gen} yields the standard expected generalization error. 
Thus, we obtain the first class-dependent bound for the standard generalization error by taking the expectation of $y \sim P_{\rmY}$ in Theorem~\ref{MI_class_bound}.  

\begin{corollary} \label{MI_geny2gen}
    Assume that for every $y \in \gY$, the loss $\ell(\overline{\rmW},\overline{\rmX},y)$ is $\sigma_y$ sub-gaussian under $P_{\overline{\rmW}} \otimes P_{\overline{\rmX}|\overline{\rmY}=y }$, then 
\begin{equation}
  | \overline{\mathrm{gen}}(P_{\rmX,\rmY}, P_{\rmW|\rmS}) | \leq \E_{Y'} \bigg[  \sqrt{2\sigma_{\rmY'}^2D(P_{\rmW|\rmZ} \otimes P_{\rmX|\rmY=\rmY'} ||P_{\rmW}\otimes P_{\rmX|\rmY=\rmY'}) } \bigg].
\end{equation}
\end{corollary}

We note that in the case where sub-gaussian parameter $\sigma_y = \sigma$ is independent of $y$, we can further show that the bound in~\ref{MI_geny2gen} is tighter than the individual sample bound in~\cite{bu2020tightening}. The proof is available in Appendix~\ref{coro_bu2020}.
We also extend the results in Theorems~\ref{CMI_class_bound}, ~\ref{fCMI_class_bound}, and~\ref{deltaLCMI_class_bound} into standard generalization bounds. For example, in Corollary~\ref{deltaCMI_geny2gen}, we provide such an extension of  Theorem~\ref{deltaLCMI_class_bound}. 



\begin{corollary} \label{deltaCMI_geny2gen}
    Assume that the loss $\ell(\hat{y},y) \in [0,1]$, then 
\begin{equation}
  | \overline{\mathrm{gen}}(P_{\rmX,\rmY}, P_{\rmW|\rmS}) | \leq \E_{\mY} \bigg[ \E_{\rmZ_{[2n]}}   
\Big[ \frac{1}{n^\rmY}  \sum_{i=1}^n   \sqrt{2 I_{\rmZ_{[2n]}} (\Delta_{\rmY} \rmL_i;\rmU_i )} \Big] \bigg].
\end{equation}
\end{corollary}

To the best of our knowledge,  Corollaries~\ref{MI_geny2gen} and~\ref{deltaCMI_geny2gen} are the first generalization bounds to provide explicit dependency on the labels. While prior bounds~\citep{steinke2020reasoning,harutyunyan2021information,wang2022information,wang2023tighter} are tight and efficient to estimate, they fail to provide any insights on how different classes affect the standard generalization error. The results presented here address this gap and provide explicit label-dependent bounds. It shows that the generalization error of an algorithm scales as the expectation of the bound depending only on a single class. 
Moreover, in classification with $m$ classes, the bounds become a sum of each class-generalization error weighted by the probability of the class, i.e., $P(\rmY = y)$. This shows that classes with a higher occurrence probability affect the standard generalization error more. From this perspective, our results can also provide insights into developing algorithms with better generalization capabilities by focusing on the class-generalization error. For example, one can employ augmentation techniques targeted at the classes with higher class-generalization error bounds to attenuate their respective class error and thus improve the standard generalization of the model.

\subsection{Sub-task problem} \label{section_subtask}
Here, ``\textit{subtask problem}'' refers to a specific case of distribution shift in supervised learning, where the training data generated from the source domain $P_{\rmX,\rmY}$ consists of multiple classes, while the test data for the target domain $Q_{\rmX,\rmY}$ only encompasses a specific known subset of the classes encountered during training. This problem is motivated by the situation where a large model has been trained on numerous classes, potentially over thousands, but is being utilized in a target environment where only a few classes, observed during training, exist. By tackling the problem as a standard domain adaptation task, the generalization error of the subtask problem can be bounded as follows: 
\begin{equation} \label{basic_bound}
\overline{\mathrm{gen}}_{Q,E_P} \triangleq \E_{P_{\rmW,\rmS}} [  L_Q(\rmW) -  L_{E}(\rmW, \rmS) ] \leq  \sqrt{2 \sigma^2 D(Q_{\rmX,\rmY}\|P_{\rmX,\rmY})} + \sqrt{2 \sigma^2 I(\rmW;\rmS)}, 
\end{equation}
where $L_Q(w) = L_P(w, Q_{\rmX,\rmY})$ denotes the population risk of $w$ under distribution $Q_{\rmX,\rmY}$.
The full details of deriving such bound are available in appendix~\ref{subtask_sup}. We note that~\citep{wu2020information} further tightens the result in~\eqref{basic_bound}, but these bounds are all based on the KL divergence $D(Q_{\rmX,\rmY}\|P_{\rmX,\rmY})$ for any generic distribution shift problem and do not leverage the fact that the target task is encapsulated in the source task.

Obtaining tighter generalization error bounds for the subtask problem is straightforward using our class-wise generalization bounds.  In fact, the generalization error bound of the subtask can be obtained by summing the class-wise generalization over the space of the subtask classes $\gA$. Formally, by taking the expectation of $\rmY \sim Q_{\rmY}$, we obtain the following notion of the subtask generalization error:
\begin{equation}
\label{main_gen_yq}
\overline{\mathrm{gen}}_{Q,E_{Q}} \triangleq  \E_{Q_\rmY} \big[ \overline{\mathrm{gen}}_\rmY\big] 
 =  \E_{P_{\rmW,\rmS}} [  L_Q(w) -  L_{E_Q}(\rmW, \rmS)
 ], 
\end{equation}
where  $L_{E_Q}(w, S) = \frac{1}{n_{\gA}} \sum_{y_i \in \gA} \ell(w,x_i,y_i)$ is the empirical risk relative to the target domain $Q$, and $n_{\gA}$ is the number of samples in $S$ such that their labels $y_i \in \gA$. 
We are interested in deriving generalization bounds for $\overline{\mathrm{gen}}_{Q,E_{Q}}$, as it only differs from $\overline{\mathrm{gen}}_{Q,E_P}$ by the difference in the empirical risk $L_{E_Q}(\rmW, \rmS)-L_{E}(\rmW, \rmS)$, which can be computed easily in practice. 

Using Jensen's inequality, we have $ |   \overline{\mathrm{gen}}_{ Q,E_{Q}} | = |\E_{\rmY \sim Q_\rmY} \big[ \overline{\mathrm{gen}}_\rmY\big] | \leq \E_{\rmY \sim Q_\rmY} \big[ | \overline{\mathrm{gen}}_\rmY| \big] $. Thus, we can use the results from Section~\ref{section_classgenerror} to obtain tighter bounds. For example, using Theorem~\ref{deltaLCMI_class_bound}, we can obtain the corresponding subtask generalization error bound in Theorem~\ref{subtaskdeltacmi_bound}.

\begin{theorem} (subtask-CMI) \label{subtaskcmi_bound}
Assume that the loss $\ell(w,x,y) \in [0,1]$ is bounded, then the subtask generalization error  defined in~\ref{main_gen_yq} can be bounded as
\begin{equation}
|\overline{\mathrm{gen}}_{ Q,E_{Q}}|  \leq 
     \E_{\rmY \sim Q_\rmY} \bigg[  \E_{\rmZ_{[2n]}}  \Big[ \frac{1}{n^\rmY}  \sum_{i=1}^n   \sqrt{2 \max(\mathds{1}_{\rmY_i^{-}=\rmY},\mathds{1}_{\rmY_i^{+}=\rmY})  I_{\rmZ_{[2n]}} (\rmW; \rmU_i )} \Big] \bigg]. \nonumber
\end{equation}  
\end{theorem}

\begin{theorem} (subtask-$\Delta~L_y$-CMI) \label{subtaskdeltacmi_bound}
Assume that the loss $\ell(w,x,y) \in [0,1]$ is bounded, Then the subtask generalization error  defined in~\ref{main_gen_yq} can be bounded as
\begin{equation}
   |   \overline{\mathrm{gen}}_{Q,E_{Q}}|  \leq 
\E_{\rmY \sim Q_\rmY} \bigg[  \E_{\rmZ_{[2n]}}   
\Big[ \frac{1}{n^\rmY}  \sum_{i=1}^n    \sqrt{2 I_{\rmZ_{[2n]}} (\Delta_{\rmY} \rmL_i;\rmU_i )}\Big] \bigg]. \nonumber
\end{equation}  
\end{theorem}
Similarly, we can extend Theorems~\ref{fCMI_class_bound} and~\ref{CMI_class_bound} to use the model's output or weights instead of $\Delta_{y} \rmL_i$. 

\begin{remark}
Existing distribution shift bounds, e.g., the bound in \eqref{basic_bound}, typically depend on some measure that quantifies the discrepancy between the target and domain distributions, e.g., KL divergence. However, as can be seen in Theorem~\ref{subtaskdeltacmi_bound}, the proposed bounds here are discrepancy-independent and involve only a quantity analog to the mutual information term in \eqref{basic_bound}.
\end{remark}

\subsection{Generalization certificates with sensitive attributes } \label{section_fairness}

One main concern hindering the use of machine learning models in high-stakes applications is the potential biases on sensitive attributes such as gender and skin color~\citep{mehrabi2021survey,barocas2017fairness}. Thus, it is critical not only to reduce sensitivity to such attributes but also to be able to provide guarantees on the fairness of the models~\citep{holstein2019improving,rajkomar2018ensuring}. One aspect of fairness is that the machine learning model should generalize equally well for each minor group with different sensitive attributes~\citep{barocas2017fairness,williamson2019fairness}. 


By tweaking the definition of our class-generalization error, we show that the theoretical tools developed in this paper can be used to obtain bounds for attribute-generalization errors. Suppose we have a random variable $\rmT \in \gT$ representing a sensitive feature. One might be interested in studying the generalization of the model for the sub-population with the attribute $\rmT = t $. Inspired by our class-generalization, we define the attribute-generalization error as follows:

\begin{definition} (attribute-generalization error)
 \label{attri_gen} Given  $t \in \gT$, the attribute-generalization error is defined as follows:
\begin{equation}
       \overline{\mathrm{gen}_t}(P_{\rmX,\rmY}, P_{\rmW|\rmS}) = \E_{P_{\rmW}\otimes P_{\rmZ|\rmT=t}} [\ell(\overline{\rmW},\overline{\rmZ})] - \E_{P_{\rmW|\rmZ} \otimes P_{\rmZ|\rmT=t}}[ \ell(\rmW,\rmZ)]. 
\end{equation}
\end{definition}

By exchanging $\rmX$ and $\rmY$ with $\rmZ$ and $\rmT$ in Theorem~\ref{MI_class_bound}, respectively, we can show the following bound for the attribute-generalization error.
\begin{theorem} \label{MI_attribute_bound}
    Given $t \in \gT$, assume that the loss $\ell(\rmW,\rmZ)$ is $\sigma$ sub-gaussian under $P_{\overline{\rmW}} \otimes P_{\overline{\rmZ} }$, then  the  attribute-generalization error of the sub-population $\rmT=t$, can be bounded as follows:
\begin{equation}
   |  \overline{\mathrm{gen}_t}(P_{\rmX,\rmY}, P_{\rmW|\rmS}) | \leq \sqrt{2\sigma^2D(P_{\rmW|\rmZ} \otimes P_{\rmZ|\rmT=t}||P_{{\rmW}}\otimes P_{{\rmZ}|{\rmT}=t}) }.
\end{equation}
\end{theorem}

We note extending our results to the CMI settings is also straightforward. Using the attribute generalization, we can show that the standard expected generalization error can be bounded as follows:
\begin{corollary} \label{MI_genatt2gen}
    Assume that the loss $\ell(\rmW,\rmZ)$ is $\sigma$ sub-gaussian under $P_{\overline{\rmW}} \otimes P_{\overline{\rmZ} }$, then 
\begin{equation}
  | \overline{\mathrm{gen}}(P_{\rmX,\rmY}, P_{\rmW|\rmS}) | \leq \E_{\rmT'} \Big[  \sqrt{2\sigma^2D(P_{\rmW|\rmZ} \otimes P_{\rmZ|\rmT=\rmT'}||P_{{\rmW}}\otimes P_{\rmZ|\rmT=\rmT'}) }\Big].
\end{equation}
\end{corollary}

The result of Corollary~\ref{MI_genatt2gen} shows that the average generalization error is upper-bounded by the expectation over the attribute-wise generalization. This shows that it is possible to improve the overall generalization by reducing the generalization of each population relative to the sensitive attribute.

\section{Conclusion \& Future Work}


This paper studied the deep learning generalization puzzle of the noticeable heterogeneity of overfitting among different classes by introducing and exploring the concept of ``class-generalization error''. To our knowledge, we provided the first rigorous generalization bounds for this concept in the MI and CMI settings. We also empirically strengthened the findings with supporting experiments validating the efficiency of the proposed bounds. Furthermore, we demonstrated the versatility of our theoretical tools in providing tight bounds for various contexts. 

Overall, our goal is to understand generalization in deep learning through the lens of information theory, which motivates future work on how to prevent high class-generalization error variability and ensure 'equal' generalization among the different classes. Other possible future research endeavors focus on obtaining tighter bounds for the class-generalization error and studying this concept in different contexts beyond supervised learning, e.g., transfer and self-supervised learning.


\bibliography{main}

\begin{thebibliography}{56}
\providecommand{\natexlab}[1]{#1}
\providecommand{\url}[1]{\texttt{#1}}
\expandafter\ifx\csname urlstyle\endcsname\relax
  \providecommand{\doi}[1]{doi: #1}\else
  \providecommand{\doi}{doi: \begingroup \urlstyle{rm}\Url}\fi

\bibitem[Alabdulmohsin(2020)]{alabdulmohsin2020towards}
Ibrahim Alabdulmohsin.
\newblock Towards a unified theory of learning and information.
\newblock \emph{Entropy}, 22\penalty0 (4):\penalty0 438, 2020.

\bibitem[Aminian et~al.(2021)Aminian, Bu, Toni, Rodrigues, and Wornell]{aminian2021exact}
Gholamali Aminian, Yuheng Bu, Laura Toni, Miguel Rodrigues, and Gregory Wornell.
\newblock An exact characterization of the generalization error for the gibbs algorithm.
\newblock \emph{Advances in Neural Information Processing Systems}, 34:\penalty0 8106--8118, 2021.

\bibitem[Arpit et~al.(2017)Arpit, Jastrz{k{e}}bski, Ballas, Krueger, Bengio, Kanwal, Maharaj, Fischer, Courville, Bengio, et~al.]{arpit2017closer}
Devansh Arpit, Stanis{\l}aw Jastrz{k{e}}bski, Nicolas Ballas, David Krueger, Emmanuel Bengio, Maxinder~S Kanwal, Tegan Maharaj, Asja Fischer, Aaron Courville, Yoshua Bengio, et~al.
\newblock A closer look at memorization in deep networks.
\newblock In \emph{International conference on machine learning}, pp.\  233--242. PMLR, 2017.

\bibitem[Balestriero et~al.(2022)Balestriero, Bottou, and LeCun]{NEURIPS2022_f73c0453}
Randall Balestriero, Leon Bottou, and Yann LeCun.
\newblock The effects of regularization and data augmentation are class dependent.
\newblock In S.~Koyejo, S.~Mohamed, A.~Agarwal, D.~Belgrave, K.~Cho, and A.~Oh (eds.), \emph{Advances in Neural Information Processing Systems}, volume~35, pp.\  37878--37891. Curran Associates, Inc., 2022.
\newblock URL \url{https://proceedings.neurips.cc/paper_files/paper/2022/file/f73c04538a5e1cad40ba5586b4b517d3-Paper-Conference.pdf}.

\bibitem[Barocas et~al.(2017)Barocas, Hardt, and Narayanan]{barocas2017fairness}
Solon Barocas, Moritz Hardt, and Arvind Narayanan.
\newblock Fairness in machine learning.
\newblock \emph{Nips tutorial}, 1:\penalty0 2017, 2017.

\bibitem[Bitterwolf et~al.(2022)Bitterwolf, Meinke, Boreiko, and Hein]{bitterwolf2022classifiers}
Julian Bitterwolf, Alexander Meinke, Valentyn Boreiko, and Matthias Hein.
\newblock Classifiers should do well even on their worst classes.
\newblock In \emph{ICML 2022 Shift Happens Workshop}, 2022.

\bibitem[Bousquet \& Elisseeff(2000)Bousquet and Elisseeff]{bousquet2000algorithmic}
Olivier Bousquet and Andr{\'e} Elisseeff.
\newblock Algorithmic stability and generalization performance.
\newblock \emph{Advances in Neural Information Processing Systems}, 13, 2000.

\bibitem[Bu et~al.(2020)Bu, Zou, and Veeravalli]{bu2020tightening}
Yuheng Bu, Shaofeng Zou, and Venugopal~V Veeravalli.
\newblock Tightening mutual information-based bounds on generalization error.
\newblock \emph{IEEE Journal on Selected Areas in Information Theory}, 1\penalty0 (1):\penalty0 121--130, 2020.

\bibitem[Chatterjee(2018)]{chatterjee2018learning}
Satrajit Chatterjee.
\newblock Learning and memorization.
\newblock In \emph{International conference on machine learning}, pp.\  755--763. PMLR, 2018.

\bibitem[Chen et~al.(2020)Chen, He, and Su]{chen2020label}
Shuxiao Chen, Hangfeng He, and Weijie Su.
\newblock Label-aware neural tangent kernel: Toward better generalization and local elasticity.
\newblock \emph{Advances in Neural Information Processing Systems}, 33:\penalty0 15847--15858, 2020.

\bibitem[Deng et~al.(2009)Deng, Dong, Socher, Li, Li, and Fei-Fei]{deng2009imagenet}
Jia Deng, Wei Dong, Richard Socher, Li-Jia Li, Kai Li, and Li~Fei-Fei.
\newblock Imagenet: A large-scale hierarchical image database.
\newblock In \emph{2009 IEEE conference on computer vision and pattern recognition}, pp.\  248--255. Ieee, 2009.

\bibitem[Deng et~al.(2021)Deng, He, and Su]{deng2021toward}
Zhun Deng, Hangfeng He, and Weijie Su.
\newblock Toward better generalization bounds with locally elastic stability.
\newblock In \emph{International Conference on Machine Learning}, pp.\  2590--2600. PMLR, 2021.

\bibitem[Gallager(1968)]{gallager1968information}
Robert~G Gallager.
\newblock \emph{Information theory and reliable communication}, volume 588.
\newblock Springer, 1968.

\bibitem[Goodfellow et~al.(2016)Goodfellow, Bengio, Courville, and Bengio]{goodfellow2016deep}
Ian Goodfellow, Yoshua Bengio, Aaron Courville, and Yoshua Bengio.
\newblock \emph{Deep learning}, volume~1.
\newblock MIT Press, 2016.

\bibitem[Hardt et~al.(2016)Hardt, Recht, and Singer]{hardt2016train}
Moritz Hardt, Ben Recht, and Yoram Singer.
\newblock Train faster, generalize better: Stability of stochastic gradient descent.
\newblock In \emph{International conference on machine learning}, pp.\  1225--1234. PMLR, 2016.

\bibitem[Harutyunyan et~al.(2021)Harutyunyan, Raginsky, Ver~Steeg, and Galstyan]{harutyunyan2021information}
Hrayr Harutyunyan, Maxim Raginsky, Greg Ver~Steeg, and Aram Galstyan.
\newblock Information-theoretic generalization bounds for black-box learning algorithms.
\newblock \emph{Advances in Neural Information Processing Systems}, 34:\penalty0 24670--24682, 2021.

\bibitem[Harvey et~al.(2017)Harvey, Liaw, and Mehrabian]{harvey2017nearly}
Nick Harvey, Christopher Liaw, and Abbas Mehrabian.
\newblock Nearly-tight vc-dimension bounds for piecewise linear neural networks.
\newblock In \emph{Conference on learning theory}, pp.\  1064--1068. PMLR, 2017.

\bibitem[He \& Tao(2020)He and Tao]{he2020recent}
Fengxiang He and Dacheng Tao.
\newblock Recent advances in deep learning theory.
\newblock \emph{arXiv preprint arXiv:2012.10931}, 2020.

\bibitem[He \& Su(2020)He and Su]{he2019local}
Hangfeng He and Weijie~J Su.
\newblock The local elasticity of neural networks.
\newblock In \emph{International Conference on Learning Representations}, 2020.

\bibitem[He et~al.(2016)He, Zhang, Ren, and Sun]{he2016deep}
Kaiming He, Xiangyu Zhang, Shaoqing Ren, and Jian Sun.
\newblock Deep residual learning for image recognition.
\newblock In \emph{Proceedings of the IEEE conference on computer vision and pattern recognition}, pp.\  770--778, 2016.

\bibitem[Hellstr{\"o}m \& Durisi(2022)Hellstr{\"o}m and Durisi]{hellstrom2022new}
Fredrik Hellstr{\"o}m and Giuseppe Durisi.
\newblock A new family of generalization bounds using samplewise evaluated cmi.
\newblock \emph{Advances in Neural Information Processing Systems}, 35:\penalty0 10108--10121, 2022.

\bibitem[Hoeffding(1994)]{hoeffding1994probability}
Wassily Hoeffding.
\newblock Probability inequalities for sums of bounded random variables.
\newblock \emph{The collected works of Wassily Hoeffding}, pp.\  409--426, 1994.

\bibitem[Holstein et~al.(2019)Holstein, Wortman~Vaughan, Daum{\'e}~III, Dudik, and Wallach]{holstein2019improving}
Kenneth Holstein, Jennifer Wortman~Vaughan, Hal Daum{\'e}~III, Miro Dudik, and Hanna Wallach.
\newblock Improving fairness in machine learning systems: What do industry practitioners need?
\newblock In \emph{Proceedings of the 2019 CHI conference on human factors in computing systems}, pp.\  1--16, 2019.

\bibitem[Kawaguchi et~al.(2017)Kawaguchi, Kaelbling, and Bengio]{kawaguchi2017generalization}
Kenji Kawaguchi, Leslie~Pack Kaelbling, and Yoshua Bengio.
\newblock Generalization in deep learning.
\newblock \emph{arXiv preprint arXiv:1710.05468}, 1\penalty0 (8), 2017.

\bibitem[Kawaguchi et~al.(2022)Kawaguchi, Deng, Luh, and Huang]{kawaguchi2022robustness}
Kenji Kawaguchi, Zhun Deng, Kyle Luh, and Jiaoyang Huang.
\newblock Robustness implies generalization via data-dependent generalization bounds.
\newblock In \emph{International Conference on Machine Learning}, pp.\  10866--10894. PMLR, 2022.

\bibitem[Kawaguchi et~al.(2023)Kawaguchi, Deng, Ji, and Huang]{kawaguchi2023does}
Kenji Kawaguchi, Zhun Deng, Xu~Ji, and Jiaoyang Huang.
\newblock How does information bottleneck help deep learning?
\newblock \emph{arXiv preprint arXiv:2305.18887}, 2023.

\bibitem[Kirichenko et~al.(2023)Kirichenko, Balestriero, Ibrahim, Vedantam, Firooz, and Wilson]{kirichenko2023understanding}
Polina Kirichenko, Randall Balestriero, Mark Ibrahim, Shanmukha~Ramakrishna Vedantam, Hamed Firooz, and Andrew~Gordon Wilson.
\newblock Understanding the class-specific effects of data augmentations.
\newblock In \emph{ICLR 2023 Workshop on Pitfalls of limited data and computation for Trustworthy ML}, 2023.

\bibitem[Krizhevsky et~al.(2009)Krizhevsky, Hinton, et~al.]{krizhevsky2009learning}
Alex Krizhevsky, Geoffrey Hinton, et~al.
\newblock Learning multiple layers of features from tiny images.
\newblock 2009.

\bibitem[Lee et~al.(2023)Lee, Lee, and Kim]{lee2023dropmix}
Haeil Lee, Hansang Lee, and Junmo Kim.
\newblock Dropmix: Reducing class dependency in mixed sample data augmentation.
\newblock \emph{arXiv preprint arXiv:2307.09136}, 2023.

\bibitem[Mehrabi et~al.(2021)Mehrabi, Morstatter, Saxena, Lerman, and Galstyan]{mehrabi2021survey}
Ninareh Mehrabi, Fred Morstatter, Nripsuta Saxena, Kristina Lerman, and Aram Galstyan.
\newblock A survey on bias and fairness in machine learning.
\newblock \emph{ACM computing surveys (CSUR)}, 54\penalty0 (6):\penalty0 1--35, 2021.

\bibitem[Modak et~al.(2021)Modak, Asnani, and Prabhakaran]{modak2021renyi}
Eeshan Modak, Himanshu Asnani, and Vinod~M Prabhakaran.
\newblock R{\'e}nyi divergence based bounds on generalization error.
\newblock In \emph{2021 IEEE Information Theory Workshop (ITW)}, pp.\  1--6. IEEE, 2021.

\bibitem[Neu et~al.(2021)Neu, Dziugaite, Haghifam, and Roy]{neu2021information}
Gergely Neu, Gintare~Karolina Dziugaite, Mahdi Haghifam, and Daniel~M Roy.
\newblock Information-theoretic generalization bounds for stochastic gradient descent.
\newblock In \emph{Conference on Learning Theory}, pp.\  3526--3545. PMLR, 2021.

\bibitem[Neyshabur et~al.(2017)Neyshabur, Bhojanapalli, McAllester, and Srebro]{neyshabur2017exploring}
Behnam Neyshabur, Srinadh Bhojanapalli, David McAllester, and Nati Srebro.
\newblock Exploring generalization in deep learning.
\newblock \emph{Advances in neural information processing systems}, 30, 2017.

\bibitem[Rajkomar et~al.(2018)Rajkomar, Hardt, Howell, Corrado, and Chin]{rajkomar2018ensuring}
Alvin Rajkomar, Michaela Hardt, Michael~D Howell, Greg Corrado, and Marshall~H Chin.
\newblock Ensuring fairness in machine learning to advance health equity.
\newblock \emph{Annals of internal medicine}, 169\penalty0 (12):\penalty0 866--872, 2018.

\bibitem[Roberts et~al.(2022)Roberts, Yaida, and Hanin]{roberts_yaida_hanin_2022}
Daniel~A. Roberts, Sho Yaida, and Boris Hanin.
\newblock \emph{Frontmatter}, pp.\  i--iv.
\newblock Cambridge University Press, 2022.

\bibitem[Rodr{\'\i}guez~G{\'a}lvez et~al.(2021)Rodr{\'\i}guez~G{\'a}lvez, Bassi, Thobaben, and Skoglund]{rodriguez2021tighter}
Borja Rodr{\'\i}guez~G{\'a}lvez, Germ{\'a}n Bassi, Ragnar Thobaben, and Mikael Skoglund.
\newblock Tighter expected generalization error bounds via wasserstein distance.
\newblock \emph{Advances in Neural Information Processing Systems}, 34:\penalty0 19109--19121, 2021.

\bibitem[Saxe et~al.(2019)Saxe, Bansal, Dapello, Advani, Kolchinsky, Tracey, and Cox]{saxe2019information}
Andrew~M Saxe, Yamini Bansal, Joel Dapello, Madhu Advani, Artemy Kolchinsky, Brendan~D Tracey, and David~D Cox.
\newblock On the information bottleneck theory of deep learning.
\newblock \emph{Journal of Statistical Mechanics: Theory and Experiment}, 2019\penalty0 (12):\penalty0 124020, 2019.

\bibitem[Shannon(1948)]{shannon1948mathematical}
Claude~Elwood Shannon.
\newblock A mathematical theory of communication.
\newblock \emph{The Bell system technical journal}, 27\penalty0 (3):\penalty0 379--423, 1948.

\bibitem[Shui et~al.(2020)Shui, Chen, Wen, Zhou, Gagn{\'e}, and Wang]{shui2020beyond}
Changjian Shui, Qi~Chen, Jun Wen, Fan Zhou, Christian Gagn{\'e}, and Boyu Wang.
\newblock Beyond h-divergence: Domain adaptation theory with jensen-shannon divergence.
\newblock \emph{arXiv preprint arXiv:2007.15567}, 6, 2020.

\bibitem[Sontag et~al.(1998)]{sontag1998vc}
Eduardo~D Sontag et~al.
\newblock Vc dimension of neural networks.
\newblock \emph{NATO ASI Series F Computer and Systems Sciences}, 168:\penalty0 69--96, 1998.

\bibitem[Steinke \& Zakynthinou(2020)Steinke and Zakynthinou]{steinke2020reasoning}
Thomas Steinke and Lydia Zakynthinou.
\newblock Reasoning about generalization via conditional mutual information.
\newblock In \emph{Conference on Learning Theory}, pp.\  3437--3452. PMLR, 2020.

\bibitem[Tishby \& Zaslavsky(2015)Tishby and Zaslavsky]{tishby2015deep}
Naftali Tishby and Noga Zaslavsky.
\newblock Deep learning and the information bottleneck principle.
\newblock In \emph{2015 ieee information theory workshop (itw)}, pp.\  1--5. IEEE, 2015.

\bibitem[Tishby et~al.(2000)Tishby, Pereira, and Bialek]{tishby2000information}
Naftali Tishby, Fernando~C Pereira, and William Bialek.
\newblock The information bottleneck method.
\newblock \emph{arXiv preprint physics/0004057}, 2000.

\bibitem[Wang et~al.(2023)Wang, Gao, and Calmon]{wang2023generalization}
Hao Wang, Rui Gao, and Flavio~P Calmon.
\newblock Generalization bounds for noisy iterative algorithms using properties of additive noise channels.
\newblock \emph{J. Mach. Learn. Res.}, 24:\penalty0 26--1, 2023.

\bibitem[Wang et~al.(2010)Wang, Wainwright, and Ramchandran]{wang2010information}
Wei Wang, Martin~J Wainwright, and Kannan Ramchandran.
\newblock Information-theoretic bounds on model selection for gaussian markov random fields.
\newblock In \emph{2010 IEEE International Symposium on Information Theory}, pp.\  1373--1377. IEEE, 2010.

\bibitem[Wang \& Mao(2021)Wang and Mao]{wang2021generalization}
Ziqiao Wang and Yongyi Mao.
\newblock On the generalization of models trained with sgd: Information-theoretic bounds and implications.
\newblock \emph{arXiv preprint arXiv:2110.03128}, 2021.

\bibitem[Wang \& Mao(2022)Wang and Mao]{wang2022information}
Ziqiao Wang and Yongyi Mao.
\newblock Information-theoretic analysis of unsupervised domain adaptation.
\newblock \emph{arXiv preprint arXiv:2210.00706}, 2022.

\bibitem[Wang \& Mao(2023)Wang and Mao]{wang2023tighter}
Ziqiao Wang and Yongyi Mao.
\newblock Tighter information-theoretic generalization bounds from supersamples.
\newblock \emph{arXiv preprint arXiv:2302.02432}, 2023.

\bibitem[Williamson \& Menon(2019)Williamson and Menon]{williamson2019fairness}
Robert Williamson and Aditya Menon.
\newblock Fairness risk measures.
\newblock In \emph{International conference on machine learning}, pp.\  6786--6797. PMLR, 2019.

\bibitem[Wu et~al.(2020)Wu, Manton, Aickelin, and Zhu]{wu2020information}
Xuetong Wu, Jonathan~H Manton, Uwe Aickelin, and Jingge Zhu.
\newblock Information-theoretic analysis for transfer learning.
\newblock In \emph{2020 IEEE International Symposium on Information Theory (ISIT)}, pp.\  2819--2824. IEEE, 2020.

\bibitem[Xu \& Raginsky(2017)Xu and Raginsky]{xu2017information}
Aolin Xu and Maxim Raginsky.
\newblock Information-theoretic analysis of generalization capability of learning algorithms.
\newblock \emph{Advances in Neural Information Processing Systems}, 30, 2017.

\bibitem[Xu \& Mannor(2012)Xu and Mannor]{xu2012robustness}
Huan Xu and Shie Mannor.
\newblock Robustness and generalization.
\newblock \emph{Machine learning}, 86:\penalty0 391--423, 2012.

\bibitem[Zhang et~al.(2016)Zhang, Bengio, Hardt, Recht, and Vinyals]{zhang2016understanding}
Chiyuan Zhang, Samy Bengio, Moritz Hardt, Benjamin Recht, and Oriol Vinyals.
\newblock Understanding deep learning requires rethinking generalization.
\newblock In \emph{International Conference on Learning Representations}, 2016.

\bibitem[Zhang et~al.(2021)Zhang, Bengio, Hardt, Recht, and Vinyals]{zhang2021understanding}
Chiyuan Zhang, Samy Bengio, Moritz Hardt, Benjamin Recht, and Oriol Vinyals.
\newblock Understanding deep learning (still) requires rethinking generalization.
\newblock \emph{Communications of the ACM}, 64\penalty0 (3):\penalty0 107--115, 2021.

\bibitem[Zhou et~al.(2022)Zhou, Tian, and Liu]{zhou2022individually}
Ruida Zhou, Chao Tian, and Tie Liu.
\newblock Individually conditional individual mutual information bound on generalization error.
\newblock \emph{IEEE Transactions on Information Theory}, 68\penalty0 (5):\penalty0 3304--3316, 2022.

\bibitem[Zhou et~al.(2023)Zhou, Tian, and Liu]{zhou2023exactly}
Ruida Zhou, Chao Tian, and Tie Liu.
\newblock Exactly tight information-theoretic generalization error bound for the quadratic gaussian problem.
\newblock \emph{arXiv preprint arXiv:2305.00876}, 2023.

\end{thebibliography}
\bibliographystyle{tmlr}

\appendix
\section{Appendix:  Related work}
\label{related}
\textbf{Information-theoretic generalization error bounds:} Information-theoretic bounds have attracted a lot of attention recently to characterize the generalization of learning algorithms~\citep{neu2021information,wang2010information,aminian2021exact,wu2020information,wang2022information,modak2021renyi,wang2021generalization,shui2020beyond,wang2023generalization,alabdulmohsin2020towards}. In the supervised learning context, several standard generalization error bounds have been proposed to based on different tools, e.g., KL divergence \citep{zhou2023exactly}, Wasserstein distance~\citep{rodriguez2021tighter}, mutual information between the samples and the weights~\citep{xu2017information,bu2020tightening}. Recently, it was shown that tighter generalization bounds could be obtained based on the conditional mutual information (CMI) setting~\citep{steinke2020reasoning,zhou2022individually}. Based on this framework, \cite{harutyunyan2021information} derived $f$-CMI bounds based on the model output. In~\cite{hellstrom2022new}, tighter bounds have been obtained based on the CMI of the loss function, which is further tightened by~\cite{wang2023tighter} using the $\delta L$ CMI. 

\textbf{Class-dependent analysis: }
Incorporating label information in generalization analysis is not entirely new~\citep{he2019local,chen2020label,deng2021toward}. For example, in \cite{he2019local}, the questions ``When and how does the update of weights of neural networks using induced gradient at an example impact the prediction at another example?'' have been extensively studied, and it was observed that the impact is significant if the two samples are from the same class. In \cite{NEURIPS2022_f73c0453,kirichenko2023understanding,bitterwolf2022classifiers,lee2023dropmix}, it has been showed that while standard data augmentation techniques~\citep{goodfellow2016deep} help improve overall performance, it yields lower performance on minority classes. 
From a theoretical perspective,  \citep{deng2021toward} noticed that in uniform stability context, the sensitivity of neural networks depends highly on the label information and thus proposed the concept of ``\textit{Locally Elastic Stability}'' to derive tighter algorithmic stability generalization bounds. 
In~\cite{tishby2000information}, an information bottleneck principle is proposed which states that an optimal feature map simultaneously minimizes its mutual information with the feature distribution and maximizes its mutual information with the label distribution, thus incorporating the class information.  In \cite{tishby2015deep,saxe2019information,kawaguchi2023does}, this principle was used to explain the generalization of neural networks.

\section{Appendix: Proofs of the Theorems in Section~\ref{section_classgenerror}}
This appendix includes the missing proofs of the results presented in the main text in Section~\ref{section_classgenerror}.

\begin{lemma} \label{boundedsubgaussian}
    Let $\rmX$ be a bounded random variable, i.e., $\rmX \in [a,b]$ almost surely. If $\E[\rmX]=0$, then $\rmX$ is $(b-a)$-subgaussian and we have:
    \begin{equation}
        \E[e^{\lambda \mX}] \leq e^{\frac{\lambda^2 (b-a)^2}{8}}, \: \: \forall \lambda \in \sR.
    \end{equation}
\end{lemma}

Next, we need the following Lemma  \ref{CMI-lemma} as a main tool to prove Theorem~\ref{CMI_class_bound},~\ref{fCMI_class_bound}, and~\ref{LCMI_class_bound} in Section~\ref{section_cmisettings} in the CMI setting.
\begin{lemma} \label{CMI-lemma}
Consider the CMI setting. Let $\rmV \in \mathcal{V}$ be a random variable, possibly depending on $\mW$. For any function g that can be written as $g(\rmV,\rmU_i,z_{[2n]}) = \mathds{1}_{\{y^{\rmU_i}=y\}} h(\rmV, z^{\rmU_i}_i) - \mathds{1}_{\{y^{-\rmU_i}=y\}} h(\mV,  z^{-\rmU_i}_i)$ such that  $h \in [0,1]$ is a bounded function, we have
\begin{equation}
   \E_{\rmV; \rmU_i|\rmZ_{[2n]}= z_{[2n]}}[g(\rmV,\rmU_i,z_{[2n]})] \leq \sqrt{2 \max(\mathds{1}_{\{y_i^{-}=y\}},\mathds{1}_{\{y_i^{+}=y\}}) I_{z_{[2n]} }(\rmV; \rmU_i )}. 
 \end{equation}
\end{lemma}
\begin{proof}
Let $(\overline{\rmV}, \overline{\mU_i})$ be an independent copy of $(\rmV ,\rmU_i)$. The disintegrated mutual information $I_{z_{[2n]}} (\rmV; \rmU_i )$ is equal to:
\begin{equation}    
    I_{z_{[2n]}} (\rmU_i;\rmV  )  = 
    D\big(P_{\rmV, \rmU_i|\rmZ_{[2n]}= z_{[2n]} }\|P_{\rmV|\rmZ_{[2n]}= z_{[2n]}} P_{\rmU_i}\big), 
\end{equation}
Thus, by the Donsker–Varadhan variational representation of KL
divergence,  $\forall \lambda \in \sR$ and for every function $g$, we have
\begin{equation} \label{classvariational}
       I_{z_{[2n]} } (\rmV; \rmU_i )  \geq   \lambda \E_{\rmV, \rmU_i|\rmZ_{[2n]}= z_{[2n]}}[g(\rmV,\rmU_i,z_{[2n]})]  
   -     \log \E_{\overline{\rmV},\overline{\rmU}_i|\rmZ_{[2n]}= z_{[2n]}} [e^{ \lambda g(\overline{\rmV},\overline{\rmU}_i,z_{[2n]})}]. 
\end{equation}
Next, let $g(\rmV,\rmU_i,z_{[2n]}) = \mathds{1}_{\{y^{\rmU_i}=y\}} h(\rmV, z^{\rmU_i}_i) - \mathds{1}_{\{y^{-\rmU_i}=y\}} h(\mV,  z^{-\rmU_i}_i) $. It is easy to see that $ g(\overline{\rmV},\overline{\rmU}_i,z_{[2n]})$ can be rewritten as follows:
\begin{equation}
     g(\overline{\rmV},\overline{\rmU}_i,z_{[2n]}) = \overline{\rmU}_i (\mathds{1}_{\{y_i^{-}=y\}} h(\overline{\rmV},z^{-}_i) - \mathds{1}_{\{y_i^{+}=y\}} h(\overline{\rmV},  z^{+}_i)).
\end{equation}
Thus, we have
\begin{equation}
   \log \E_{\overline{\rmV},\overline{\rmU}_i|\rmZ_{[2n]}= z_{[2n]}} [e^{ \lambda g(\overline{\rmV},\overline{\rmU}_i,z_{[2n]})}] =\log \E_{\overline{\rmV},\overline{\rmU}_i|\rmZ_{[2n]}= z_{[2n]}} [e^{ \lambda \overline{\rmU}_i (\mathds{1}_{\{y_i^{-}=y\}} h(\overline{\rmV},z^{-}_i) - \mathds{1}_{\{y_i^{+}=y\}} h(\overline{\rmV},  z^{+}_i)) }] .
\end{equation}
Note that $ \E_{\overline{\rmU}_i } [\overline{\rmU}_i (\mathds{1}_{\{y_i^{-}=y\}} h(\overline{\rmV}, z^{-}_i) - \mathds{1}_{\{y_i^{+}=y\}} h(\overline{\rmV},  z^{+}_i))] =0$ and $\overline{\rmU}_i \in \{-1,+1\}$. Thus, using Hoeffding's Lemma, we have
\begin{equation}
     \log \E_{\overline{\rmV},\overline{\rmU}_i|\rmZ_{[2n]}= z_{[2n]}} [e^{ \lambda g(\overline{\rmV},\overline{\rmU}_i,z_{[2n]})}] \leq   \log \E_{\overline{\rmV}|\rmZ_{[2n]}= z_{[2n]}} [e^{ \frac{\lambda^2}{2} \big(\mathds{1}_{\{y_i^{-}=y\}} h(\overline{\rmV}, z^{-}_i) - \mathds{1}_{\{y_i^{+}=y\}} h(\overline{\rmV},  z^{+}_i)\big)^2 }]  .
\end{equation}
Next, as $h \in [0,1]$, $ \Big|\mathds{1}_{\{y_i^{-}=y\}} h(\overline{\rmV}, z^{-}_i) - \mathds{1}_{\{y_i^{+}=y\}} h(\overline{\rmV},  z^{+}_i) \Big| \leq \max(\mathds{1}_{\{y_i^{-}=y\}},\mathds{1}_{\{y_i^{+}=y\}})$.
Thus, 
\begin{equation}
     \log \E_{\overline{\rmV},\overline{\rmU}_i|\rmZ_{[2n]}= z_{[2n]}} [e^{ \lambda g(\overline{\rmV},\overline{\rmU}_i,z_{[2n]})}] \leq  \frac{\lambda^2}{2} \max(\mathds{1}_{\{y_i^{-}=y\}}, \mathds{1}_{\{y_i^{+}=y\}})^2 =  \frac{\lambda^2}{2} \max(\mathds{1}_{\{y_i^{-}=y\}}, \mathds{1}_{\{y_i^{+}=y\}}).
\end{equation}

Replacing in \eqref{classvariational}, we have
\begin{multline}
     I_{z_{[2n]} } (\rmV; \rmU_i )  \geq   \lambda \E_{\rmV, \rmU_i|\rmZ_{[2n]}= z_{[2n]}}[\mathds{1}_{y^{\rmU_i}=y} h(\rmV, z^{\rmU_i}_i) - \mathds{1}_{\{y^{-\rmU_i}=y\}} h(\mW,  z^{-\rmU_i}_i)]  \\
   -      \frac{\lambda^2}{2} \max(\mathds{1}_{\{y_i^{-}=y\}}, \mathds{1}_{\{y_i^{+}=y\}}). 
\end{multline}

So, $\forall \lambda \in \sR$
\begin{equation} \label{parabolaeq}
     \frac{\lambda^2}{2} \max(\mathds{1}_{\{y_i^{-}=y\}}, \mathds{1}_{\{y_i^{+}=y\}})  - \lambda   \E_{\rmV; \rmU_i|\rmZ_{[2n]}= z_{[2n]}}[g(\rmV,\rmU_i,z_{[2n]})]  +  I_{z_{[2n]}} (\rmV; \rmU_i ) \geq  0.
\end{equation}
The \eqref{parabolaeq} is a non-negative parabola with respect to $\lambda$. Thus its discriminant must be non-positive. This implies
\begin{equation}
   \E_{\rmV; \rmU_i|\rmZ_{[2n]}= z_{[2n]}}[g(\rmV,\rmU_i,z_{[2n]})] \leq \sqrt{2 \max(\mathds{1}_{\{y_i^{-}=y\}},\mathds{1}_{\{y_i^{+}=y\}}) I_{z_{[2n]} }(\rmV; \rmU_i )}.
\end{equation}

\end{proof}

\subsection{ Proof of Theorem~\ref{MI_class_bound}} \label{MI_class_bound_proof}
\textbf{Theorem~\ref{MI_class_bound}} (restated)  For $y \in \gY$, assume Assumption~\ref{fixed_joint} holds and the loss $\ell(\overline{\rmW},\overline{\rmX},y)$ is $\sigma_y$ sub-gaussian under $P_{\overline{\rmW}} \otimes P_{\overline{\rmX}|\overline{\rmY}=y }$, then  the  class-generalization error of class $y$ in Definition~\ref{class_gen} can be bounded as:
\begin{equation}
   |  \overline{\mathrm{gen}_y}(P_{\rmX,\rmY}, P_{\rmW|\rmS}) | \leq \sqrt{2\sigma_{y}^2D(P_{\rmW,\rmX|\rmY=y} ||P_{\rmW}\otimes P_{\rmX|\rmY=y}) }.
\end{equation}  
\begin{proof}
From lemma~\ref{class_gen_lemma}, we have 
\begin{equation}
\overline{\mathrm{gen}_y}(P_{\rmX,\rmY}, P_{\rmW|\rmS}) = \E_{P_{\overline{\rmW}}\otimes P_{\overline{\rmX}|\rmY=y}} [\ell(\overline{\rmW},\overline{\rmX},y)] - \E_{P_{\rmW,\rmX|\rmY=y}}[ \ell(\rmW,\rmX,y)].
\end{equation}
Using the Donsker–Varadhan variational representation of the relative entropy,  we have
\begin{multline} \label{main_eq}
     D(P_{\rmW,\rmX|\rmY=y}||P_{\rmW}\otimes P_{\rmX|\rmY=y}) \geq \E_{P_{\rmW,\rmX|\rmY=y}}[\lambda \ell(\rmW,\rmX,y)] \\ - \log \E_{P_{\overline{\rmW}}\otimes P_{\overline{\rmX}|\overline{\rmY}=y}}[e^{\lambda \ell(\overline{\rmW},\overline{\rmX},y)} ], \forall \lambda \in \sR.
   \end{multline} 
On the other hand, we have:
\begin{align*}   
    &\log  \E_{P_{\overline{\rmW}}\otimes P_{\overline{\rmX}|\overline{\rmY}=y}} \Big[ e^{\lambda \ell(\overline{\rmW},\overline{\rmX},y) - \lambda\E [\ell(\overline{\rmW},\overline{\rmX},y)]  }\Big]\\
    &=   \log  \E_{P_{\overline{\rmW}}\otimes P_{\overline{\rmX}|\overline{\rmY}=y}} \Big[ e^{\lambda \ell(\overline{\rmW},\overline{\rmX},y) }  e^{- \lambda  \E [\ell(\overline{\rmW},\overline{\rmX},y)]  \big)}\Big]  \\
    &= \log \E_{P_{\overline{\rmW}}\otimes P_{\overline{\rmX}|\overline{\rmY}=y}}[e^{\lambda \ell(\overline{\rmW},\overline{\rmX},y)} ] - \lambda \E_{P_{\overline{\rmW}}\otimes P_{\overline{\rmX}|\overline{\rmY}=y}} [\ell(\overline{\rmW},\overline{\rmX},y)].
\end{align*}
Using the sub-gaussian assumption, we have
\begin{equation}
    \log \E_{P_{\overline{\rmW}}\otimes P_{\overline{\rmX}|\overline{\rmY}=y}}[e^{\lambda \ell(\overline{\rmW},\overline{\rmX},y)} ] \leq \lambda \E_{P_{\overline{\rmW}}\otimes P_{\overline{\rmX}|\overline{\rmY}=y}} (\ell(\overline{\rmW},\overline{\rmX},y)) + \frac{\lambda^2 \sigma_y^2}{2}.
\end{equation}
By replacing in \eqref{main_eq}, we have 
\begin{multline}   D(P_{\rmW,\rmX|\rmY=y}||P_{\rmW}\otimes P_{\rmX|\rmY=y}) \geq \lambda \big(\E_{P_{\rmW,\rmX|\rmY=y}}[ \ell(\rmW,\rmX,y)] -  \\ \E_{P_{\overline{\rmW}}\otimes P_{\overline{\rmX}|\overline{\rmY}=y}} [\ell(\overline{\rmW},\overline{\rmX},y)] \big)  -  \frac{\lambda^2 \sigma_y^2}{2}.
\end{multline}

Thus, 
\begin{multline} \label{eq11}   D(P_{\rmW,\rmX|\rmY=y}||P_{\rmW}\otimes P_{\rmX|\rmY=y}) -  \lambda (\E_{P_{\rmW,\rmX|\rmY=y}}[ \ell(\rmW,\rmX,y)] -  \E_{P_{\overline{\rmW}}\otimes P_{\overline{\rmX}|\overline{\rmY}=y}} [\ell(\overline{\rmW},\overline{\rmX},y)] )   \\ + \lambda^2 \sigma_{y}^2 \geq 0 , \forall \lambda  \in \sR.
\end{multline}
Equation \eqref{eq11} is a non-negative parabola with respect to $\lambda$, which implies its discriminant must be non-positive. Thus,
\begin{equation}
   |  \E_{P_{\rmW,\rmX|\rmY=y}}[ \ell(\rmW,\rmX,y)] -  \E_{P_{\overline{\rmW}}\otimes P_{\overline{\rmX}|\overline{\rmY}=y}} [\ell(\overline{\rmW},\overline{\rmX},y)] | \leq \sqrt{2\sigma_{y}^2D(P_{\rmW,\rmX|\rmY=y}||P_{\rmW}\otimes P_{\rmX|\rmY=y}) }.
\end{equation}
   This completes the proof.
\end{proof}

\subsection{Proof of Theorem~\ref{CMI_class_bound}} \label{CMI_class_bound_proof}

\textbf{Theorem~\ref{CMI_class_bound}} (restated)
Assume that the loss $\ell(w,x,y) \in [0,1]$ is bounded, then the class-generalization error for class $y$ in Definition~\ref{CMI_class_gen} can be bounded as
\begin{equation}
| \overline{\mathrm{gen}_y}(P_{\rmX,\rmY}, P_{\rmW|\rmS}) |\leq  \E_{\rmZ_{[2n]}}   
\Big[ \frac{1}{n^y}  \sum_{i=1}^n   \sqrt{2 \max(\mathds{1}_{\{\rmY_i^{-}=y\}}, \mathds{1}_{\{\rmY_i^{+}=y\}}) I_{\rmZ_{[2n]}} (\rmW; \rmU_i )} \Big]. 
 \end{equation} 
\begin{proof}
Using  Lemma~\ref{CMI-lemma} with $\rmV=\rmW$ and $h(\rmV,z)=\ell(\rmW,z)$ in, we have
\begin{equation}
   \E_{\rmW; \rmU_i|\rmZ_{[2n]}= z_{[2n]}}[g(\rmW,\rmU_i,z_{[2n]})] \leq \sqrt{2 \max(\mathds{1}_{\{y_i^{-}=y\}},\mathds{1}_{\{y_i^{+}=y\}}) I_{z_{[2n]} }(\rmW; \rmU_i )},
\end{equation}
where $g(\rmW,\rmU_i,z_{[2n]})= \mathds{1}_{\{y^{\rmU_i}=y\}} \ell(\rmW, z^{\rmU_i}_i) - \mathds{1}_{\{y^{-\rmU_i}=y\}} \ell(\mW,  z^{-\rmU_i}_i) $. Thus, by summing over the different terms in Definition~\ref{CMI_class_gen} and taking expectation over $\rmZ_{[2n]}$, 
\begin{equation}
| \overline{\mathrm{gen}_y}(P_{\rmX,\rmY}, P_{\rmW|\rmS}) | \leq  \E_{\rmZ_{[2n]}}   
\Big[ \frac{1}{n^y}  \sum_{i=1}^n   \sqrt{2 \max(\mathds{1}_{\{\rmY_i^{-}=y\}}, \mathds{1}_{\{\rmY_i^{+}=y\}}) I_{\rmZ_{[2n]}} (\rmW; \rmU_i )} \Big]. 
\end{equation}
\end{proof}

\subsection{Proof of Theorem~\ref{fCMI_class_bound}}

\textbf{Theorem~\ref{fCMI_class_bound}} (restated) Assume that the loss $\ell(\rmW,\rmX,y) \in [0,1]$. The  class-generalization error of class $y$, as defined in~\ref{CMI_class_gen}, can be bounded as follows:
\begin{equation}
| \overline{\mathrm{gen}_y}(P_{\rmX,\rmY}, P_{\rmW|\rmS}) | \leq \E_{\rmZ_{[2n]}}
\Big[    \frac{1}{n^y}  \sum_{i=1}^n   \sqrt{2 \max(\mathds{1}_{\{\rmY_i^{-}=y\}}, \mathds{1}_{\{\rmY_i^{+}=y\}}) I_{\rmZ_{[2n]}} (f_{\rmW}(\rmX^\pm_i); \rmU_i )} \Big]. 
 \end{equation}
 
\begin{proof}
Similar to the proof of Theorem ~\ref{CMI_class_bound}.   
Using  Lemma~\ref{CMI-lemma} with $\rmV=f_{\rmW}(x^{\pm}_i)$ and $h(\rmV,z_i)=\ell(f_{\rmW}(x_i),y_i)$ in, we have
\begin{equation}
   \E_{f_{\rmW}(x^{\pm}_i); \rmU_i|\rmZ_{[2n]}= z_{[2n]}}[g(f_{\rmW}(\rmX_i),\rmU_i,z_{[2n]})] \leq  \sqrt{2 \max(\mathds{1}_{\{y_i^{-}=y\}},\mathds{1}_{\{y_i^{+}=y\}}) I_{z_{[2n]} } (f_{\rmW}(x^{\pm}_i); \rmU_i )}.
\end{equation}
Thus taking expectation with respect to $\rmZ_{[2n]}$ yields the desired result
\begin{equation}
| \overline{\mathrm{gen}_y}  (P_{\rmX,\rmY}, P_{\rmW|\rmS}) | \leq  \E_{\rmZ_{[2n]}} \Big[  \frac{1}{n^y}  \sum_{i=1}^n   \sqrt{2 \max(\mathds{1}_{\{\rmY_i^{-}=y\}},\mathds{1}_{\{\rmY_i^{+}=y\}})  I_{\rmZ_{[2n]} } (f_{\rmW}(\rmX^{\pm}_i); \rmU_i )} \Big].
\end{equation}
\end{proof}

\subsection{Extra bound of class-generalization error using the Loss pair $\rmL^{\pm}_i$:}

\begin{theorem}
\label{LCMI_class_bound} 
 (class-e-CMI)
Assume that the loss $\ell(\hat{y},y) \in [0,1]$.  The  class-generalization error of class $y$, as defined in~\ref{CMI_class_gen}, can be bounded as follows:
    \begin{equation}
| \overline{\mathrm{gen}_y}  (P_{\rmX,\rmY}, P_{\rmW|\rmS}) | \leq  \E_{\rmZ_{[2n]}} \Big[  \frac{1}{n^y} \sum_{i=1}^n   \sqrt{2 \max(\mathds{1}_{\{\rmY_i^{-}=y\}},\mathds{1}_{\{\rmY_i^{+}=y\}})  I_{\rmZ_{[2n]} } (\rmL^{\pm}_i; \rmU_i )} \Big].
\end{equation}
\end{theorem}

\begin{proof}
Similar to the proof of Theorems ~\ref{CMI_class_bound} and~\ref{fCMI_class_bound}. 
Using  Lemma~\ref{CMI-lemma} with $\rmV=\rmL^{\pm}_i$ and $h(\rmV,z_i)=\rmL_i$ in, we have

\begin{equation}
   \E_{\rmL^{\pm}_i; \rmU_i|\rmZ_{[2n]}= z_{[2n]}}[g(f_{\rmW}(\rmX_i),\rmU_i,z_{[2n]})] \leq  \sqrt{2 \max(\mathds{1}_{\{y_i^{-}=y\}},\mathds{1}_{\{y_i^{+}=y\}}) I_{z_{[2n]} } (\rmL^{\pm}_i; \rmU_i )}.
\end{equation}
Thus taking expectation with respect to $\rmZ_{[2n]}$ yields the desired result
\begin{equation}
| \overline{\mathrm{gen}_y}  (P_{\rmX,\rmY}, P_{\rmW|\rmS}) | \leq  \E_{\rmZ_{[2n]}} \Big[  \frac{1}{n^y}  \sum_{i=1}^n   \sqrt{2 \max(\mathds{1}_{\{\rmY_i^{-}=y\}},\mathds{1}_{\{\rmY_i^{+}=y\}})  I_{\rmZ_{[2n]} } (\rmL^{\pm}_i; \rmU_i )} \Big].
\end{equation} 
\end{proof}

\subsection{Proof of Theorem~\ref{deltaLCMI_class_bound}}

\textbf{Theorem~\ref{deltaLCMI_class_bound}} (restated)
 ($\Delta_y L$-CMI)
$\Delta_y \rmL_i \triangleq \mathds{1}_{\{y_i^{-}=y\}} \ell(f_{\rmW}(\rmX_i)^{-}, y^{-}_i) - \mathds{1}_{\{y_i^{+}=y\}} \ell(f_{\rmW}(\rmX_i)^+,  y^{+}_i) $.  Assume that the loss $\ell(\hat{y},y) \in [0,1]$.  The  class-generalization error of class $y$, as defined in~\ref{CMI_class_gen}, can be bounded as follows:
    \begin{equation}
| \overline{\mathrm{gen}_y}  (P_{\rmX,\rmY}, P_{\rmW|\rmS}) | \leq  \E_{\rmZ_{[2n]}} \Big[  \frac{1}{n^y}  \sum_{i=1}^n   \sqrt{2 I_{\rmZ_{[2n]}} (\Delta_y \rmL_i; \rmU_i )} \Big].
    \end{equation}
\begin{proof}
First, we notice that for a fixed realization $z_{[2n]}$, $ \mathds{1}_{\{y^{\rmU_i}=y\}} \ell(\rmW, z^{\rmU_i}_i) - \mathds{1}_{\{y^{-\rmU_i}=y\}} \ell(\mW,  z^{-\rmU_i}_i) = \rmU_i (\mathds{1}_{\{y_i^{-}=y\}} \ell(\rmW,z^{-}_i) - \mathds{1}_{\{y_i^{+}=y\}} \ell(\rmW,  z^{+}_i)) = \mU_i \Delta_y \rmL_i$. 

Next, let $(\overline{\Delta_y \rmL}_i,\overline{\rmU}_i)$ be an independent copy of $(\Delta_y \rmL_i; \rmU_i)$. Using the Donsker–Varadhan variational representation of KL
divergence,  we have  $\forall \lambda \in \sR$ and for every function $g$
\begin{multline} \label{classvariational1delta}
       I_{z_{[2n]} } (\Delta_y \rmL_i; \rmU_i )  \geq   \lambda \E_{\Delta_y \rmL_i, \rmU_i|\rmZ_{[2n]}= z_{[2n]}}[g(\Delta_y \rmL_i,\rmU_i,z_{[2n]})]  \\
   -     \log \E_{ \overline{\Delta_y \rmL}_i,\overline{\rmU}_i|\rmZ_{[2n]}= z_{[2n]}} [e^{ \lambda g(\overline{\Delta_y \rmL}_i,\overline{\rmU}_i,z_{[2n]})}].
\end{multline}

Next, let $g(\Delta_y \rmL_i,\rmU_i,z_{[2n]}) = \mU_i \Delta_y \rmL_i $. Thus, we have 
\begin{equation}
   \log \E_{\overline{\Delta_y \rmL}_i,\overline{\rmU}_i|\rmZ_{[2n]}= z_{[2n]}} [e^{ \lambda   g(\overline{\Delta_y \rmL}_i,\overline{\rmU}_i,z_{[2n]})}] =
   \log \E_{\overline{\Delta_y \rmL}_i,\overline{\rmU}_i|\rmZ_{[2n]}= z_{[2n]}} [e^{ \lambda \overline{\rmU}_i \overline{\Delta_y \rmL}_i }]. 
\end{equation} 
We have $ \E_{\overline{\rmU}_i } [\overline{\rmU}_i \overline{\Delta_y \rmL}_i ] =0$ and $\overline{\rmU}_i \in \{-1,+1\}$. Thus, using Hoeffding's Lemma, we have
\begin{equation}
     \log \E_{\overline{\rmL}^{\pm}_i,\overline{\rmU}_i|\rmZ_{[2n]}= z_{[2n]}} [e^{ \lambda   g(\overline{\Delta_y \rmL}_i,\overline{\rmU}_i,z_{[2n]})}] 
     \leq \\  \log \E_{\overline{\Delta_y \rmL}_i|\rmZ_{[2n]}= z_{[2n]}}  [e^{ \frac{\lambda^2}{2} \overline{\Delta_y \rmL}_i^2 }].  
\end{equation}
Next, as $\ell \in [0,1]$, it follows that $\Delta_y \rmL_i \in [-1,1]$. Thus, $ | \overline{\Delta_y \rmL}_i | \leq 1$.
Thus, 
\begin{equation}
     \log \E_{\overline{\Delta_y \rmL}_i,\overline{\rmU}_i|\rmZ_{[2n]}= z_{[2n]}} [e^{ \lambda   g(\overline{\Delta_y \rmL}_i,\overline{\rmU}_i,z_{[2n]})}] \leq  \frac{\lambda^2}{2}.
\end{equation}

Replacing in \eqref{classvariational1delta}, we have
\begin{equation}
     I_{z_{[2n]} } (\Delta_y \rmL_i; \rmU_i )  \geq   \lambda \E_{\Delta_y \rmL_i, \rmU_i|\rmZ_{[2n]}= z_{[2n]}}[\rmU_i \Delta_y \rmL_i ] 
   -      \frac{\lambda^2}{2}. 
\end{equation}

So $\forall \lambda \in \sR$
\begin{equation} \label{parabolaeq2delta}
     \frac{\lambda^2}{2} - \lambda   \E_{\Delta_y \rmL_i; \rmU_i|\rmZ_{[2n]}= z_{[2n]}}[g(\Delta_y \rmL_i,\rmU_i,z_{[2n]})]  +  I_{z_{[2n]} } (\Delta_y \rmL_i; \rmU_i ) \geq  0.
\end{equation}
The \eqref{parabolaeq2delta} is a non-negative parabola with respect to $\lambda$. Thus, its discriminant must be non-positive. This implies
\begin{equation}
   \E_{\Delta_y \rmL_i; \rmU_i|\rmZ_{[2n]}= z_{[2n]}}[g(f_{\rmW}(\rmX_i),\rmU_i,z_{[2n]})] \leq  \sqrt{2 I_{z_{[2n]} } (\Delta_y \rmL_i; \rmU_i )}.
\end{equation}
Thus taking expectation with respect to $\rmZ_{[2n]}$ yields the desired result
\begin{equation}
| \overline{\mathrm{gen}_y}  (P_{\rmX,\rmY}, P_{\rmW|\rmS}) | \leq  \E_{\rmZ_{[2n]}} \Big[  \frac{1}{n^y}  \sum_{i=1}^n   \sqrt{2   I_{\rmZ_{[2n]} } (\Delta_y \rmL_i; \rmU_i )} \Big].
\end{equation} 

\textbf{Proof that the $\Delta_y L$-CMI is always tighter than the class-$f$-CMI bound in Theorem~\ref{fCMI_class_bound}, and the latter is always tighter than the class-CMI bound in Theorem~\ref{CMI_class_bound}:} \\

Due to the data processing inequality , we have $\rmU \to \rmW \to f_{\rmW}(\rmX^\pm_i) \to  \Delta_y \rmL_i $. It then follows directly that the class-$f$-CMI bound is always tighter than the class-CMI bound. Moreover, we can also show that for a the $\Delta_y L$-CMI is always tighter than the class-$f$-CMI bound.  For a fixed $\rmZ_{[2n]}$, we have 4 different possible cases for each term in the sum:

\begin{itemize}
    \item i) If $y_i^-  \neq y$ and  $y_i^+  \neq y$: In this case, $ \max(\mathds{1}_{\{\rmY_i^{-}=y\}}, \mathds{1}_{\{\rmY_i^{+}=y\}}) I_{\rmZ_{[2n]}} (f_{\rmW}(\rmX^\pm_i); \rmU_i ) = 0 I_{\rmZ_{[2n]}} (f_{\rmW}(\rmX^\pm_i); \rmU_i ) =0$. On the other hand, we have $\Delta_y \rmL_i= 0$ so $ I_{\rmZ_{[2n]} } (\Delta_y \rmL_i; \rmU_i ) = I_{\rmZ_{[2n]} } (0; \rmU_i )  = 0 \leq 0 =\max(\mathds{1}_{\{\rmY_i^{-}=y\}}, \mathds{1}_{\{\rmY_i^{+}=y\}}) I_{\rmZ_{[2n]}} (f_{\rmW}(\rmX^\pm_i); \rmU_i ) $ 

    \item ii),If $y_i^- = y$ and  $y_i^+ = y$:  In this case, $ \max(\mathds{1}_{\{\rmY_i^{-}=y\}}, \mathds{1}_{\{\rmY_i^{+}=y\}}) I_{\rmZ_{[2n]}} (f_{\rmW}(\rmX^\pm_i); \rmU_i ) = I_{\rmZ_{[2n]}} (f_{\rmW}(\rmX^\pm_i); \rmU_i ) $. Due to the  data processing inequality,   $ I_{\rmZ_{[2n]}} (\Delta_y \rmL_i; \rmU_i ) \le I_{\rmZ_{[2n]}} (f_{\rmW}(\rmX^\pm_i); \rmU_i )= \max(\mathds{1}_{\{\rmY_i^{-}=y\}}, \mathds{1}_{\{\rmY_i^{+}=y\}}) I_{\rmZ_{[2n]}} (f_{\rmW}(\rmX^\pm_i); \rmU_i )  $

    \item iii) if  ,If $y_i^+ \neq y$ and  $y_i^- = y$,  In this case, $ \max(\mathds{1}_{\{\rmY_i^{-}=y\}}, \mathds{1}_{\{\rmY_i^{+}=y\}}) I_{\rmZ_{[2n]}} (f_{\rmW}(\rmX^\pm_i); \rmU_i ) = I_{\rmZ_{[2n]}} (f_{\rmW}(\rmX^\pm_i); \rmU_i ) $ and $\Delta_y \rmL_i = \rmL_i^+ $. As $ \rmW \to f_{\rmW}(\rmX^\pm_i) \to  \rmL_i^+ $ is a Markov chain, using the data processing inequality, we have  $ I_{\rmZ_{[2n]}} (\rmL_i^+; \rmU_i ) \le I_{\rmZ_{[2n]}} (f_{\rmW}(\rmX^\pm_i); \rmU_i ) $  and thus   $ I_{\rmZ_{[2n]}} (\Delta_y \rmL_i; \rmU_i ) \le I_{\rmZ_{[2n]}} (f_{\rmW}(\rmX^\pm_i); \rmU_i ) $ 

    \item iv)  If  $y_i^+ = y $ and   $y_i^- \neq  y $,  this corresponds to the same as iii) by changing the + and -. 
\end{itemize}

As for all the different cases, we have  

\begin{equation}
    I_{\rmZ_{[2n]}} (\Delta_y \rmL_i; \rmU_i ) \le  \max(\mathds{1}_{\{\rmY_i^{-}=y\}}, \mathds{1}_{\{\rmY_i^{+}=y\}}) I_{\rmZ_{[2n]}} (f_{\rmW}(\rmX^\pm_i); \rmU_i )
\end{equation}

$\Delta_y L$-CMI is always tighter than the class-$f$-CMI bound.

\end{proof}

\section{Additional Empirical results}
\subsection{Experiment Setup} \label{expi_setup}
Here, we fully describe the experimental setup used in the main body of the paper. 
We use the exact same setup as in~\cite{harutyunyan2021information}, where the code is publicly available\footnote{\url{https://github.com/hrayrhar/f-CMI/tree/master}}. For every number of training data $n$, we draw $m_1$ of the random variable $\rmZ_{[2n]}$, i.e., we select $m_1$ different 2n samples from the original dataset of size $h > 2n$.Then, for each $z_{[2n]}$, we draw $m_2$ different train/test splits, i.e., $m_2$ random realizations of $\rmU$. In total, we have $m_1m_2$ experiments. We report the mean and standard deviation on the $m_1$ results. For the CIFER10 experiments, we select $m_1=2$ and $m_2=20$. For its noisy variant, we select $m_1=5$ and $m_2=15$. For both datasets, we use ResNet50 pre-trained on ImageNet. The training is conducted for 40 epochs using SGD with a learning rate of 0.01 and a batch size of 256. 

In practice,  $P(\rmY=y)$ is unknown and needs to be estimated using the available data. One simple way is to use $\frac{n^y_{Z_{[2n]}}}{2n}$ as an estimate $P(\rmY=y)$. Thus, $n^y = n P(\rmY=y) $used in Section~\ref{section_cmisettings}, can be estimated with $\frac{2}{n_{z_{[2n]}}^y}$. Note that our empirical results validate the proposed bounds for this estimated class-wise generalization error.

\subsection{full class-generalization error vs. standard generalization results on CIFAR10}

\begin{figure*}[h]
\centering
\includegraphics[width=0.7\linewidth]{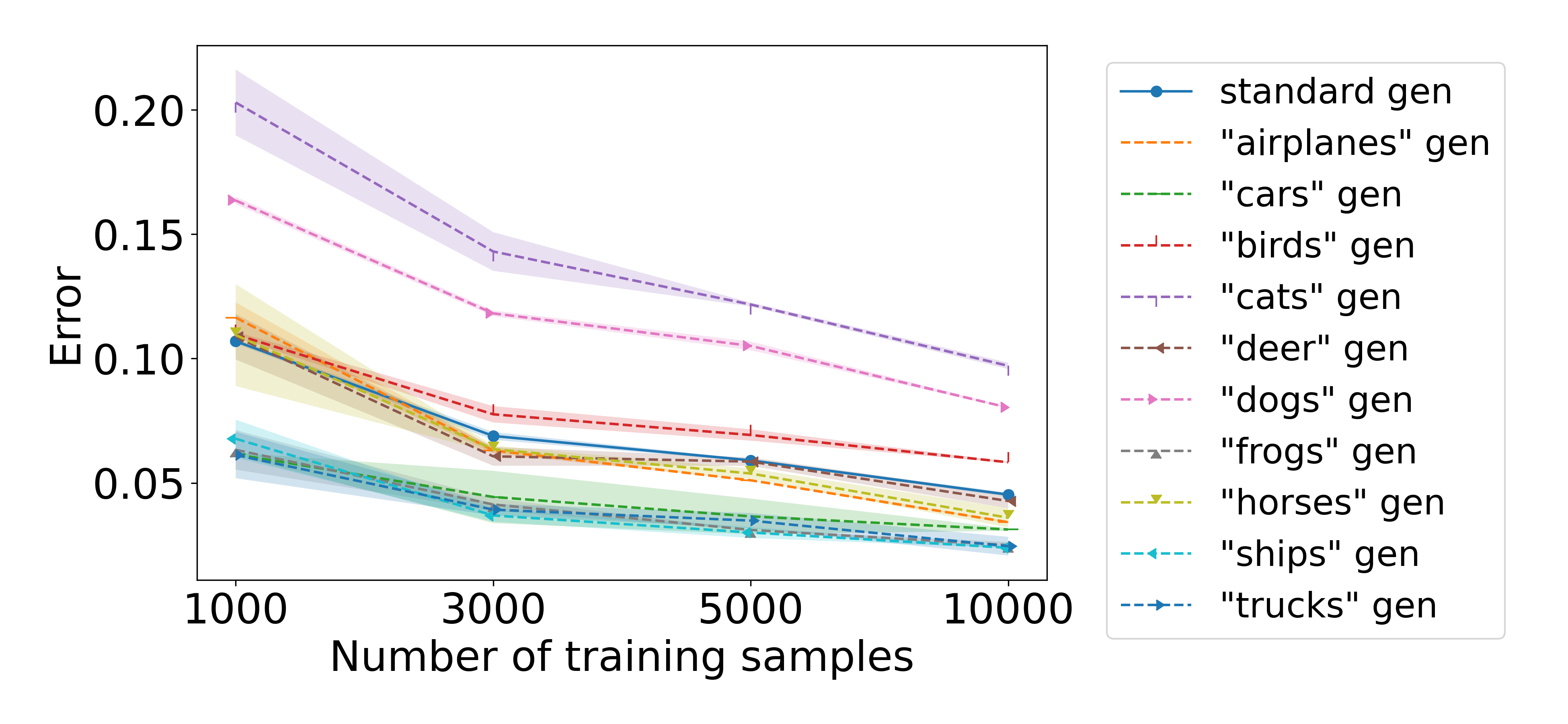}
\caption{The standard generalization error and the generalization error relative for all classes on CIFAR10 as a function of the number of training data. }
\label{full_characterizationCIFAR10}
\end{figure*}

As a supplement to Figure~\ref{motivating_example} (left), we plot the standard generalization error along with the class-generalization error of all the classes of CIFAR10 in Figure~\ref{full_characterizationCIFAR10}. Similar to the observations highlighted in Section~\ref{introduction_sec}, we notice the significant variability in generalization performance across different classes.

\subsection{Numerical results with our bounds for all classes of CIFAR10 and its noisy variant} \label{classgen_results}

\begin{figure*}[h]
\centering
\includegraphics[width=0.32\linewidth]{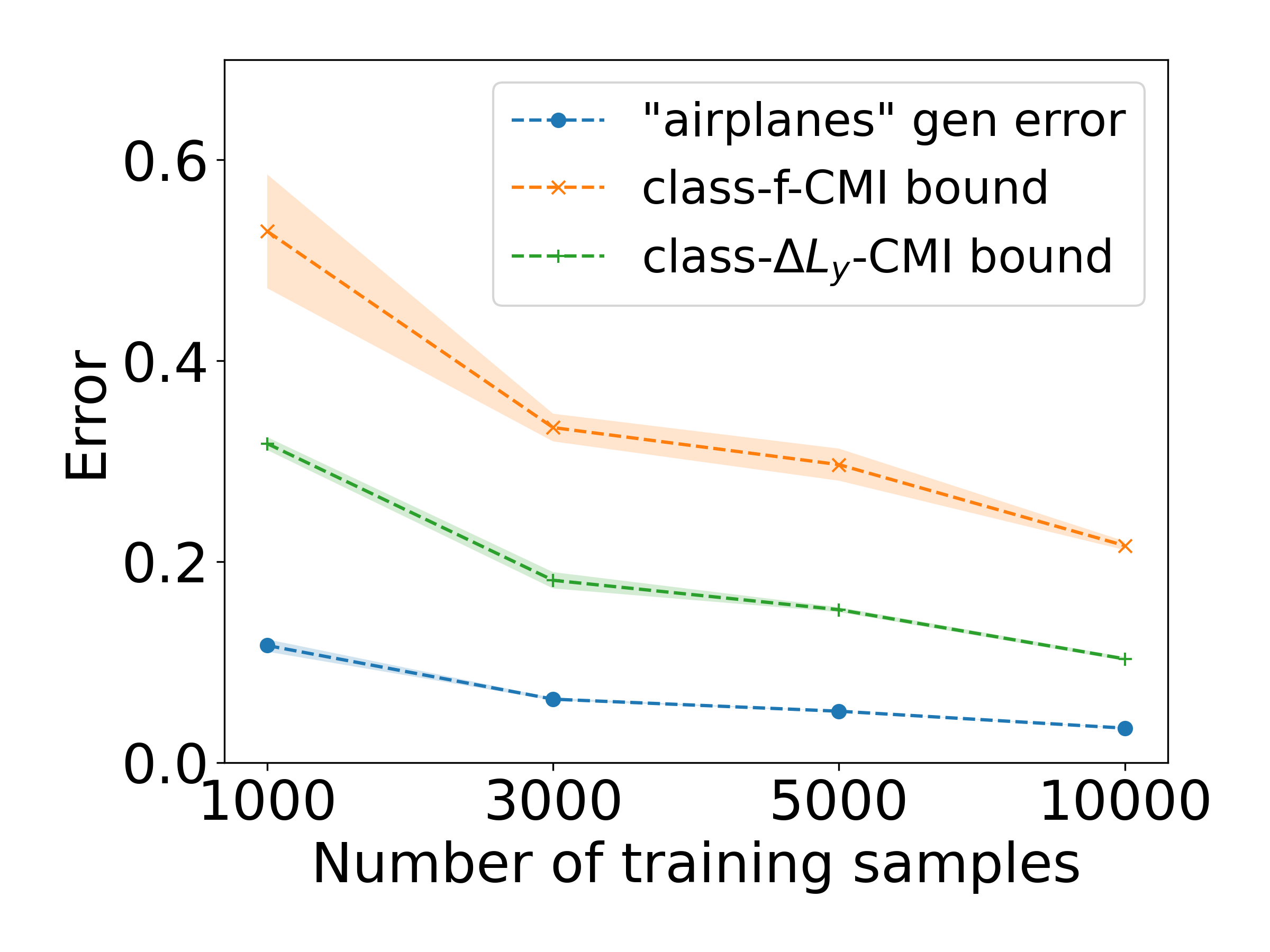}
\includegraphics[width=0.32\linewidth]{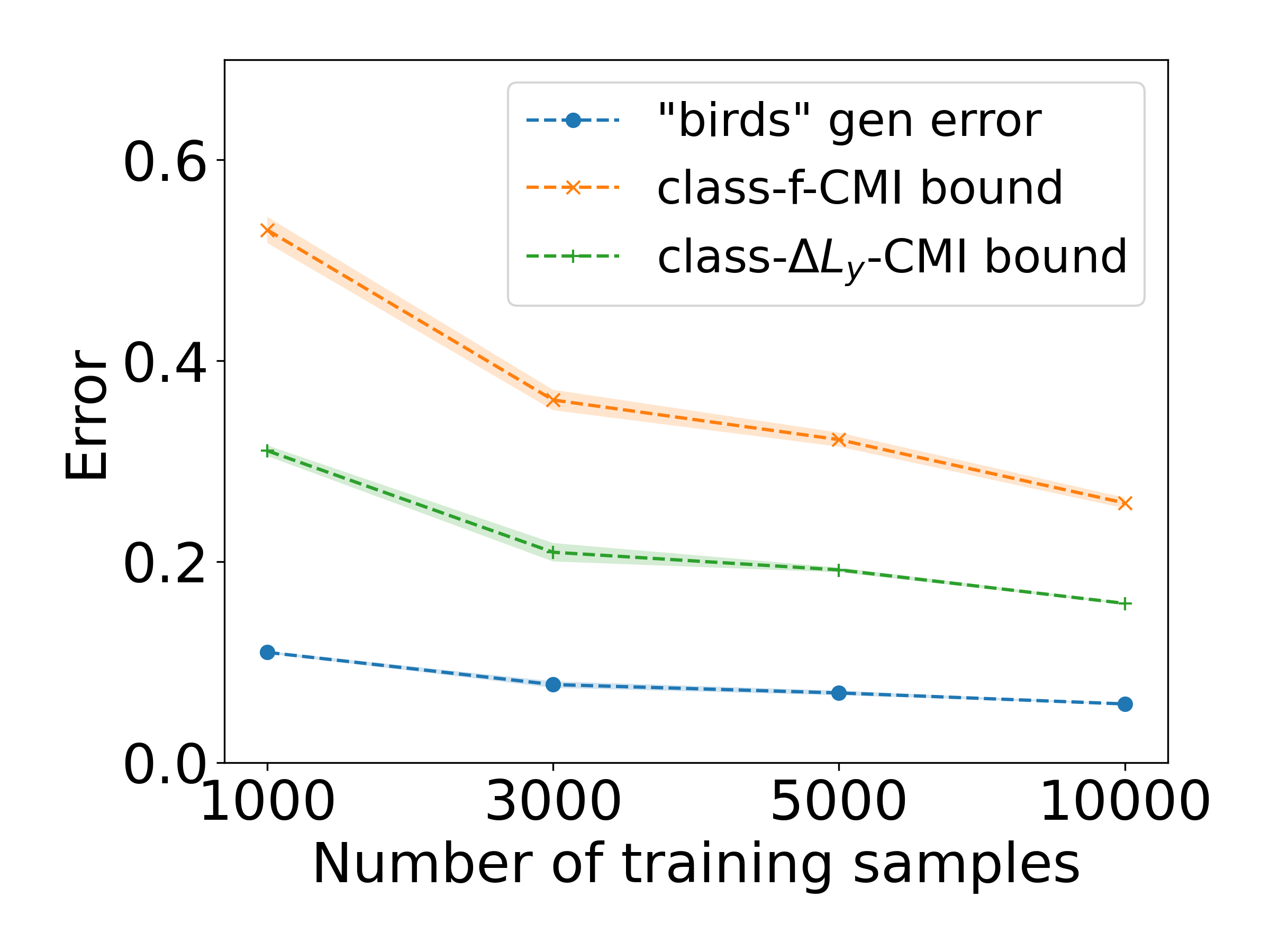}
\includegraphics[width=0.32\linewidth]{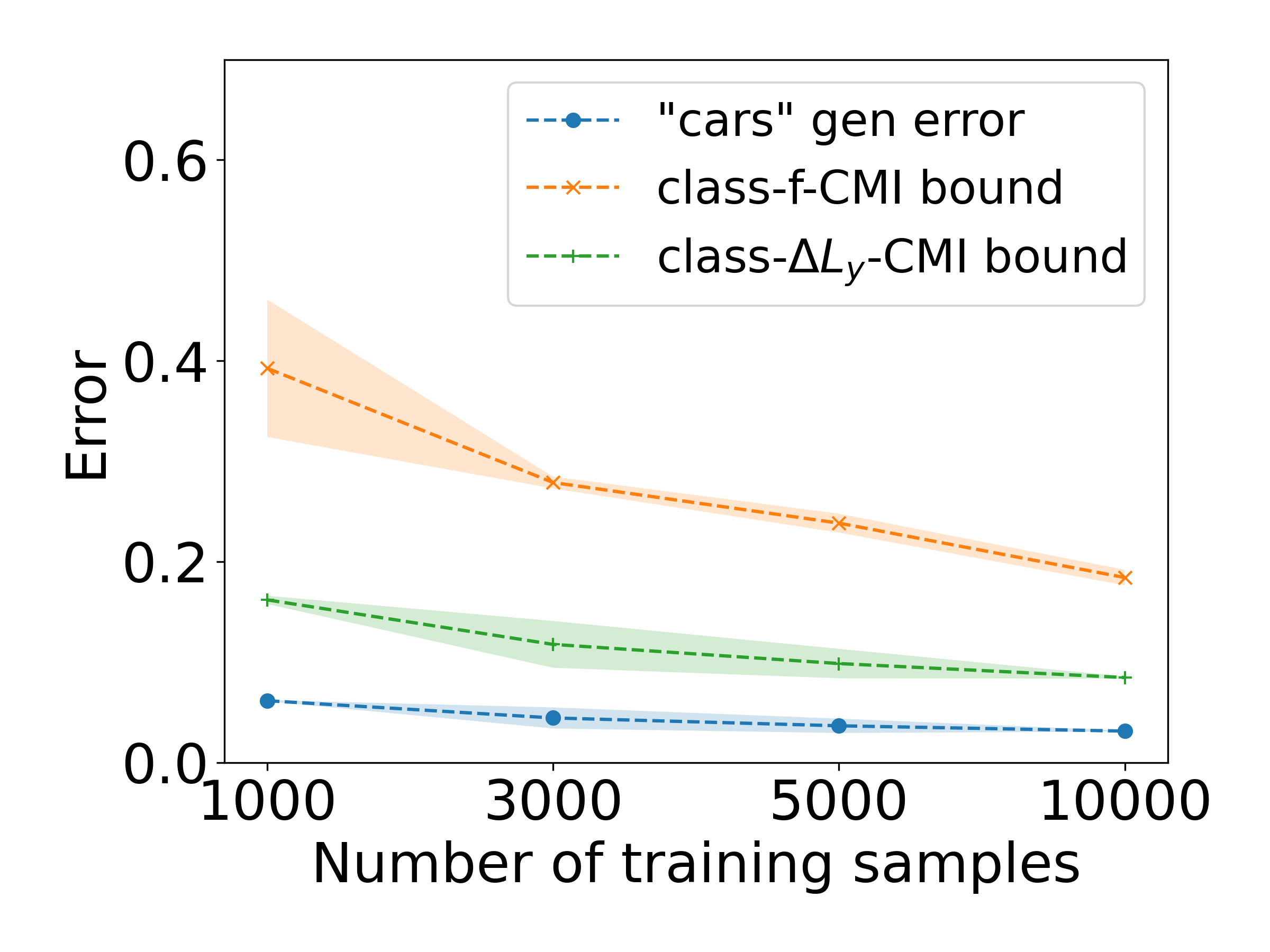}
\includegraphics[width=0.32\linewidth]{Figures/Cifar10/cifar10classcatsgenerrorplot.png}
\includegraphics[width=0.32\linewidth]{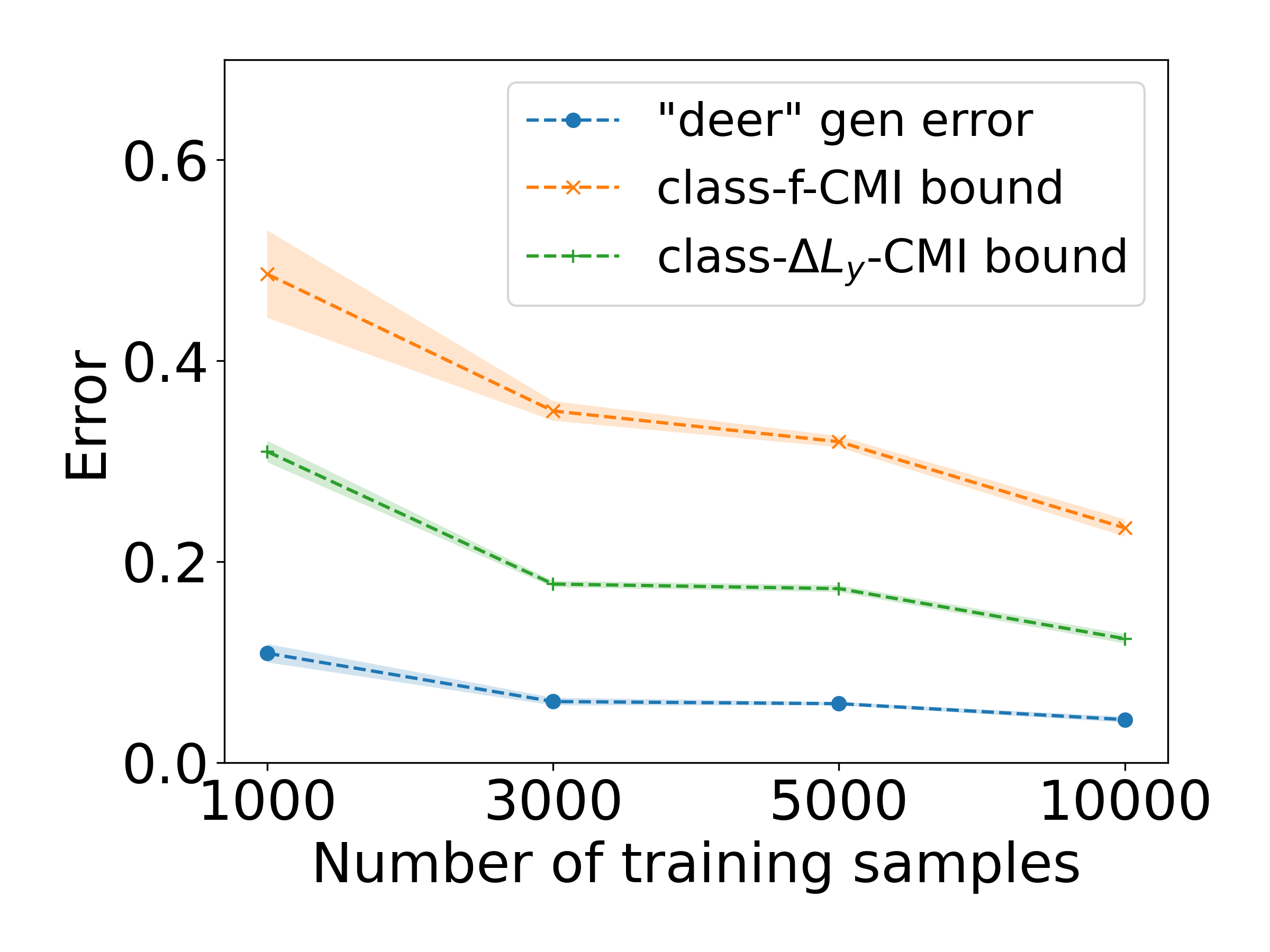}
\includegraphics[width=0.32\linewidth]{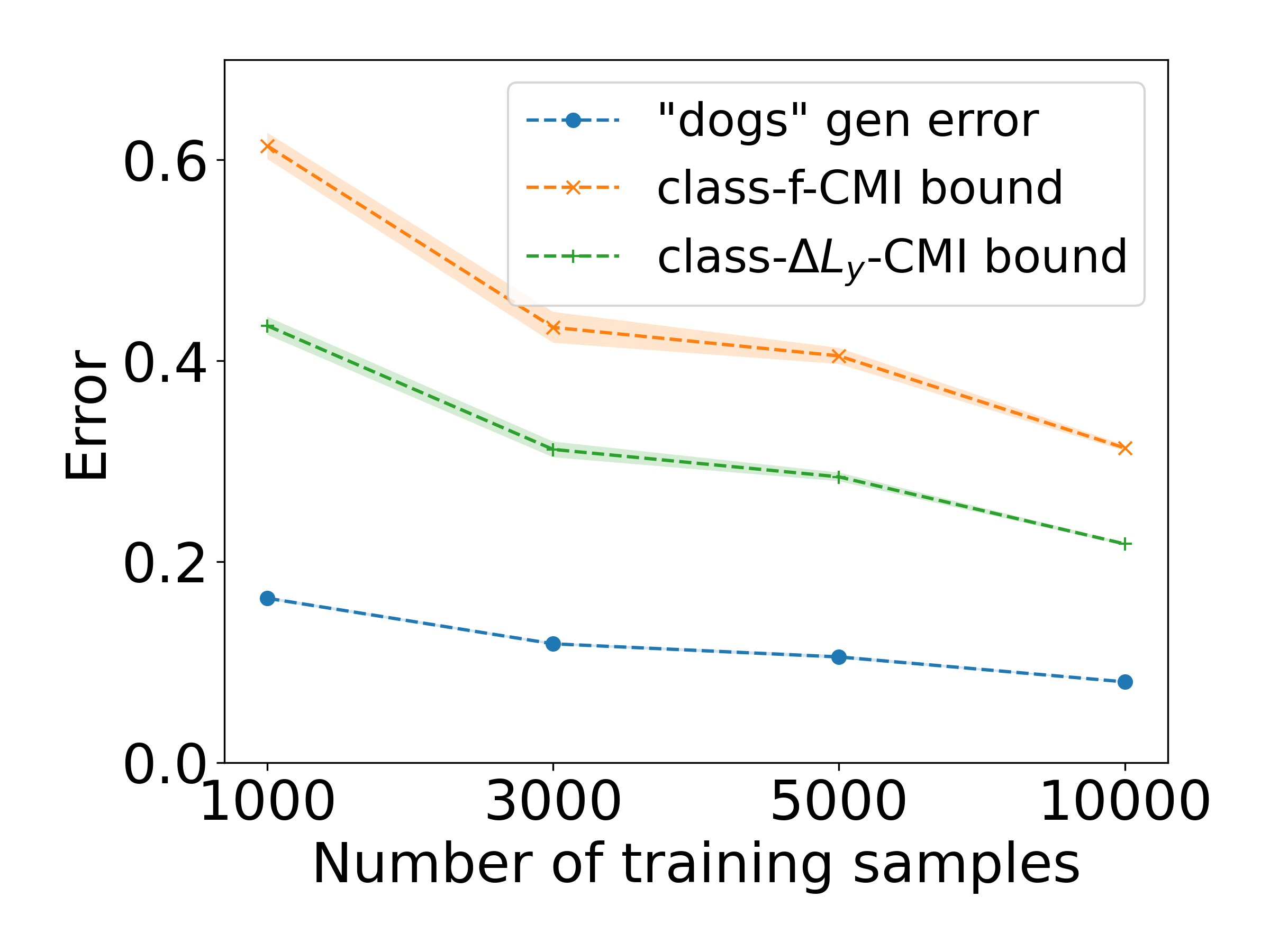}
\includegraphics[width=0.32\linewidth]{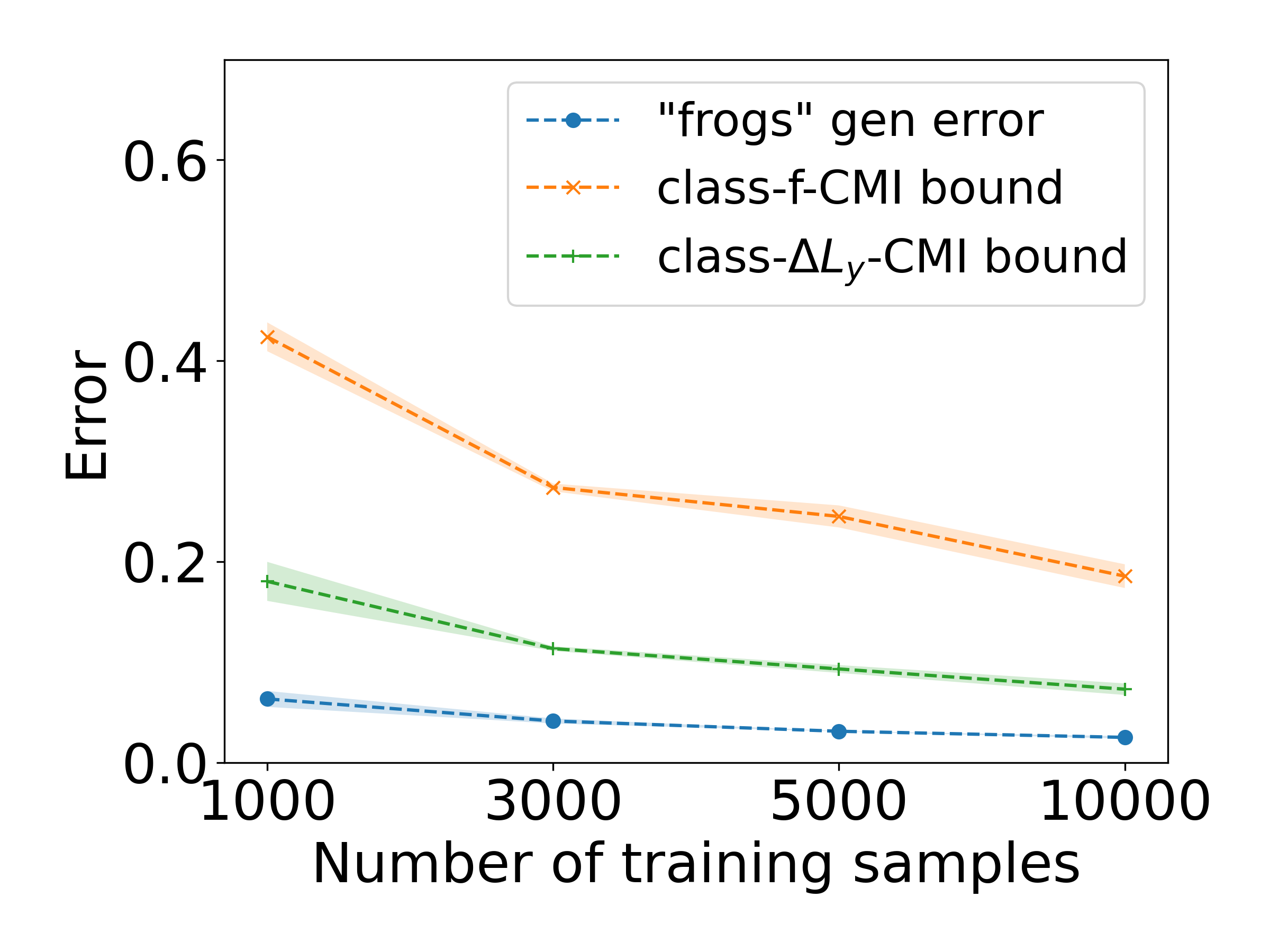}
\includegraphics[width=0.32\linewidth]{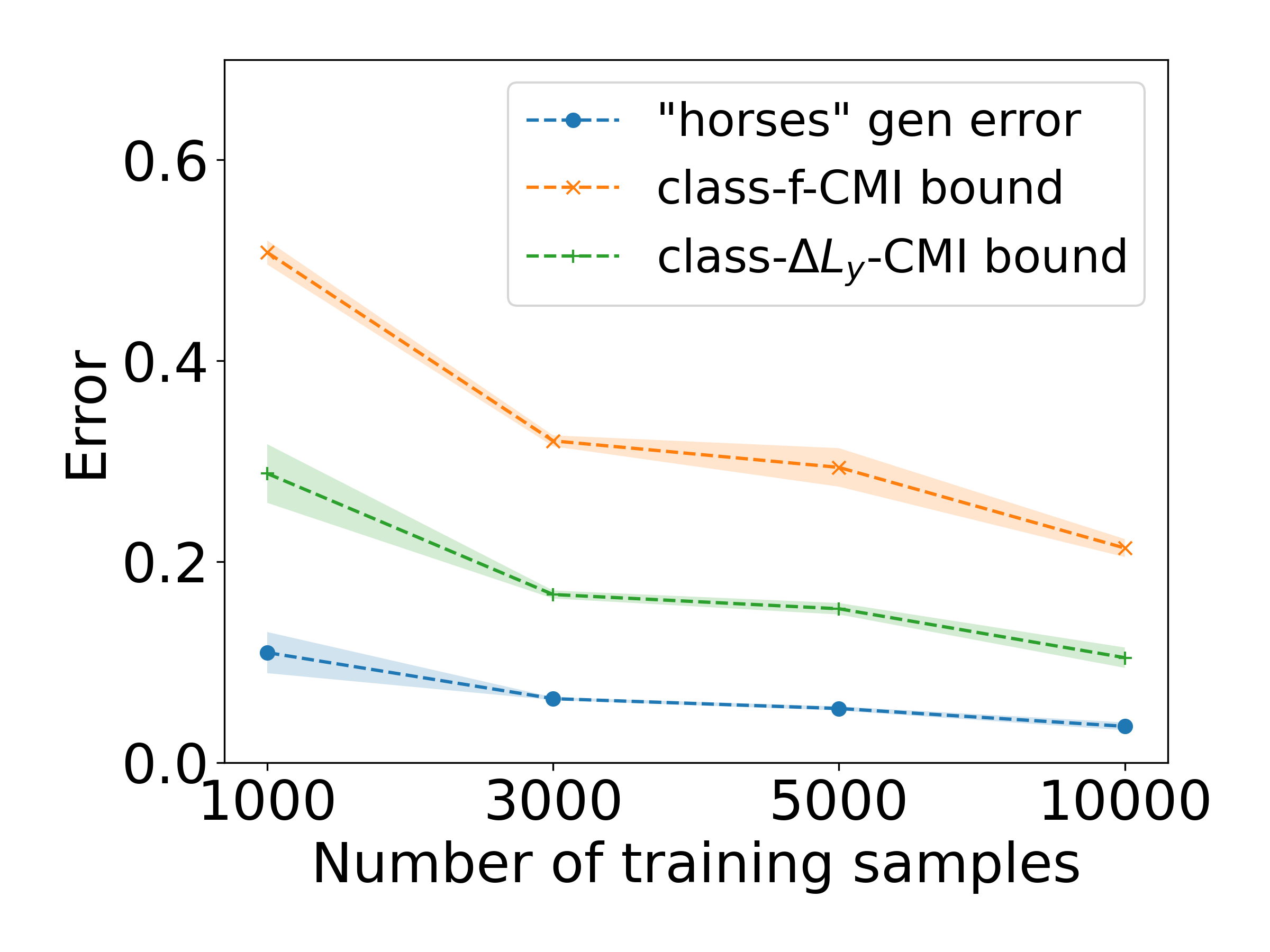}
\includegraphics[width=0.32\linewidth]{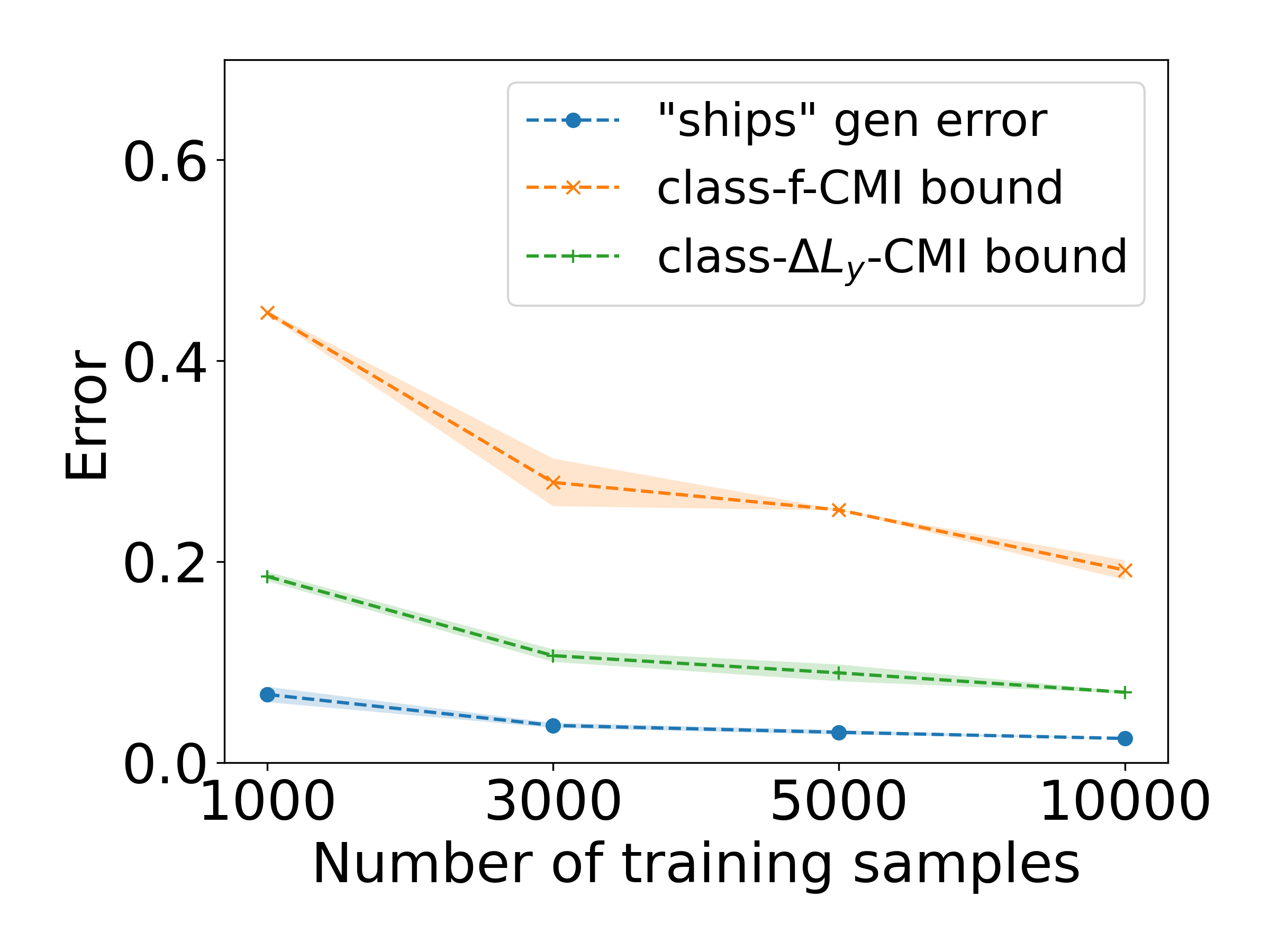}
\includegraphics[width=0.32\linewidth]{Figures/Cifar10/cifar10classtrucksgenerrorplot.png}
\includegraphics[width=0.28\linewidth]{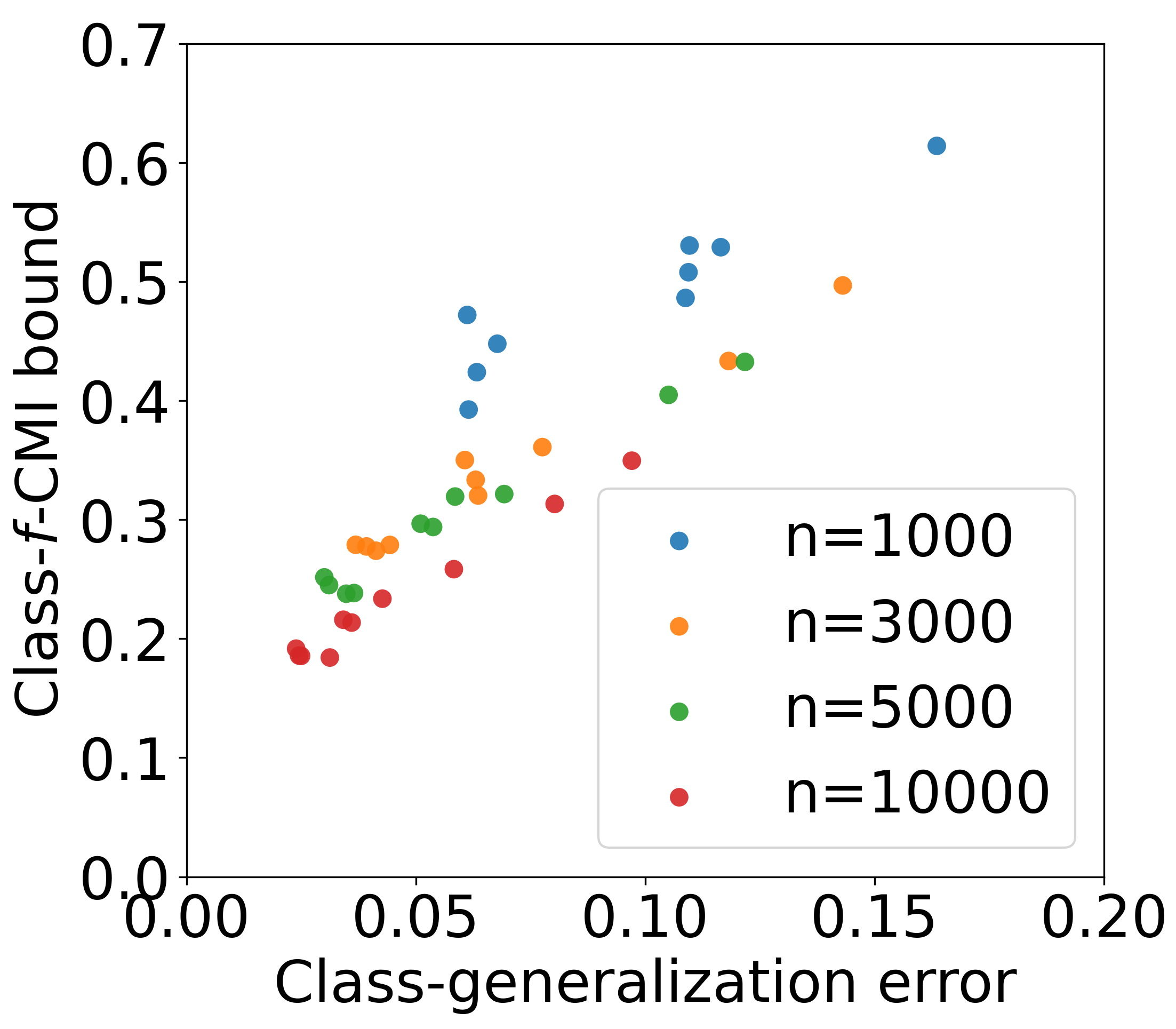}
\caption{Class-wise generalization on the 10 classes of CIFAR10 and the scatter plot between class-generalization error and the class-$f$-CMI bound in Theorem~\ref{fCMI_class_bound}. }
\label{cifar10boundsresults}
\end{figure*}

\begin{figure*}[h]
\centering
\includegraphics[width=0.32\linewidth]{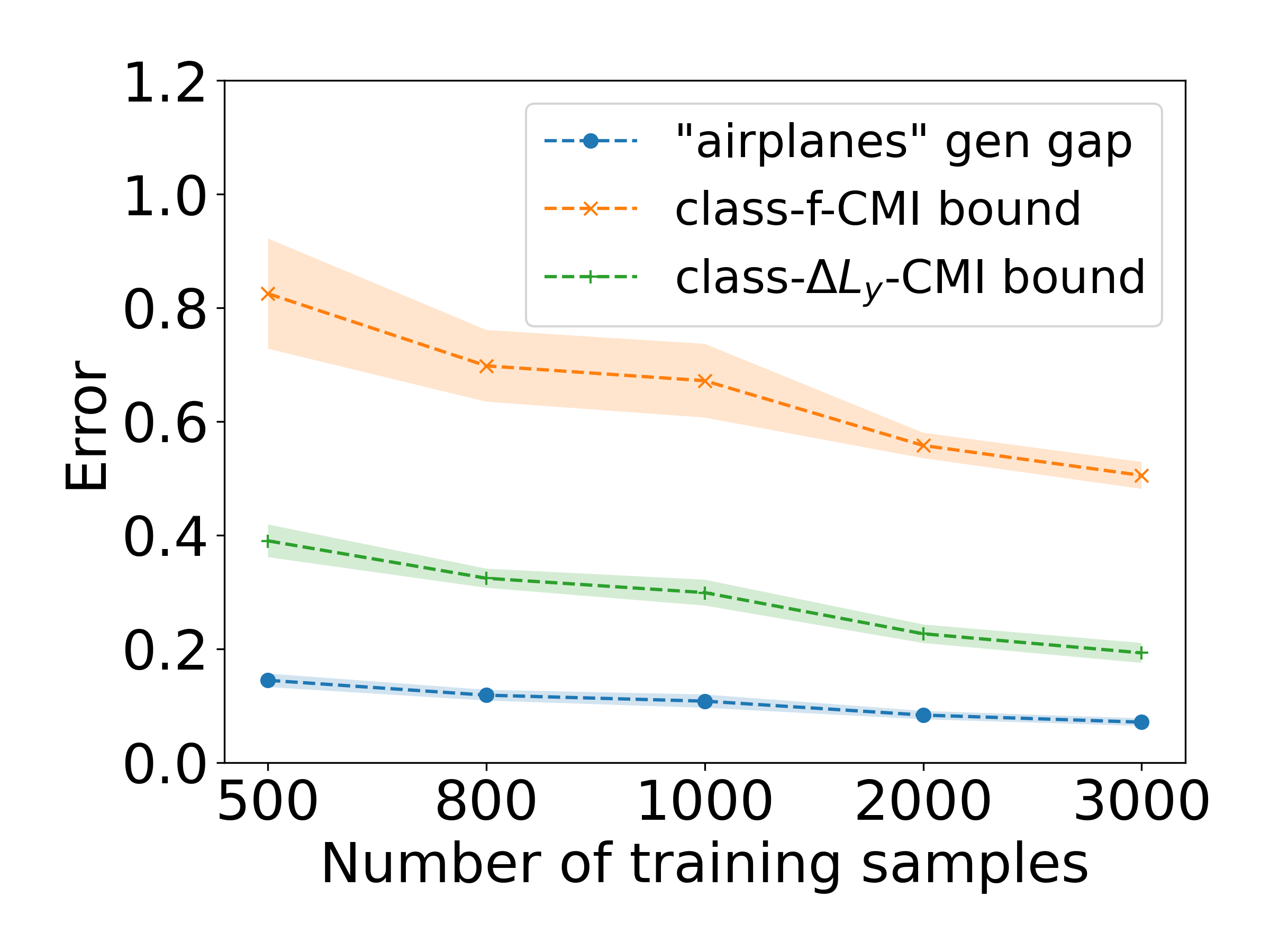}
\includegraphics[width=0.32\linewidth]{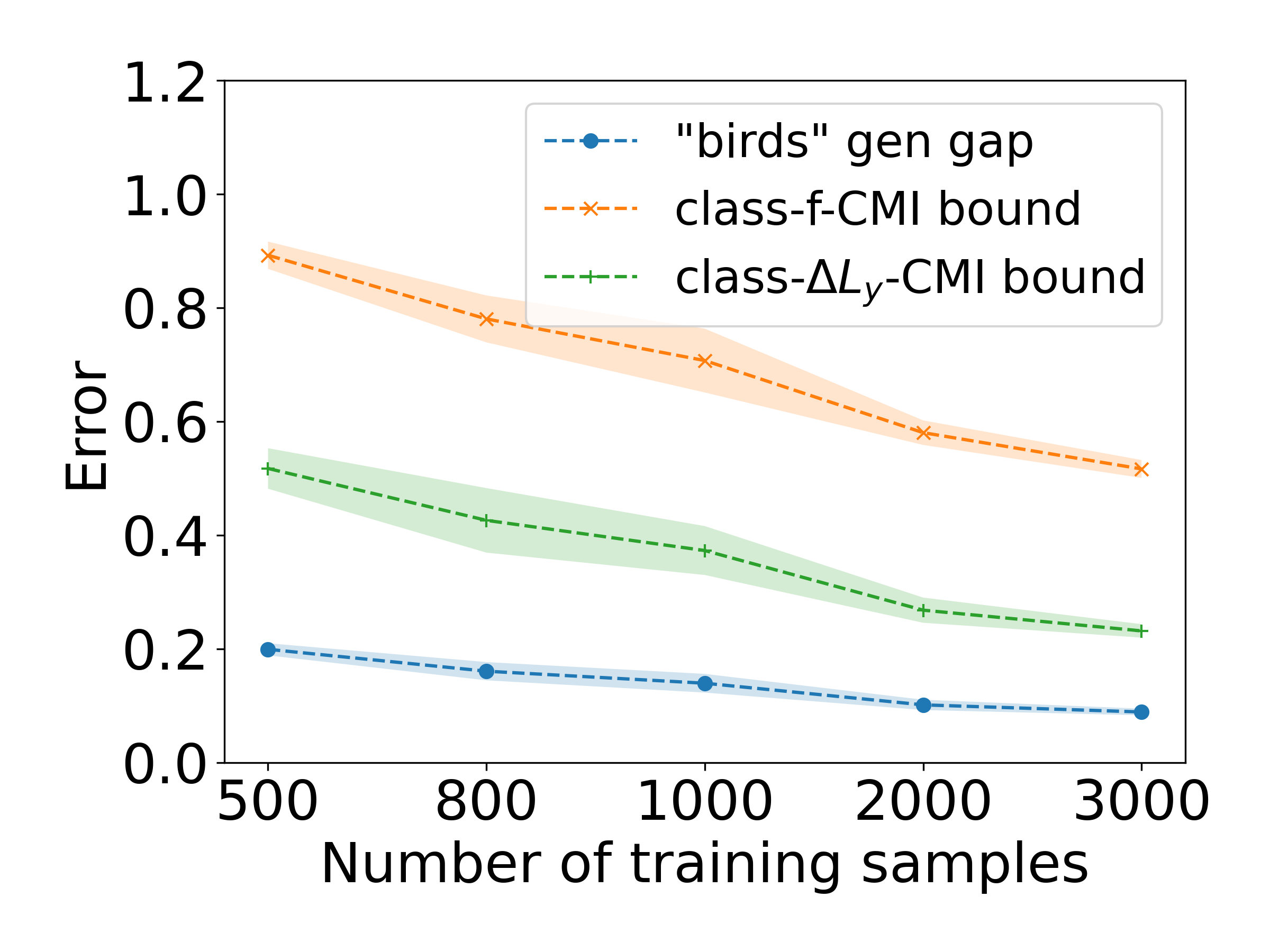}
\includegraphics[width=0.32\linewidth]{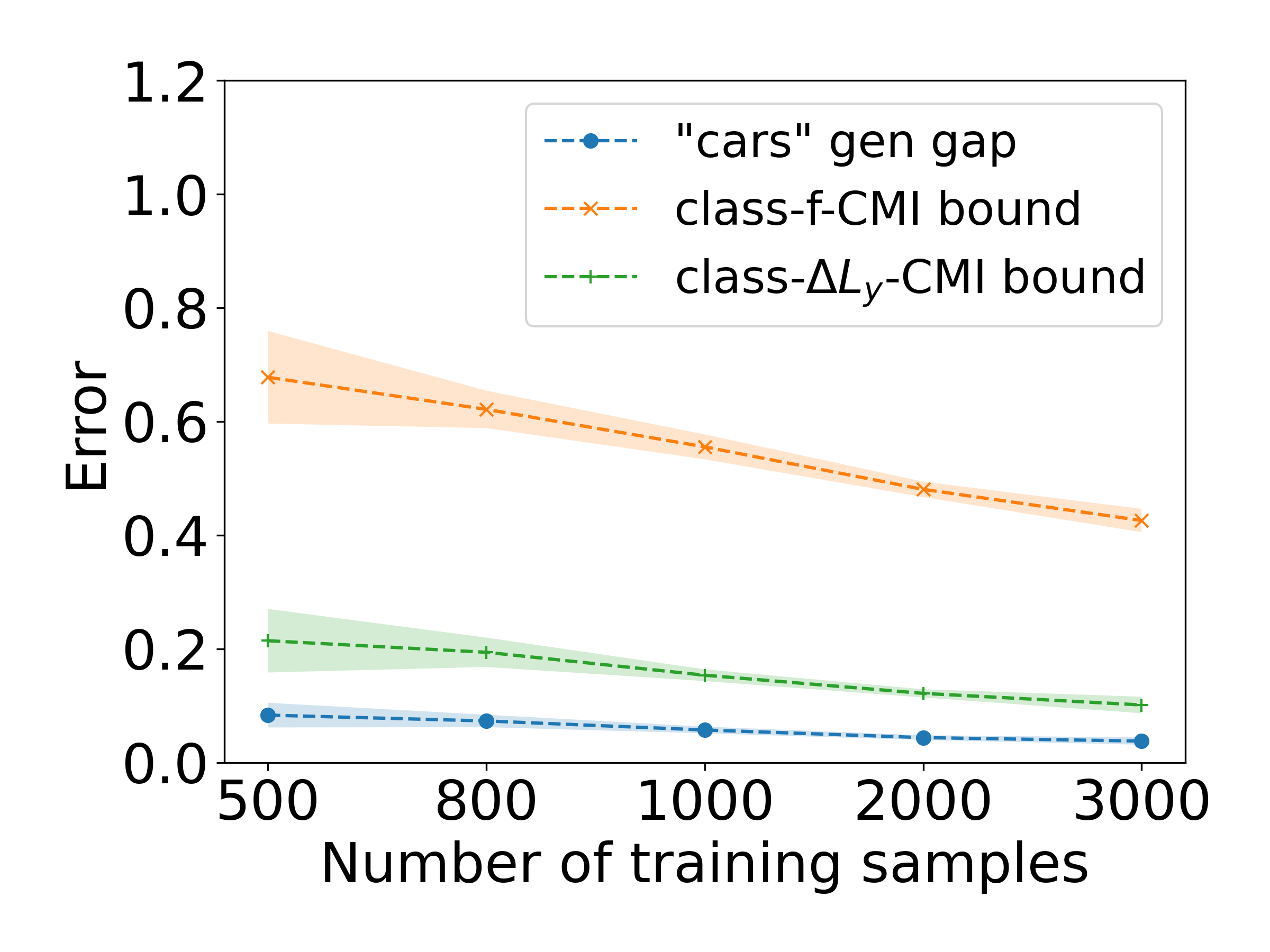}
\includegraphics[width=0.32\linewidth]{Figures/noisyCifar05/noisycifar10classcatsgenerrorplot.png}
\includegraphics[width=0.32\linewidth]{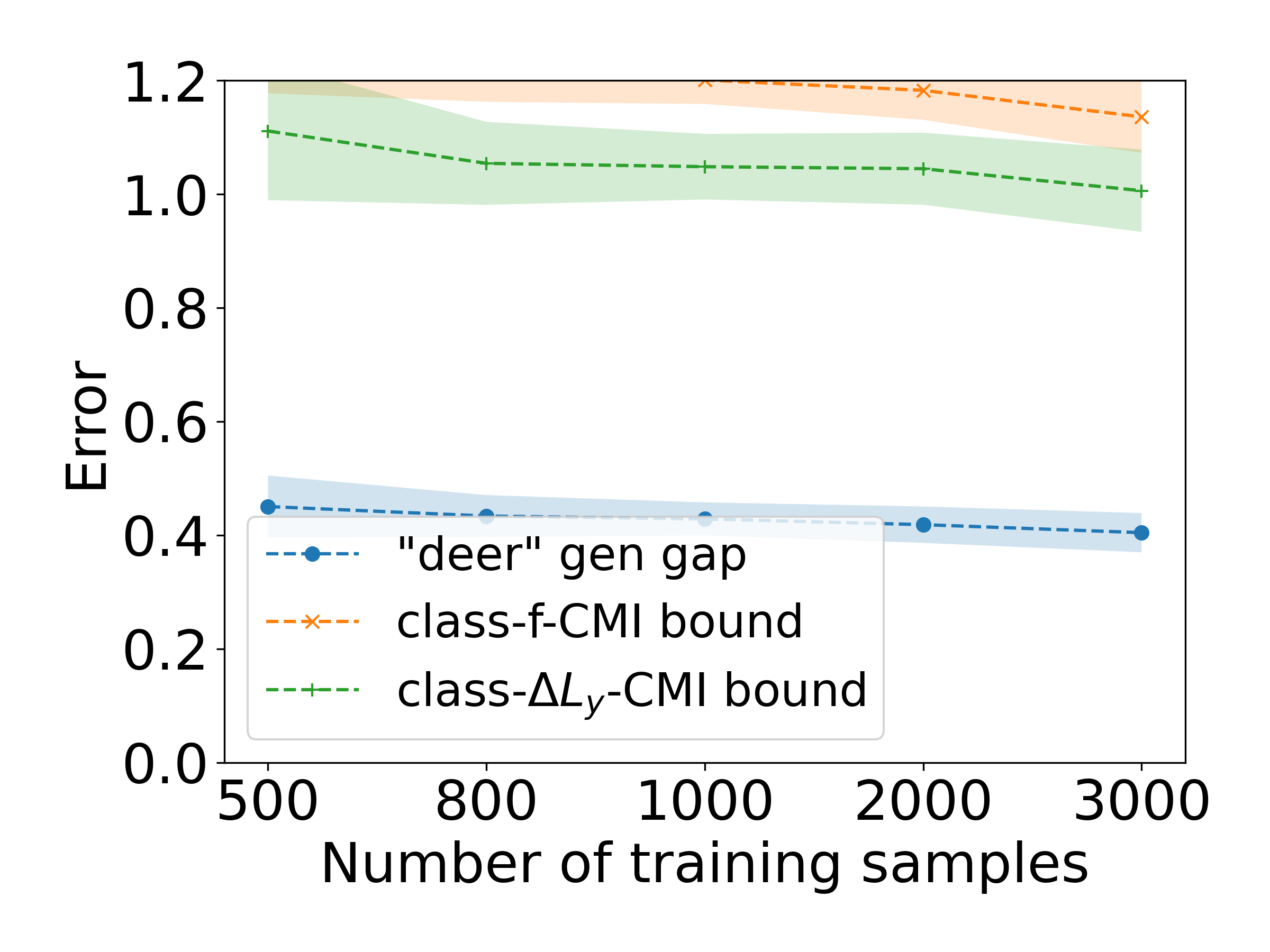}
\includegraphics[width=0.32\linewidth]{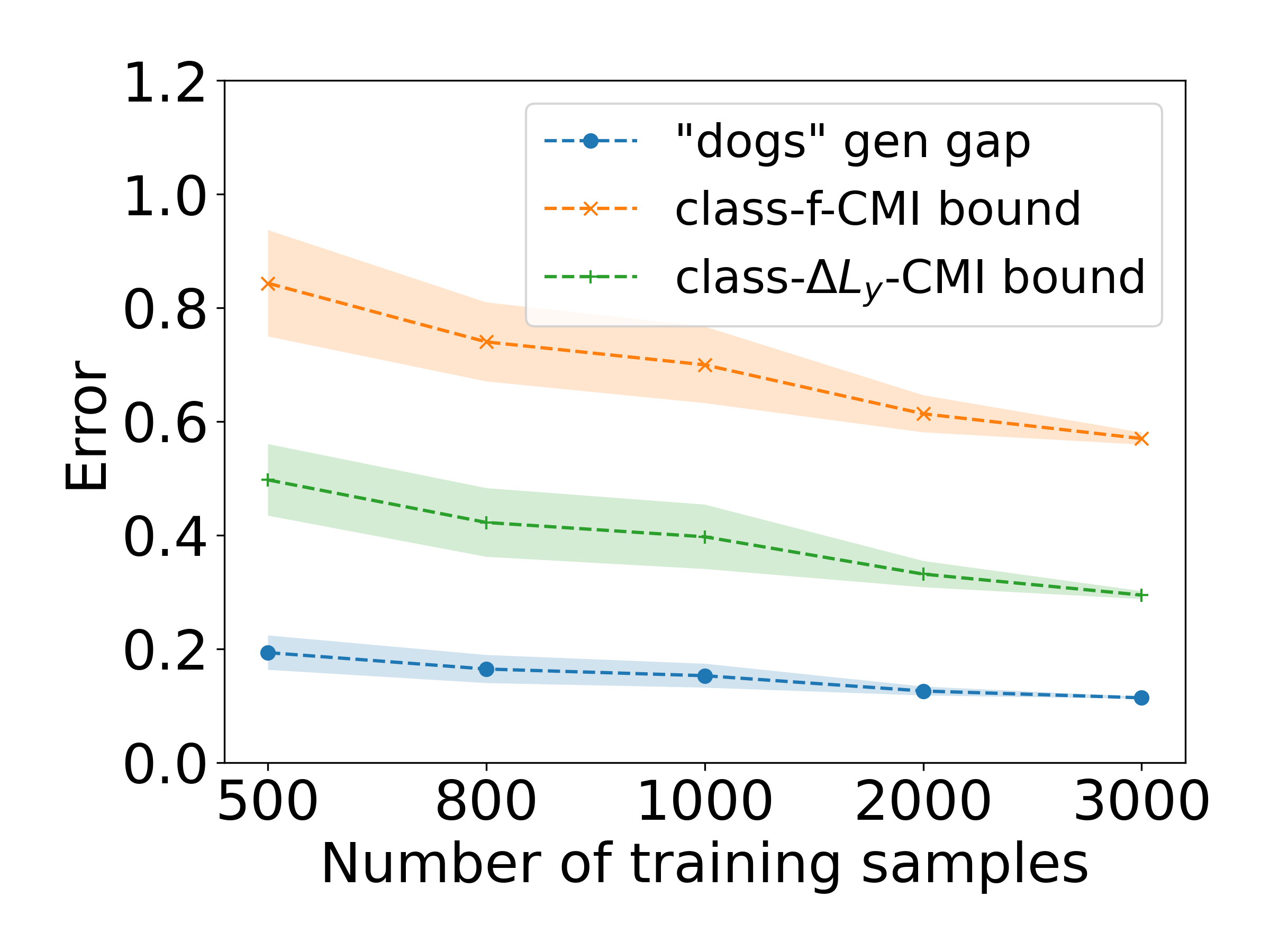}
\includegraphics[width=0.32\linewidth]{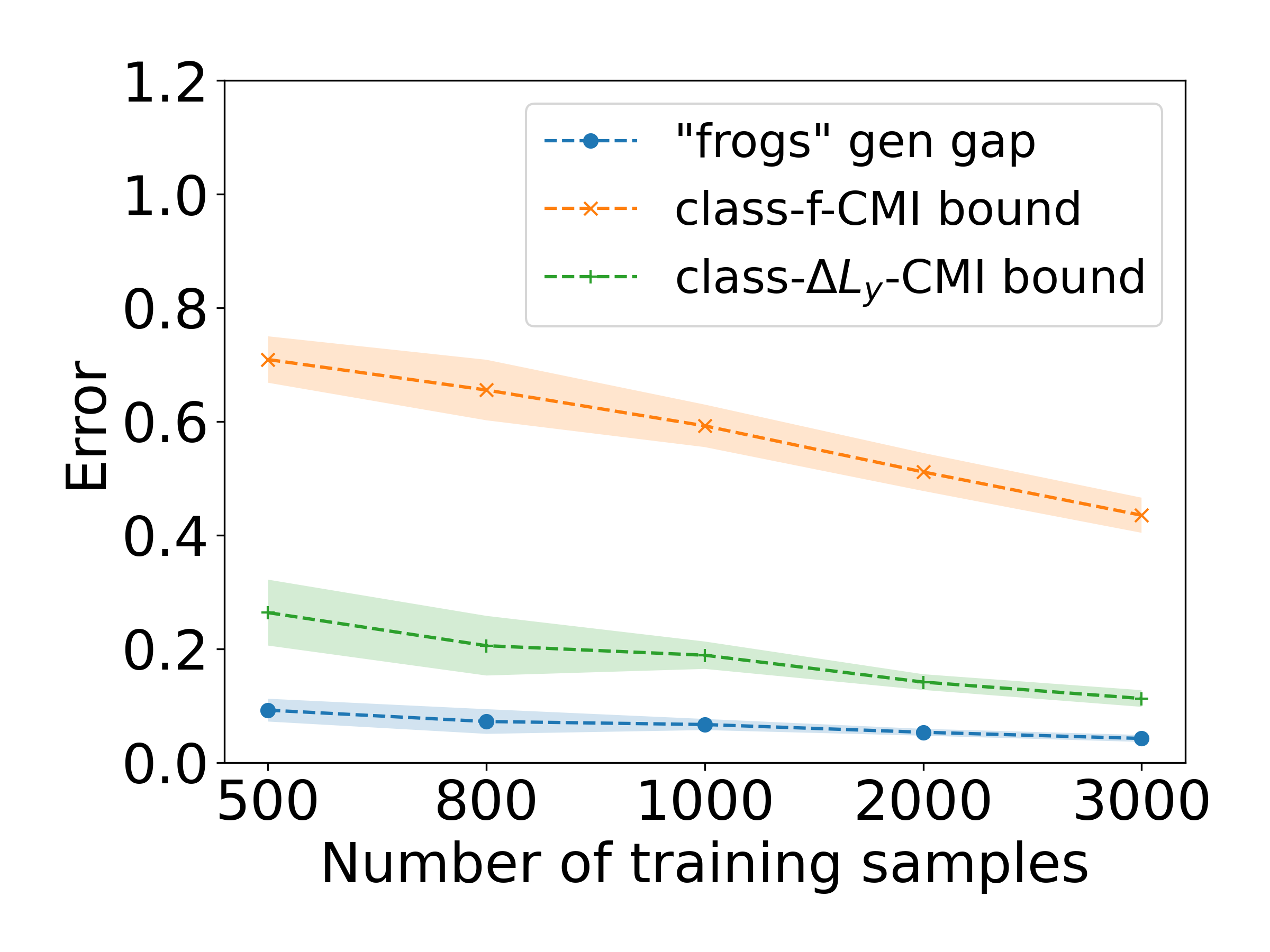}
\includegraphics[width=0.32\linewidth]{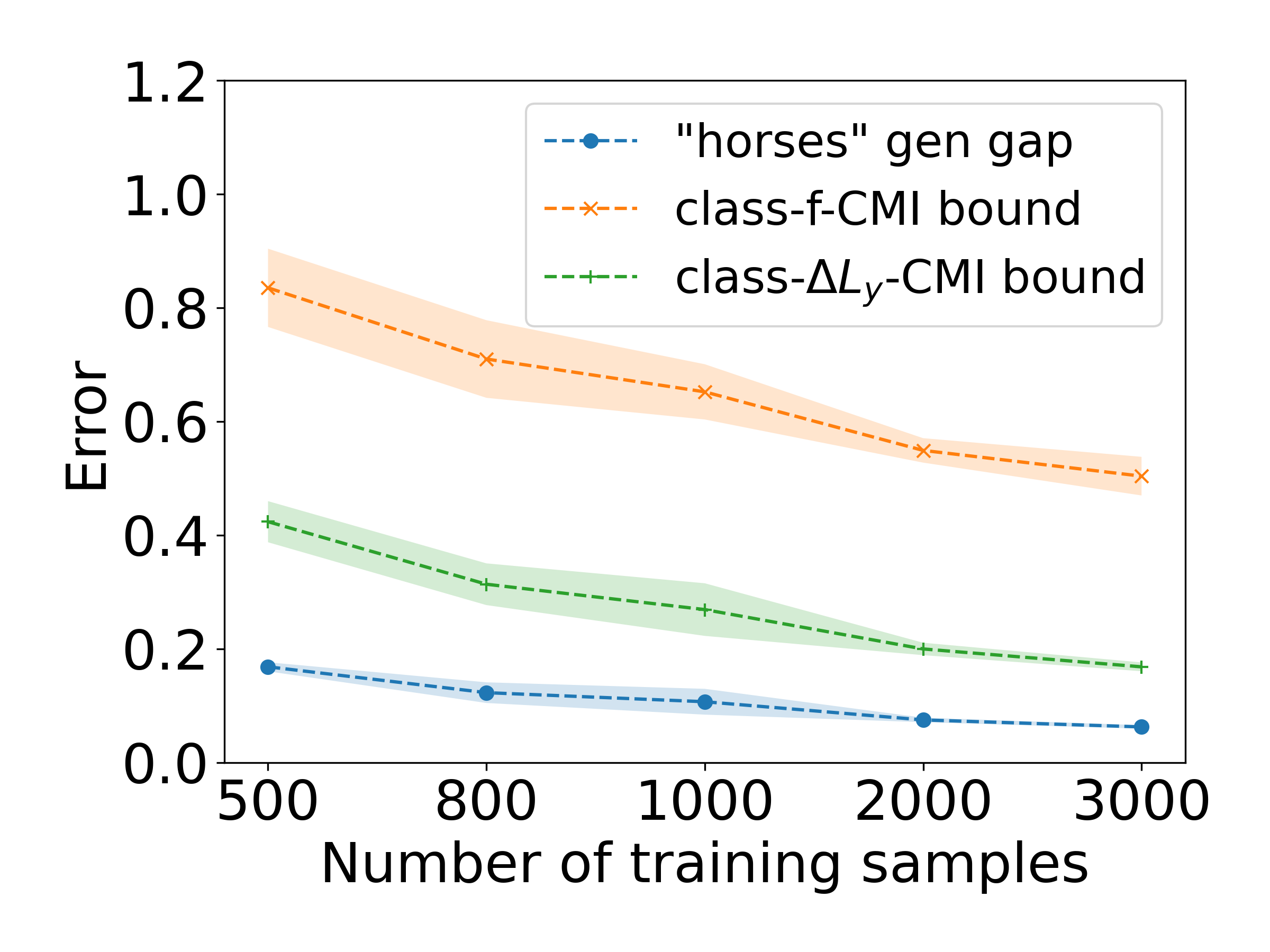}
\includegraphics[width=0.32\linewidth]{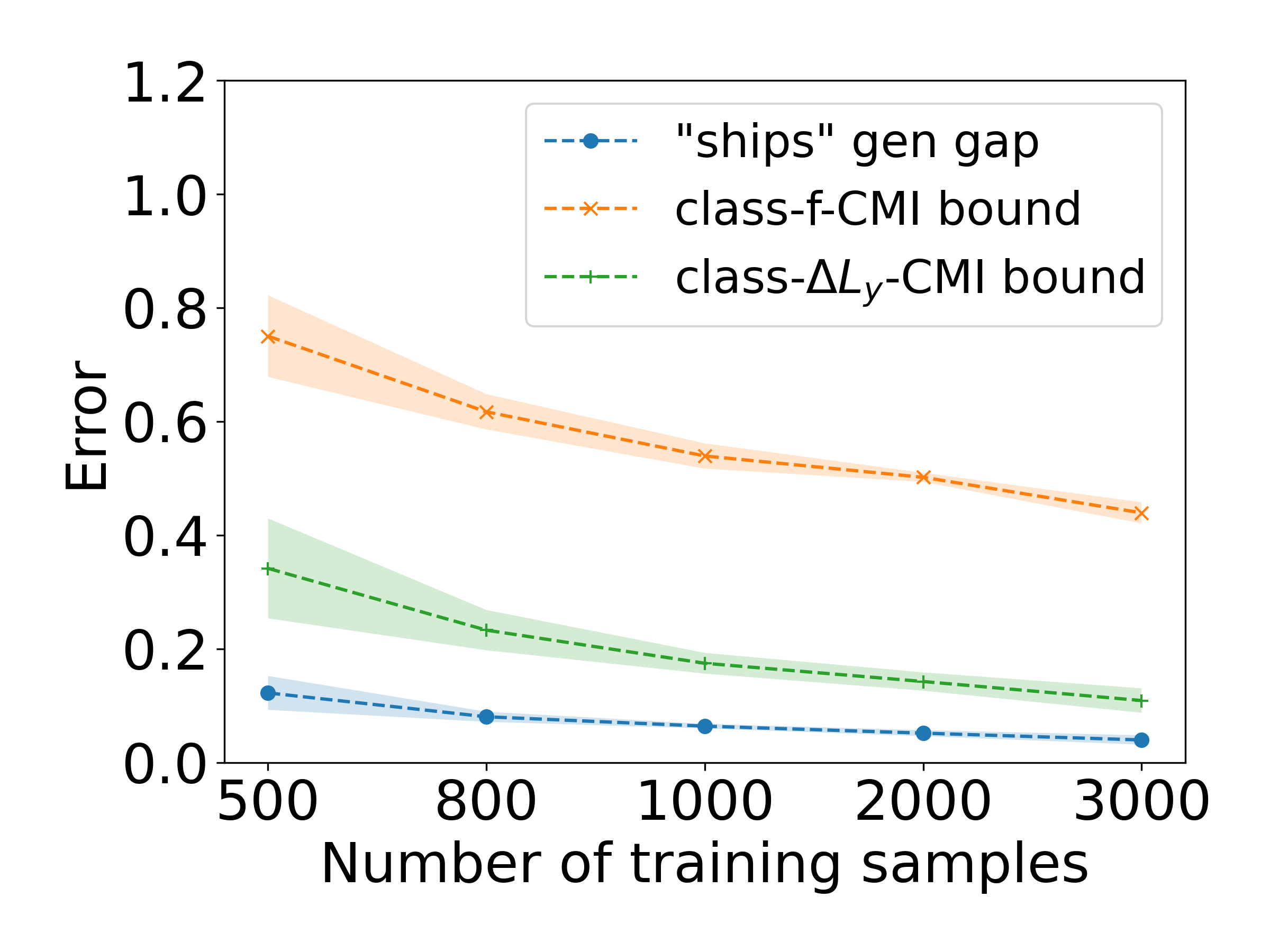}
\includegraphics[width=0.32\linewidth]{Figures/noisyCifar05/noisycifar10classtrucksgenerrorplot.png}
\includegraphics[width=0.28\linewidth]{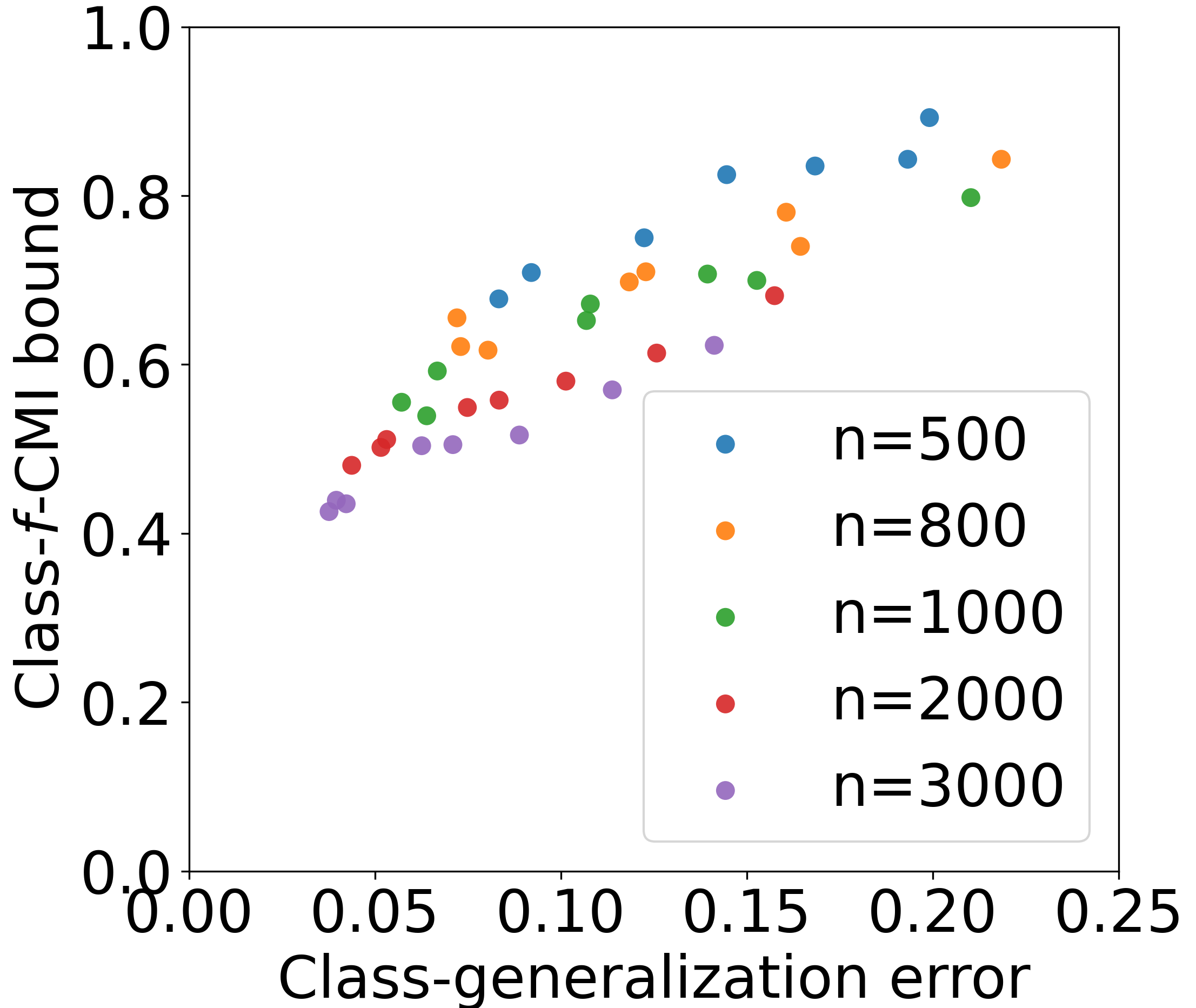}

\caption{Class-wise generalization on the 10 classes of noisy CIFAR10 (clean validation)  and the scatter plot between class-generalization error and the class-$f$-CMI bound in Theorem~\ref{fCMI_class_bound}. }
\label{noisycifar10boundsresults}

\end{figure*}

In Figure~\ref{cifar10boundsresults}, we present the results of the empirical evaluation of our bounds on all the classes of CIFAR10. Moreover, we generate the scatter plot between the class-generalization error and the class-$f$-CMI bound. We note that similar to the class-$\Delta L_y$ results in Figure~\ref{CMIbounds_results}, our bound scales linearly with the error. The results of noisy CIFAR10, presented in Figure~\ref{noisycifar10boundsresults}, are also consistent with these findings.

\begin{figure*}[h]
\centering
\includegraphics[width=0.32\linewidth]{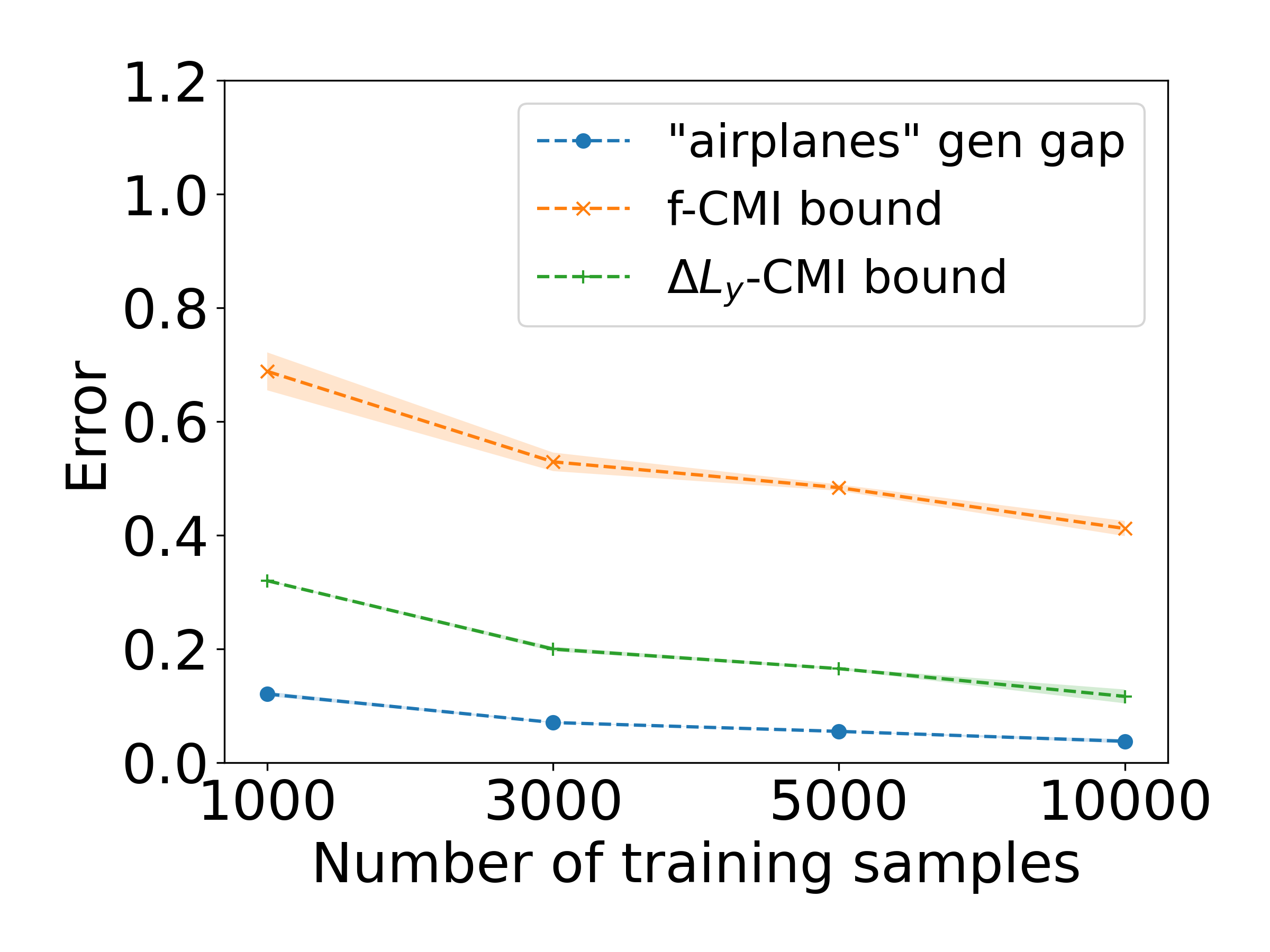}
\includegraphics[width=0.32\linewidth]{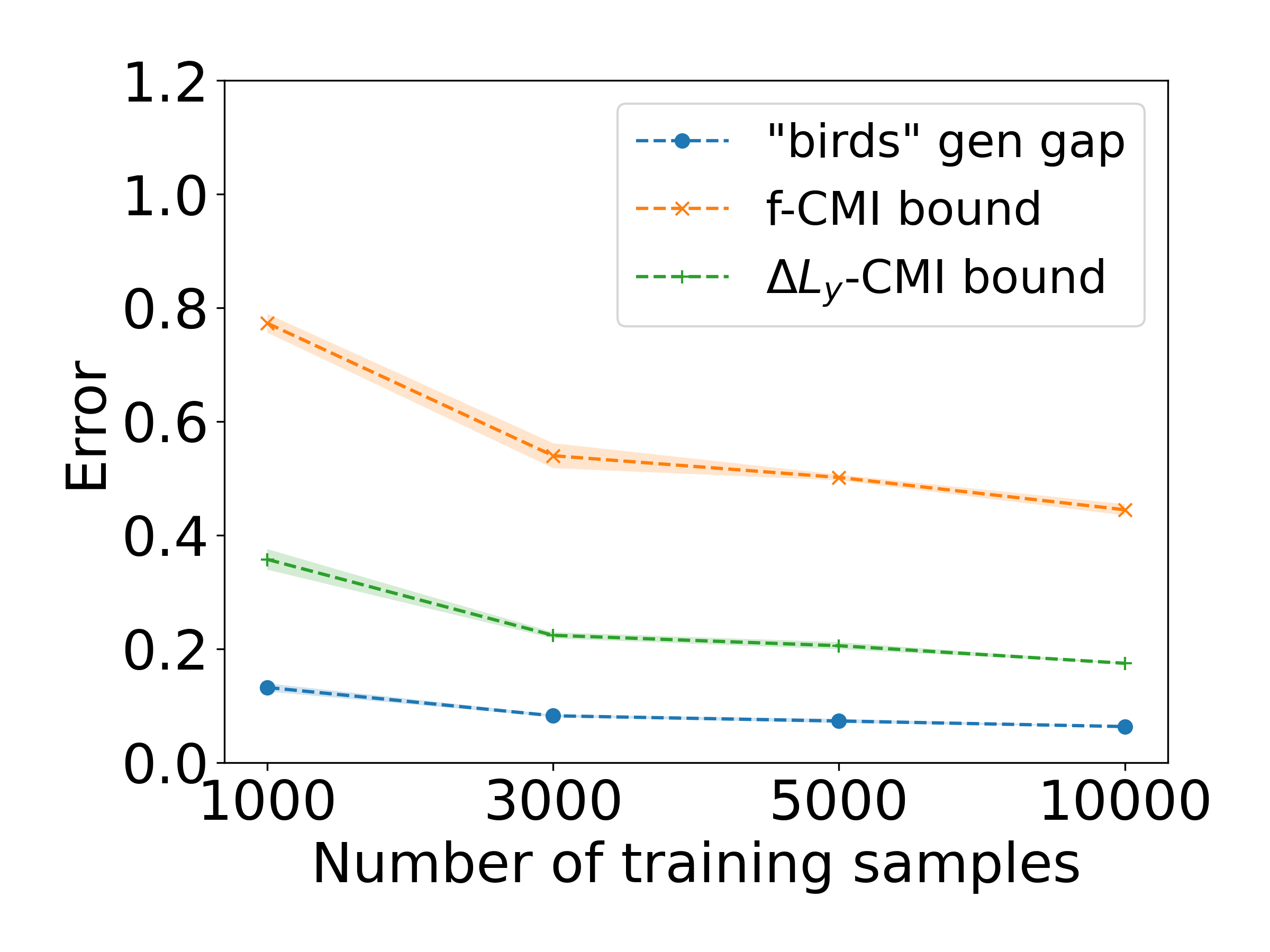}
\includegraphics[width=0.32\linewidth]{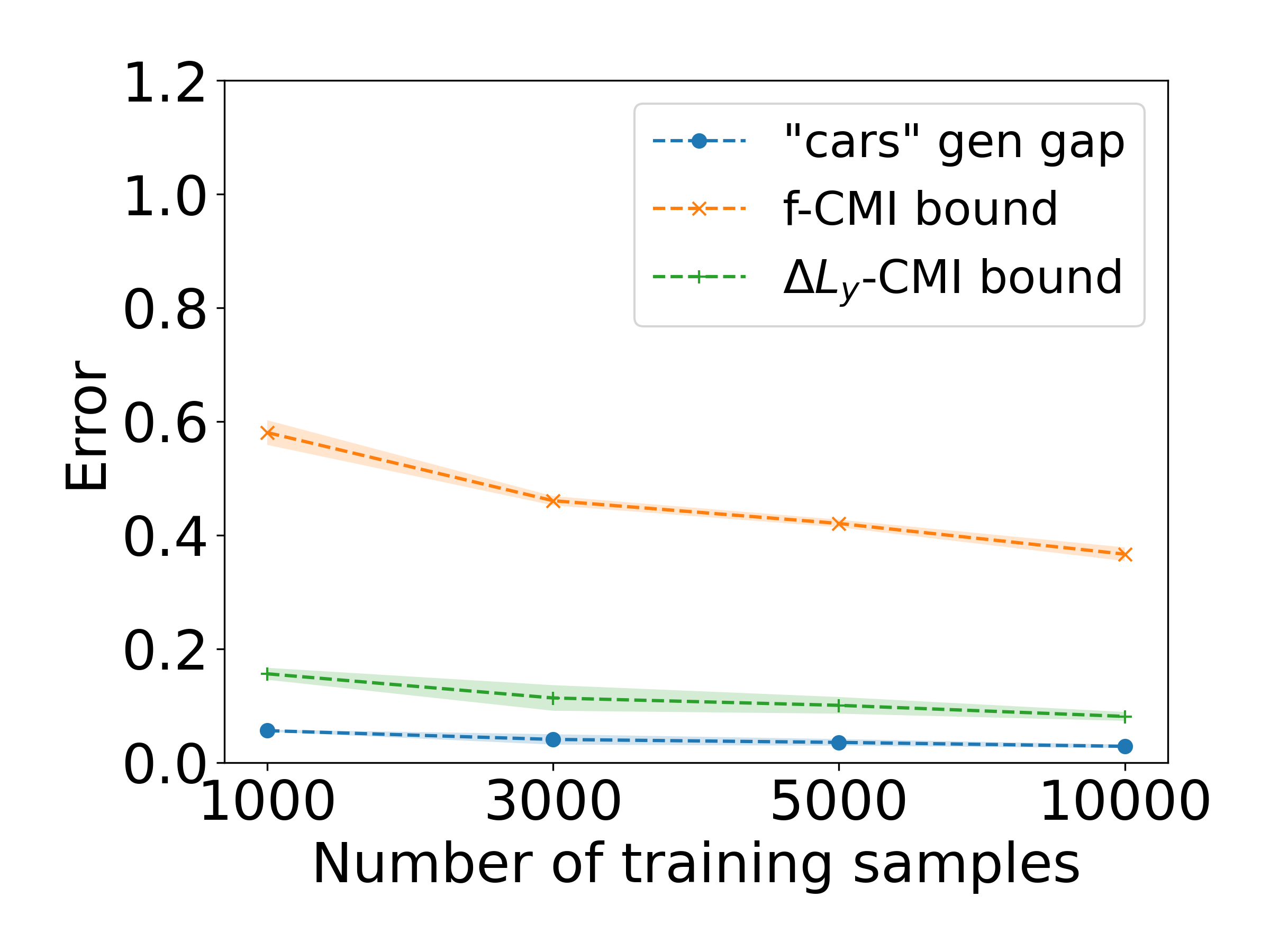}
\includegraphics[width=0.32\linewidth]{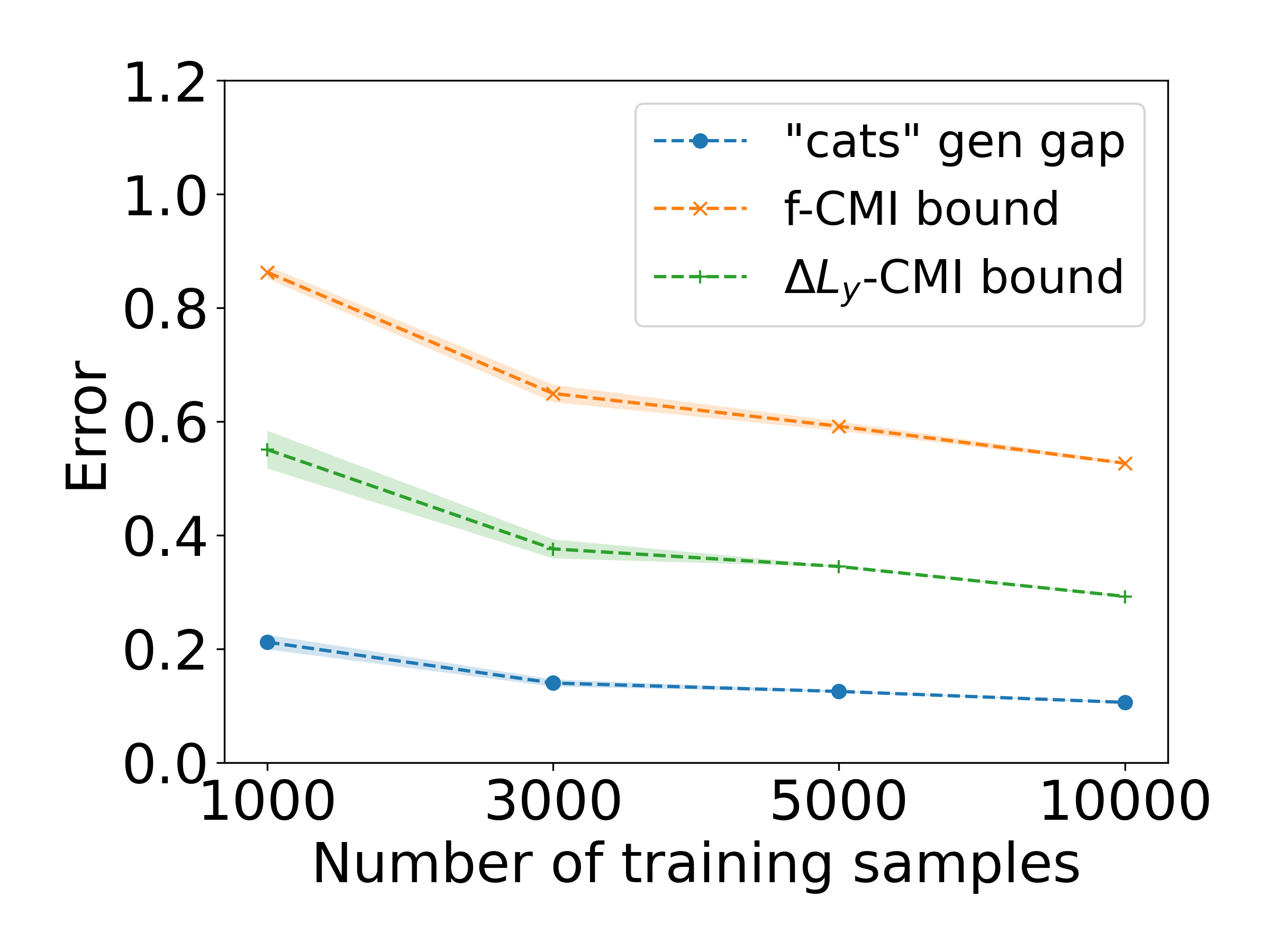}
\includegraphics[width=0.32\linewidth]{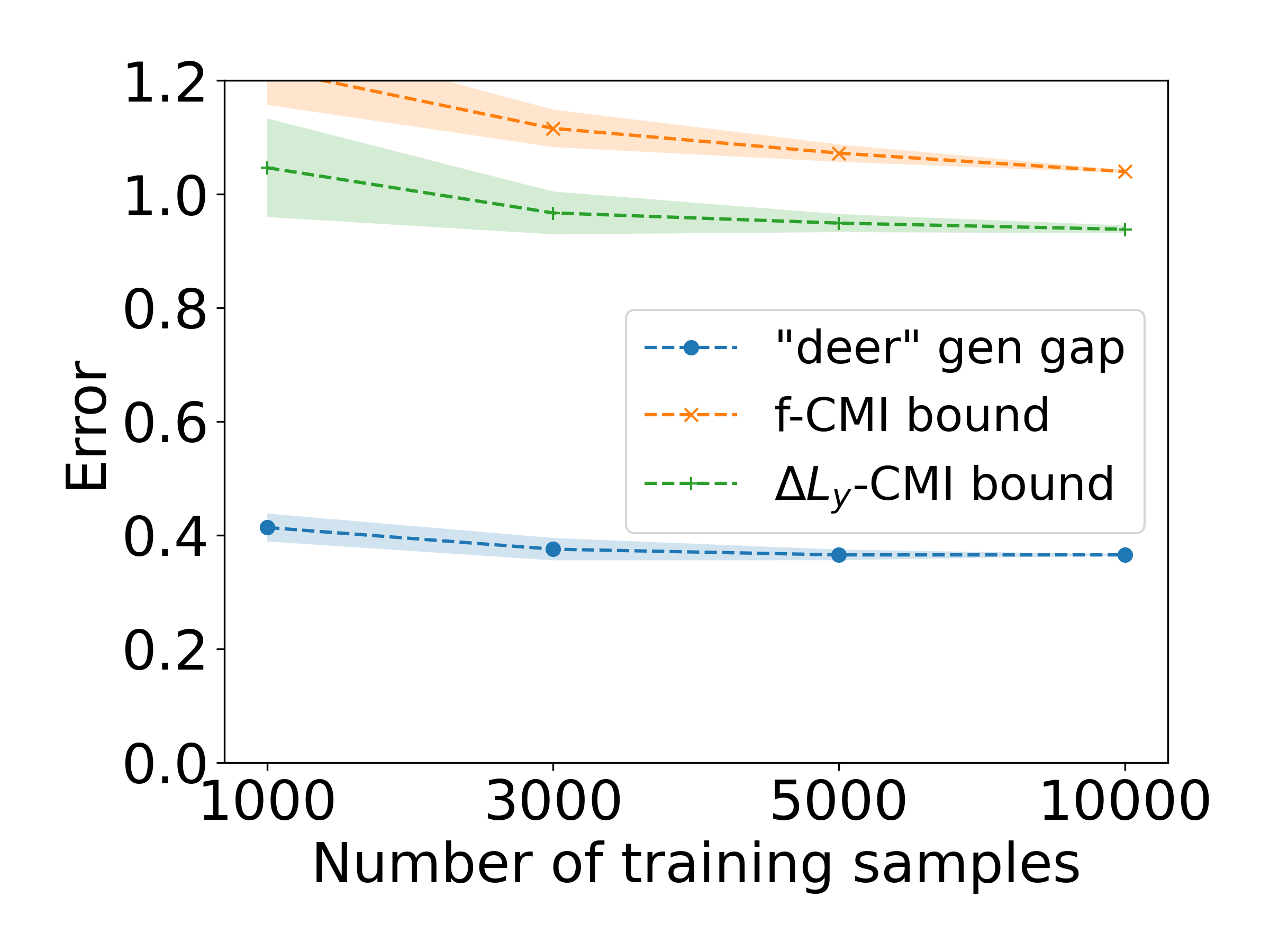}
\includegraphics[width=0.32\linewidth]{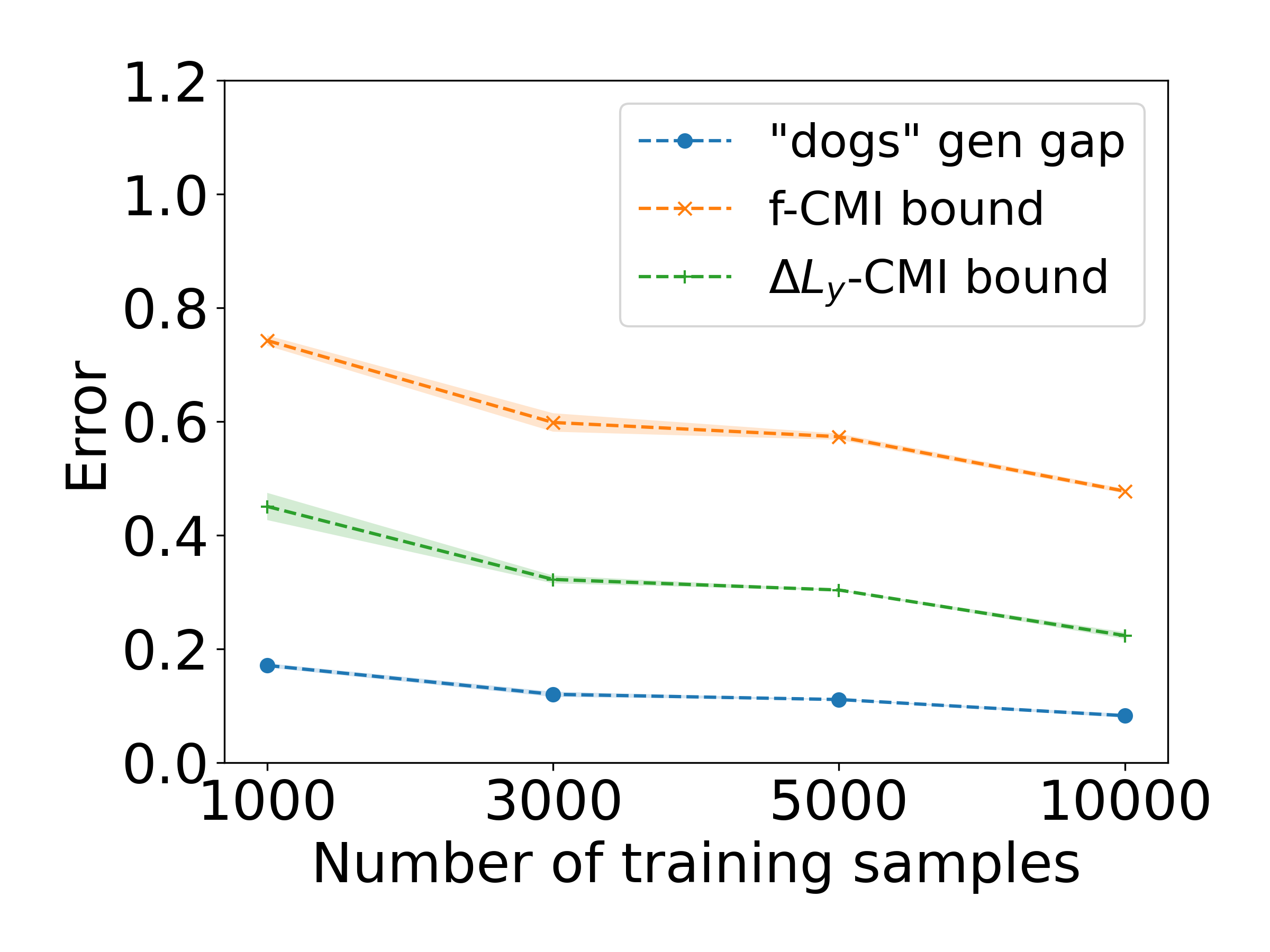}
\includegraphics[width=0.32\linewidth]{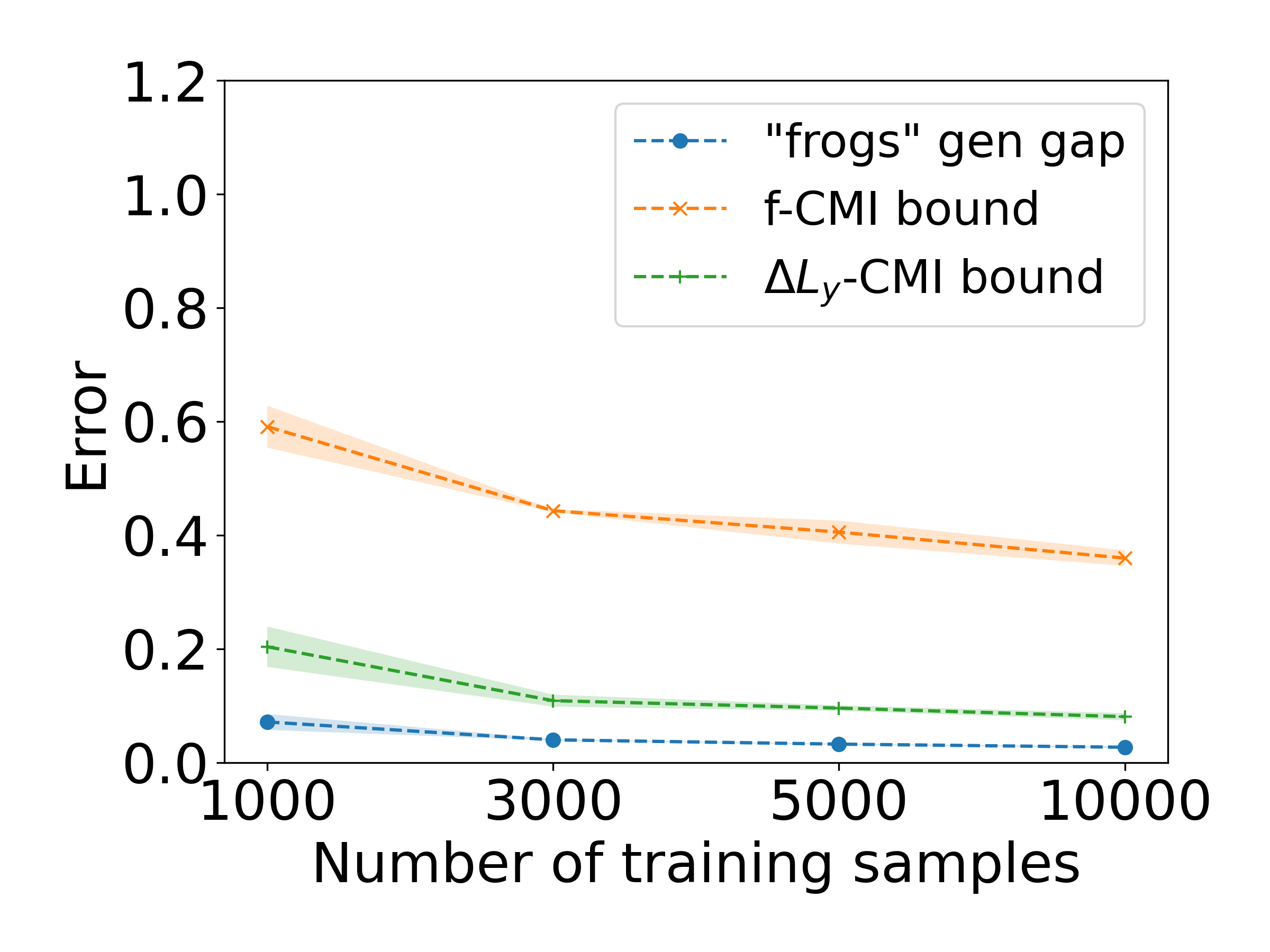}
\includegraphics[width=0.32\linewidth]{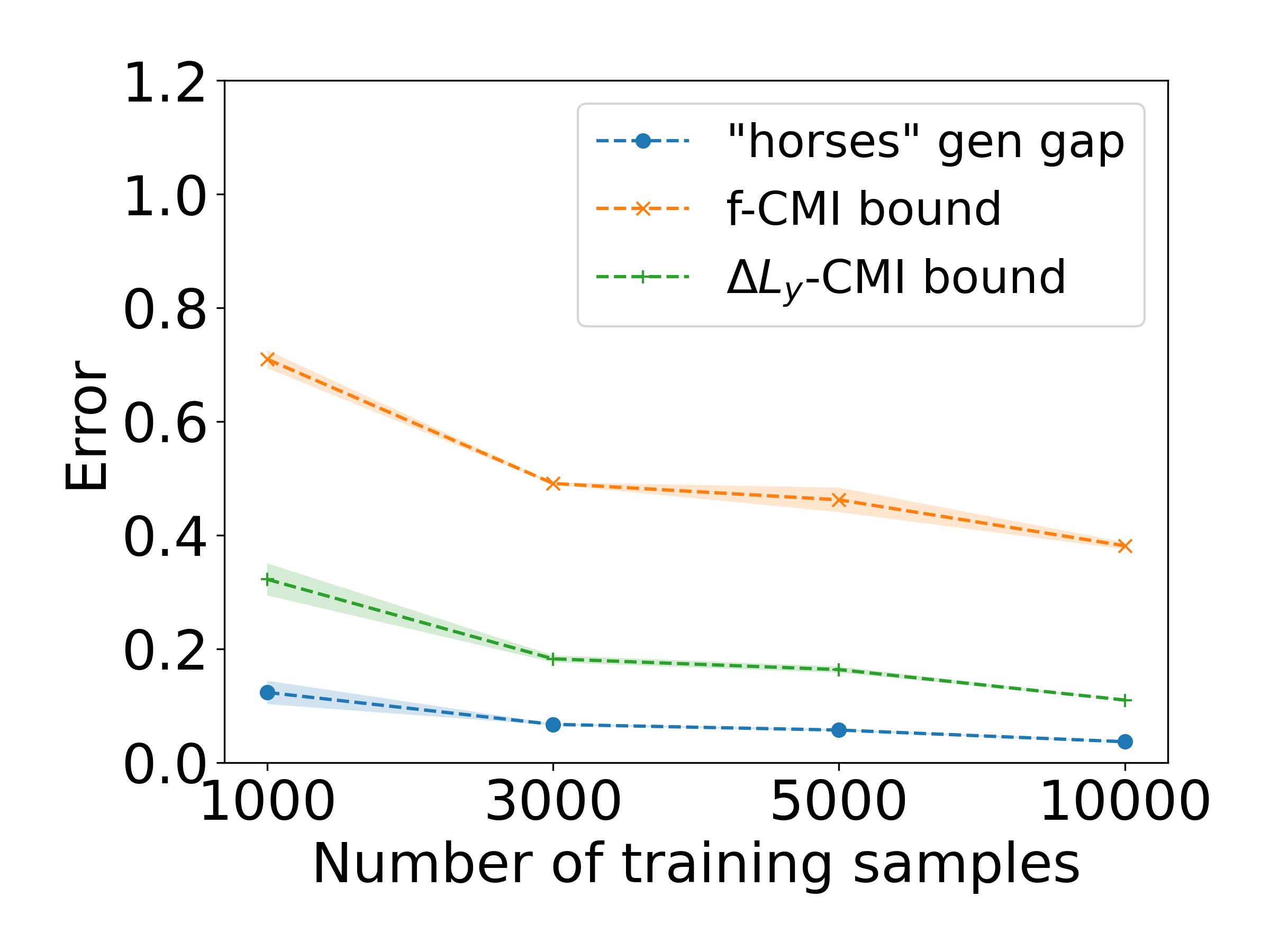}
\includegraphics[width=0.32\linewidth]{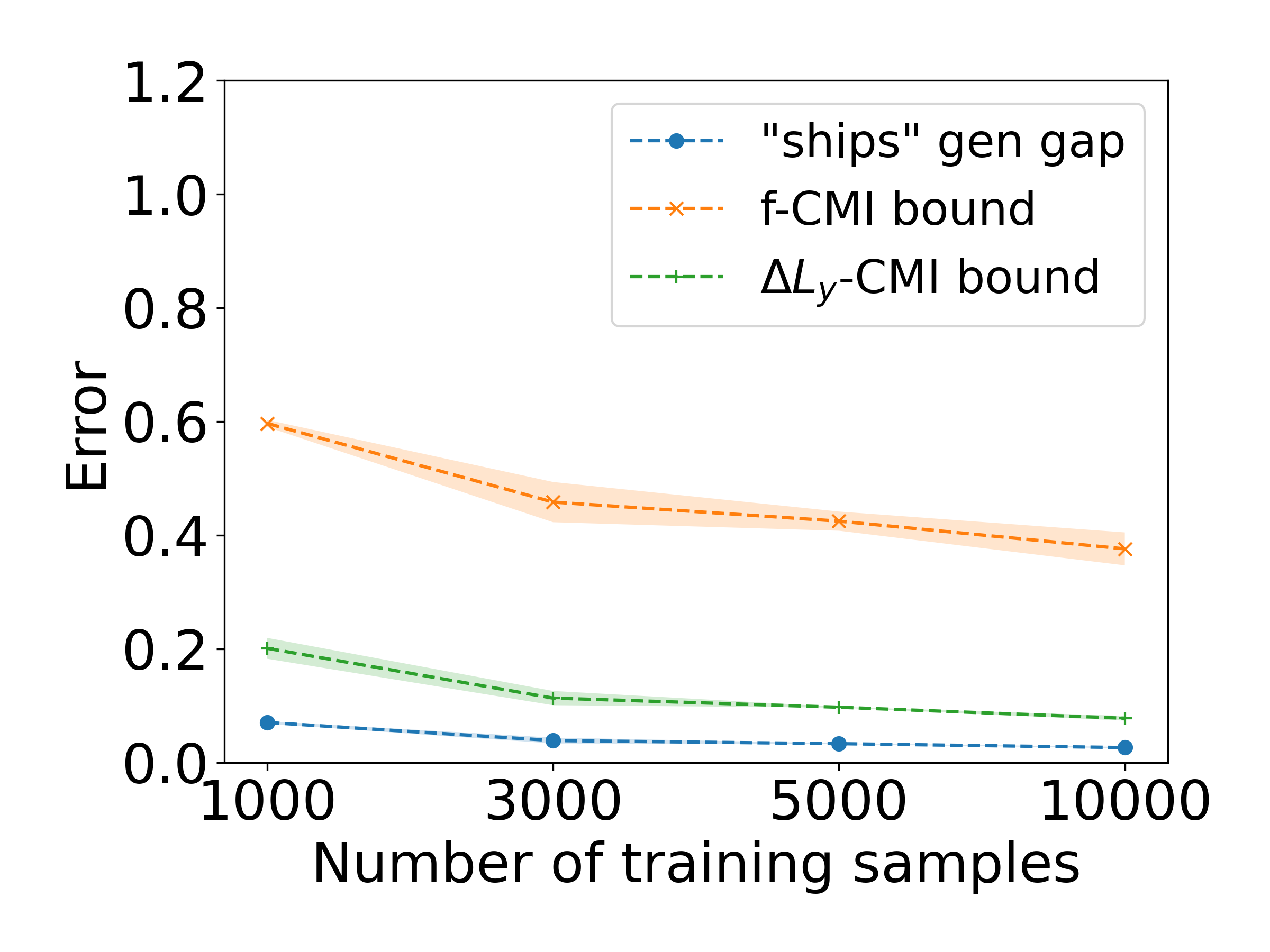}
\includegraphics[width=0.32\linewidth]{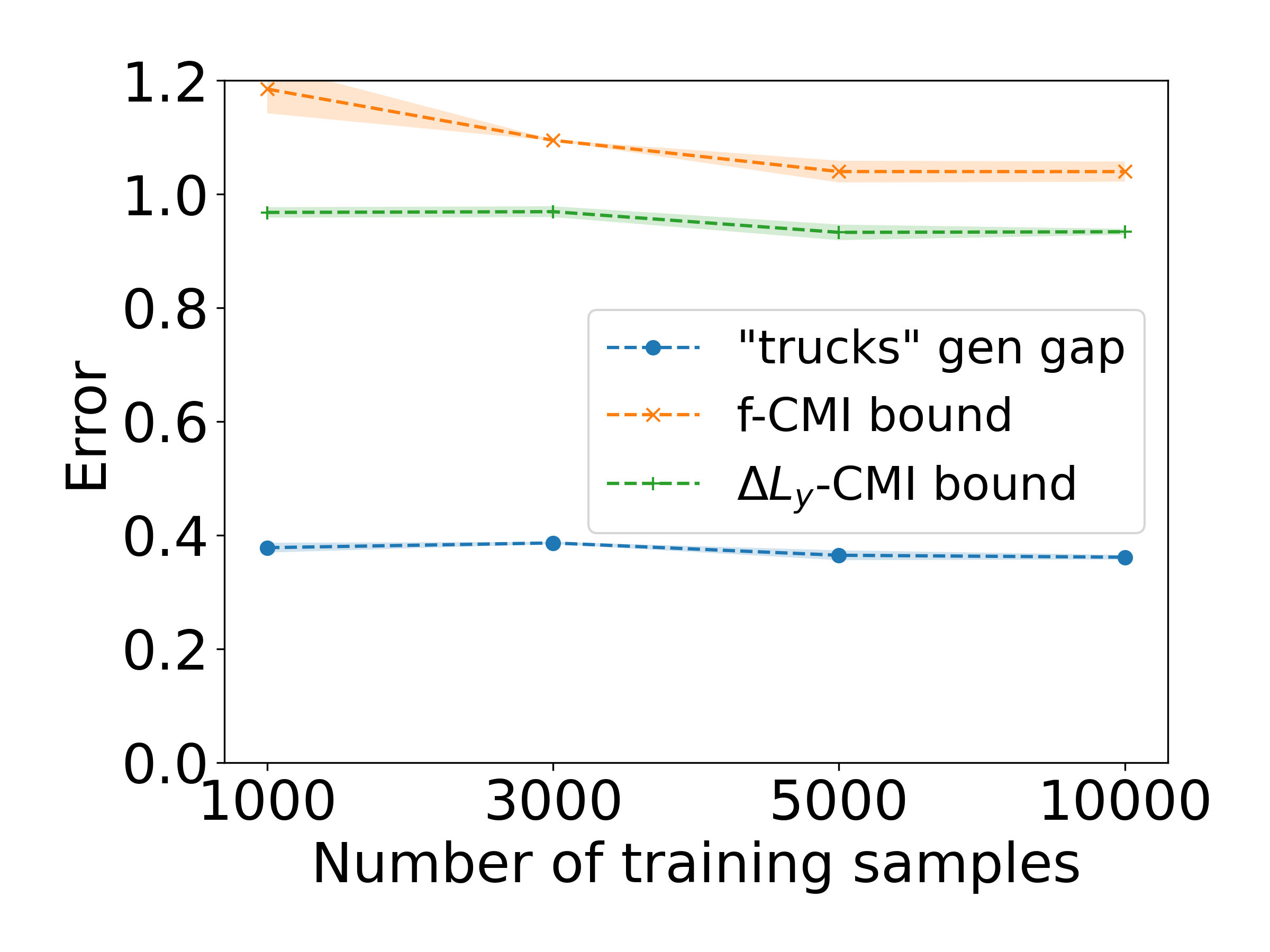}
\includegraphics[width=0.28\linewidth]{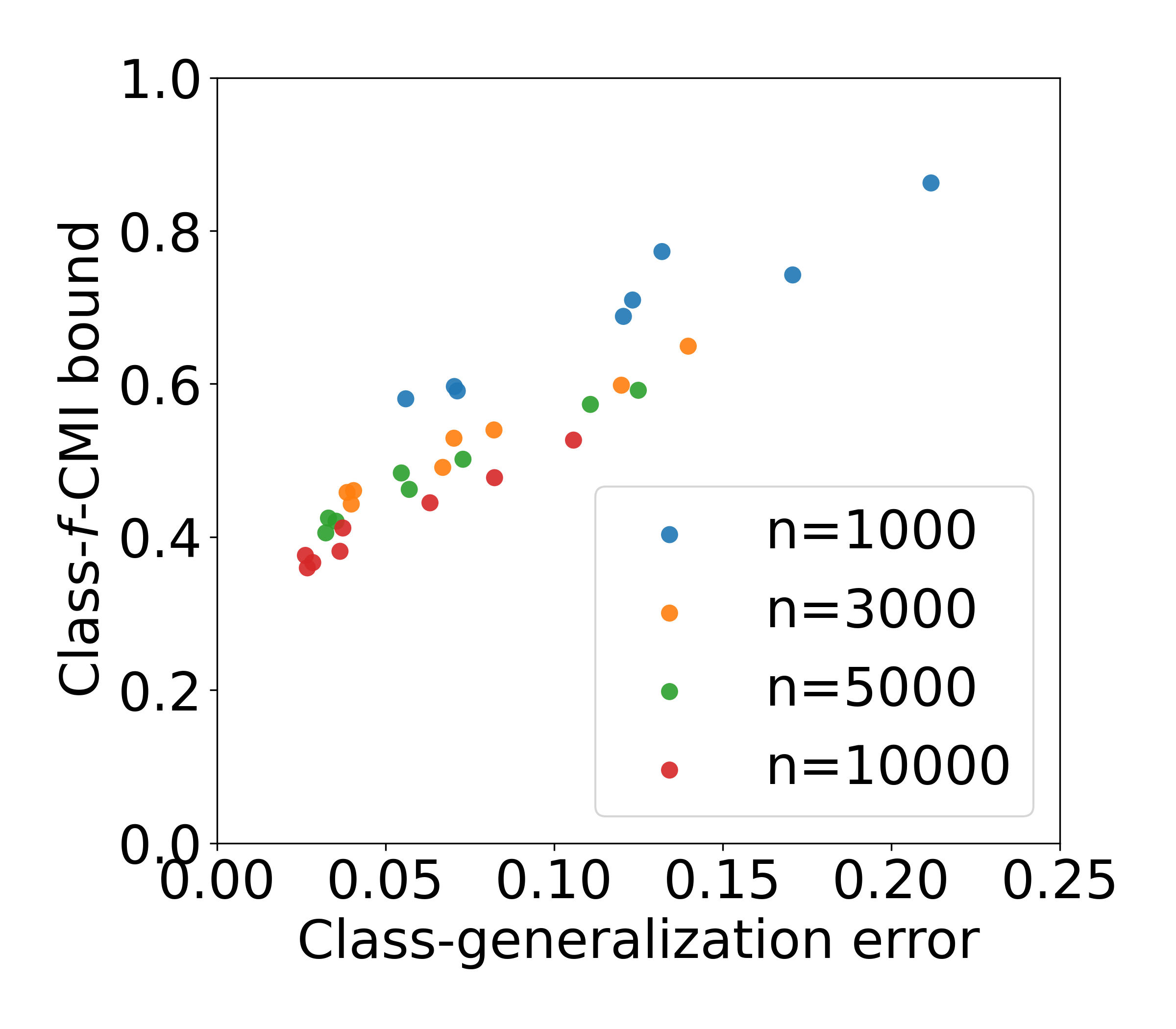}

\caption{Class-wise generalization on the 10 classes of noisy CIFAR10 (noise added to both train and validation)  and the scatter plot between class-generalization error and the class-$f$-CMI bound in Theorem~\ref{fCMI_class_bound}. }  
\end{figure*}
\newpage

\subsection{Corollary~\ref{MI_geny2gen} compared to the bound in \cite{bu2020tightening} }\label{coro_bu2020}

In the case of standard loss sub-gaussianity assumption, i.e., $\sigma_y = \sigma$ is independent of $y$, it is possible to show that the bound in~\ref{MI_geny2gen} is tighter than the bound in~\cite{bu2020tightening}. 
This is due to the fact that 
\begin{multline}
   \E_{P_Y}  \sqrt{2\sigma^2D(P_{\rmW,\rmX|\rmY} ||P_{\rmW}\otimes P_{\rmX|\rmY}) } =   \E_{P_Y}  \sqrt{2\sigma^2 \E_{P_{\rmW,\rmX|\rmY}} \frac{P_{\rmW,\rmX|\rmY} P_{\mY}}{P_{\rmW}\otimes P_{\rmX|\rmY} P_{\mY}} } \\ = \E_{P_Y}  \sqrt{2\sigma^2 \E_{P_{\rmW,\rmX|\rmY}} \frac{P_{\rmW,\rmX,\rmY}}{P_{\rmW}\otimes P_{\rmX,\rmY} }} \leq  \sqrt{2\sigma^2 I(\rmW ;\rmZ)}
\end{multline}
where the last inequality comes from Jensen's inequality. This shows that class-wise analysis can be used to derive tighter generalization bounds. 

\subsection{Extra results: Expected recall \& specificity generalization } \label{app_recall}
In the special case of binary classification with the 0-1 loss, the class-generalization errors studied within this paper correspond to generalization in terms of recall and specificity: \\ 
$expected \;  recall -  empirical \;  recall = gen_p$, where $p$ is the positive class.   \\ 
$expected \;  specificity-  empirical \;  specificity = gen_n$, where $n$ is the negative class.  

Standard Generalization bounds \citep{xu2017information,harutyunyan2021information} provide theoretical certificates for learning algorithms regarding classification error/accuracy. However, in several ML applications, e.g., an imbalanced binary dataset, accuracy/error are not considered good performance metrics. For example, consider the binary classification problem of detecting a rare cancer type. While analyzing the model's generalization error is important, in this case, we might be more interested in understanding generalization in terms of recall, as that is more critical in this case. Standard theoretical bounds~\citep{neyshabur2017exploring,wu2020information,wang2023tighter} do not provide any insights for such metrics. 

The developed tools in this paper can be used to close this gap and allow us to understand generalization for recall and specificity theoretically. 

\begin{figure*}[h]
\centering
\includegraphics[width=0.48\linewidth]{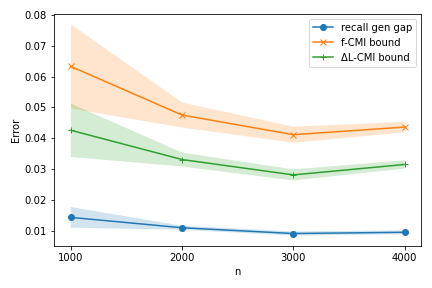}
\includegraphics[width=0.48\linewidth]{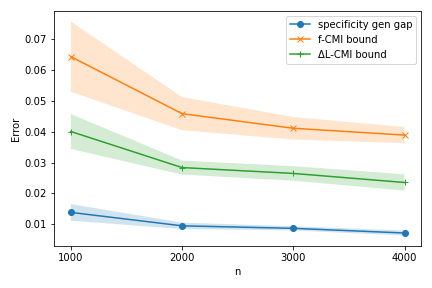}

\caption{recall generalization error (left) and specificity generalization error (right) for the binary classification 4vs9 from MNIST. The digit 4 is considered the positive class.}
\label{class-wise on MNIST4vs9 }
\end{figure*}

We conduct an experiment of binary MNIST (digit 4 vs. digit 9), similar to~\cite{harutyunyan2021information}. $m_1$ and $m_2$ discussed in Section~\ref{expi_setup} are selected to be $m_1=5$ and $m_2=30$.  Empirical results for this particular case are presented in Figure~\ref{class-wise on MNIST4vs9 }. As can be seen in the Figure, our bounds efficiently estimate the expected recall and specificity errors.

\section{Extra analysis for Section~\ref{otherapp_section}: other applications}

\subsection{Full details of Section~\ref{sec:standard_gen}: FROM CLASS-GENERALIZATION ERROR TO STANDARD GENERALIZATION ERROR} 

\textbf{Corollary~\ref{MI_geny2gen}} (restated)
    Assume that for every $y \in \gY$, the loss $\ell(\overline{\rmW},\overline{\rmX},y)$ is $\sigma_y$ sub-gaussian under $P_{\overline{\rmW}} \otimes P_{\overline{\rmX}|\overline{\rmY}=y }$, then     
\begin{equation}
  | \overline{\mathrm{gen}}(P_{\rmX,\rmY}, P_{\rmW|\rmS}) | \leq \frac{1}{n} \sum_{i=1}^n \E_Y \sqrt{2\sigma_{\rmY}^2D(P_{\rmW,\rmX_i|\rmY_i=y}||P_{\rmW}\otimes P_{\rmX_i|\rmY_i=y}) }.
\end{equation}
\begin{proof} 
The generalization error can be written as
    \begin{equation} \label{maineq22}
       \overline{\mathrm{gen}}(P_{\rmX,\rmY}, P_{\rmW|\rmS}) = \frac{1}{n} \sum_{i=1}^n \big( \E_{\rmW,\overline{\rmZ}}[\ell(\rmW,\overline{\rmZ})] - \E_{\rmW,\rmZ_i}[\ell(\rmW,\rmZ_i)] \big) .
    \end{equation}
 
As the loss $\ell$ is $\sigma_\rmY$ sub-gaussian, using Theorem~\ref{MI_class_bound}, we have  
\begin{equation}
     \E_{P_{\rmW,\rmX|\rmY=y}}[ \ell(\rmW,\rmX,\rmY)] -  \E_{P_{\rmW}\otimes P_{\rmX|\rmY=y}} [\ell(\overline{\rmW},\overline{\rmX},\overline{\rmY})]  \leq \sqrt{2\sigma_{y}^2D(P_{\rmW,\rmX|\rmY=y}||P_{\rmW}\otimes P_{\rmX|\rmY=y}) }.
\end{equation}  
Taking the expectation over $\rmY$ in both sides, we have 
\begin{equation} \label{labelgene}
     \E_{P_{\rmW,\rmX,\rmY}}[ \ell(\rmW,\rmX,\rmY)] -  \E_{P_{\rmW}\otimes P_{\rmX,\rmY}} [\ell(\overline{\rmW},\overline{\rmX},\overline{\rmY})]  \leq  \E_Y \sqrt{2\sigma_{\rmY}^2D(P_{\rmW,\rmX|\rmY=y}||P_{\rmW}\otimes P_{\rmX|\rmY=y}) }.
\end{equation}  
Applying \eqref{labelgene} on each term of \eqref{maineq22} for each $\rmZ_i$ completes the proof.

\end{proof}

With similar proofs, we can also extend the results in Theorems~\ref{CMI_class_bound}, ~\ref{fCMI_class_bound}, and~\ref{deltaLCMI_class_bound} into standard generalization bounds by taking the expectation over $y \sim P_{\rmY}$. In Corollaries~\ref{CMI_geny2gen}, ~\ref{fCMI_geny2gen}, ~\ref{LCMI_geny2gen}, and~\ref{deltaCMI_geny2gen}, we provide such an extension of  Theorems~\ref{CMI_class_bound}, ~\ref{fCMI_class_bound}, ~\ref{LCMI_class_bound}, and \ref{deltaLCMI_class_bound}, respectively. 

\textbf{Corollary~\ref{deltaCMI_geny2gen}}(restated)
    Assume that the loss $\ell(\hat{y},y) \in [0,1]$, then 
\begin{equation}
  | \overline{\mathrm{gen}}(P_{\rmX,\rmY}, P_{\rmW|\rmS}) | \leq \E_{\mY} \bigg[ \E_{\rmZ_{[2n]}}   
\Big[ \frac{1}{n^\rmY}  \sum_{i=1}^n   \sqrt{2 I_{\rmZ_{[2n]}} (\Delta_{\rmY} \rmL_i;\rmU_i )} \Big] \bigg].
\end{equation}

\begin{corollary} (extra result)\label{CMI_geny2gen}
  Assume that the loss $\ell(\hat{y},y) \in [0,1]$, then 
\begin{equation}
  | \overline{\mathrm{gen}}(P_{\rmX,\rmY}, P_{\rmW|\rmS}) | \leq \E_{\mY} \bigg[ \E_{\rmZ_{[2n]}}   
\Big[ \frac{1}{n^\rmY}  \sum_{i=1}^n   \sqrt{2 \max(\mathds{1}_{\rmY_i^{-}=\rmY}, \mathds{1}_{\rmY_i^{+}=\rmY}) I_{\rmZ_{[2n]}} (\rmW; \rmU_i )} \Big] \bigg].
\end{equation}
\end{corollary}

\begin{corollary} (extra result) \label{fCMI_geny2gen}
   Assume that the loss $\ell(\hat{y},y) \in [0,1]$, then 
\begin{equation}
  | \overline{\mathrm{gen}}(P_{\rmX,\rmY}, P_{\rmW|\rmS}) | \leq \E_{\mY} \bigg[ \E_{\rmZ_{[2n]}}   
\Big[ \frac{1}{n^\rmY}  \sum_{i=1}^n   \sqrt{2 \max(\mathds{1}_{\rmY_i^{-}=\rmY}, \mathds{1}_{\rmY_i^{+}=\rmY}) I_{\rmZ_{[2n]}} (f_{\rmW}(\rmX^\pm_i);\rmU_i )} \Big] \bigg].
\end{equation}
\end{corollary}

\begin{corollary} (extra result) \label{LCMI_geny2gen}
    Assume that the loss $\ell(\hat{y},y) \in [0,1]$, then 
\begin{equation}
  | \overline{\mathrm{gen}}(P_{\rmX,\rmY}, P_{\rmW|\rmS}) | \leq \E_{\mY} \Bigg[ \E_{\rmZ_{[2n]}}   
\Big[ \frac{1}{n^\rmY}  \sum_{i=1}^n   \sqrt{2 \max(\mathds{1}_{\rmY_i^{-}=\rmY}, \mathds{1}_{\rmY_i^{+}=\rmY}) I_{\rmZ_{[2n]}} (\rmL^{\pm}_i;\rmU_i )} \Big] \Bigg].
\end{equation}
\end{corollary}

\subsection{Full details of the sub-task problem} \label{subtask_sup}
Consider a supervised learning problem where
the machine learning model $f_\rmW(\cdot)$, parameterized with $w \in \gW$, is obtained with a training dataset $S$ consisting of $n$ i.i.d samples $z_i = (x_i, \!y_i) \in\! \mathcal{X}\! \times \!\mathcal{Y} \triangleq \mathcal{Z}$ generated from distribution $P_{\rmX \rmY}$. The quality of the model with parameter $w$ is evaluated with a loss function $\ell: \gW \times \gZ \rightarrow \sR^+$.
 
For any $w \in \gW$, the population risk is defined as follows
\begin{equation}
    L_{P}(w) = \E_{P_{\rmX,\rmY}} [\ell(w,\rmX,\rmY)].
\end{equation}
and the empirical risk is:
\begin{equation}
    L_{E_P}(w,S) =\frac{1}{n} \sum_{i=1}^n \ell(w,x_i,y_i).
\end{equation}
Here, we are interested in the subtask problem, which is a special case of distribution shift, i.e., the test performance of the model $w$ is evaluated using a specific subset of classes $\gA \subset \gY$ of the source distribution $P_{\rmX \rmY}$. Thus, the target distribution  $Q_{\rmX \rmY}$ is defined as
$Q_{\rmX \rmY}(x,y) = \frac{P_{\rmX \rmY}(x,y) \mathds{1}_{\{y \in \gA\}}}{P_{\rmY}(y\in \gA)}  $.
The population risk on the target domain $Q$ of the subtask problem is
\begin{equation} \label{poprisktarget}
    L_{Q}(w) = \E_{Q_{\rmX,\rmY}} [\ell(w,\rmX,\rmY)].
\end{equation}

A learning algorithm can be modeled as a randomized mapping from the training set $S$ onto a model parameter $w \in \mathcal{W}$  according to the conditional distribution $P_{\rmW|S}$.
The expected generalization error on the subtask problem is the difference between the population risk of $Q$ and the empirical risk evaluated using all samples from $S$:
\begin{equation}\label{eqgen1}
    \overline{\mathrm{gen}}_{Q,E_P} = \E_{P_{\rmW,\rmS}} [  L_{Q}(\rmW) -  L_{E_P}(\rmW, \rmS) ],
\end{equation} 
where the expectation is taken over the joint distribution $P_{\rmW,\rmS} =  P_{\rmW|S}\otimes P_{\rmZ}^n$.

The generalization error defined above can be decomposed as follows:
\begin{equation} \label{eqgen}
    \overline{\mathrm{gen}}_{Q,E_P} = \E_{P_{\rmW}} [  L_{Q}(\rmW) - L_{P}(\rmW)  ]  \\ 
    + \E_{P_{\rmW,\rmS}} [  L_{P}(\rmW)  -  L_{E_P}(\rmW,\rmS)  ].  
\end{equation}
The first term quantifies the gap of the population risks in two different domains, and the second term is the source domain generalization error. Assuming that loss is $\sigma$-subgaussian under $P_{\rmZ}$, it is shown in~\cite{wang2022information} that
the first term can be bounded using the KL divergence between $P$ and $Q$:
\begin{equation}
    \E_{P_{\rmW}} [  L_{Q}(w) - L_{P}(w)  ]   \leq  \sqrt{2 \sigma^2 D(Q\|P)}.
\end{equation} 
The second term can be bounded using the standard mutual information approach in~\cite{xu2017information} as
\begin{equation}
    \E_{P_{\rmW,\rmS}} [  L_{P}(\rmW)  -  L_{E_P}(\rmW, \rmS)  ]   \leq  \sqrt{2 \sigma^2 I(\rmW;\rmS)}.
\end{equation} 
Thus, the generalization error of the subtask problem can be bounded as follows: 
\begin{equation} \overline{\mathrm{gen}}_{Q,E_P} \leq  \sqrt{2 \sigma^2 D(Q\|P)} + \sqrt{2 \sigma^2 I(\rmW;\rmS)}. 
\end{equation}

Obtaining tighter generalization error bounds for the subtask problem is straightforward using our class-wise generalization bounds.  In fact, the generalization error bound of the subtask can be obtained by taking the expectation of $\rmY \sim Q_{{\rmY}}$. 

Using Jensen's inequality, we have $ |   \overline{\mathrm{gen}}_{ Q,E_{Q}} | = |\E_{\rmY \sim Q_\rmY} \big[ \overline{\mathrm{gen}}_\rmY\big] | \leq \E_{\rmY \sim Q_\rmY} \big[ | \overline{\mathrm{gen}}_\rmY| \big] $. Thus, we can use the results from Section~\ref{section_classgenerror} to obtain tighter bounds.

\textbf{Theorem~\ref{subtaskcmi_bound}} (subtask-CMI) (restated)
Assume that the loss $\ell(w,x,y) \in [0,1]$ is bounded, then the subtask generalization error  defined in~\ref{main_gen_yq} can be bounded as
\begin{equation}
|\overline{\mathrm{gen}}_{ Q,E_{Q}}|  \leq 
     \E_{\rmY \sim Q_\rmY} \bigg[  \E_{\rmZ_{[2n]}}  \Big[ \frac{1}{n^\rmY}  \sum_{i=1}^n   \sqrt{2 \max(\mathds{1}_{\rmY_i^{-}=\rmY},\mathds{1}_{\rmY_i^{+}=\rmY})  I_{\rmZ_{[2n]}} (\rmW; \rmU_i )} \Big] \bigg]. \nonumber
\end{equation}  
Similarly, we can extend the result of class-$\Delta L_y$-CMI, with similar proof, as follows:

\textbf{Theorem~\ref{subtaskdeltacmi_bound}} (subtask-$\Delta L_y$-CMI) (restated)
Assume that the loss $\ell(w,x,y) \in [0,1]$ is bounded, Then the subtask generalization error  defined in~\ref{main_gen_yq} can be bounded as
\begin{equation}
   |   \overline{\mathrm{gen}}_{Q,E_{Q}}|  \leq 
\E_{\rmY \sim Q_\rmY} \bigg[  \E_{\rmZ_{[2n]}}   
\Big[ \frac{1}{n^\rmY}  \sum_{i=1}^n    \sqrt{2 I_{\rmZ_{[2n]}} (\Delta_{\rmY} \rmL_i;\rmU_i )}\Big] \bigg]. \nonumber
\end{equation}  

Similarly, we can also extend the result of Theorem~\ref{LCMI_class_bound} to the subtask as follows:
\begin{theorem}
     (subtask-e-CMI) (extra result)
Assume that the loss $\ell(w,x,y) \in [0,1]$ is bounded, then the subtask generalization error  defined in~\ref{main_gen_yq} can be bounded as
\begin{equation}
|\overline{\mathrm{gen}}_{ Q,E_{Q}}|  \leq 
     \E_{\rmY \sim Q_\rmY} \bigg[  \E_{\rmZ_{[2n]}}  \Big[ \frac{1}{n^\rmY}  \sum_{i=1}^n   \sqrt{2 \max(\mathds{1}_{\rmY_i^{-}=\rmY},\mathds{1}_{\rmY_i^{+}=\rmY})  I_{\rmZ_{[2n]}} (f_{\rmW}(\rmX^\pm_i); \rmU_i )} \Big] \bigg]. \nonumber
\end{equation}
\end{theorem}

\subsection{Extra details of Section~\ref{section_fairness}: Generalization certificates with sensitive attributes } 

\textbf{Theorem~\ref{MI_attribute_bound}} (restated)      Given $t \in \gT$, assume that the loss $\ell(\rmW,\rmZ)$ is $\sigma$ sub-gaussian under $P_{\overline{\rmW}} \otimes P_{\overline{\rmZ} }$, then  the  attribute-generalization error of the sub-population $\rmT=t$, as defined in~\ref{attri_gen}, can be bounded as follows:
\begin{equation}
   |  \overline{\mathrm{gen}_t}(P_{\rmX,\rmY}, P_{\rmW|\rmS}) | \leq \sqrt{2\sigma^2D(P_{\rmW|\rmZ} \otimes P_{\rmZ|\rmT=t}||P_{\rmW}\otimes P_{\rmZ|\overline{\rmT}=t}) }.
\end{equation} 
\begin{proof}
We have 
\begin{equation}
       \overline{\mathrm{gen}_t}(P_{\rmX,\rmY}, P_{\rmW|\rmS}) = \E_{P_{\overline{\rmW}}\otimes P_{\overline{\rmZ}|\rmT=t}} [\ell(\overline{\rmW},\overline{\rmZ})] - \E_{P_{\rmW|\rmZ} \otimes P_{\rmZ|\rmT=t}}[ \ell(\rmW,\rmZ)]. 
\end{equation}
Using the Donsker–Varadhan variational representation of the relative entropy,  we have
\begin{multline} \label{main_eqatri}
       D(P_{\rmW|\rmZ} \otimes P_{\rmZ|\rmT=t}||P_{\rmW}\otimes P_{\rmZ|\rmT=t}) \geq \E_{P_{\rmW|\rmZ} \otimes P_{\rmZ|\rmT=t}}[\lambda \ell(\rmW,\rmZ)] \\ - \log \E_{P_{\overline{\rmW}}\otimes P_{\overline{\rmZ}|\overline{\rmT}=t}}[e^{\lambda \ell(\overline{\rmW},\overline{\rmZ})} ], \forall \lambda \in \sR.
   \end{multline} 
On the other hand, we have:
\begin{align*}   
    &\log  \E_{P_{\overline{\rmW}}\otimes P_{\overline{\rmZ}|\overline{\rmT}=t}} \Big[ e^{\lambda \ell(\overline{\rmW},\overline{\rmZ}) - \lambda\E [\ell(\overline{\rmW},\overline{\rmZ})]  }\Big]\\
    &=   \log  \E_{P_{\overline{\rmW}}\otimes P_{\overline{\rmZ}|\overline{\rmT}=t}} \Big[ e^{\lambda \ell(\overline{\rmW},\overline{\rmZ}) }  e^{- \lambda  \E [\ell(\overline{\rmW},\overline{\rmZ})]  \big)}\Big]  \\
    &= \log \E_{P_{\overline{\rmW}}\otimes P_{\overline{\rmZ}|\overline{\rmT}=t}}[e^{\lambda \ell(\overline{\rmW},\overline{\rmZ})} ] - \lambda \E_{P_{\overline{\rmW}}\otimes P_{\overline{\rmZ}|\overline{\rmT}=t}} [\ell(\overline{\rmW},\overline{\rmZ})].
\end{align*}
Using the sub-gaussian assumption, we have
\begin{equation}
    \log \E_{P_{\overline{\rmW}}\otimes P_{\overline{\rmZ}|\overline{\rmT}=t}}[e^{\lambda \ell(\overline{\rmW},\overline{\rmZ})} ] \leq \lambda \E_{P_{\overline{\rmW}}\otimes P_{\overline{\rmZ}|\overline{\rmT}=t}} (\ell(\overline{\rmW},\overline{\rmZ})) + \frac{\lambda^2 \sigma^2}{2}.
\end{equation}
By replacing in \eqref{main_eqatri}, we have 
\begin{multline}
     D(P_{\rmW|\rmZ} \otimes P_{\rmZ|\rmT=t}||P_{\rmW}\otimes P_{\rmZ|\rmT=t}) \geq \lambda \big(\E_{P_{\rmW|\rmZ} \otimes P_{\rmZ|\rmT=t}}[ \ell(\rmW,\rmZ)] -  \\ \E_{P_{\overline{\rmW}}\otimes P_{\overline{\rmZ}|\overline{\rmT}=t}} [\ell(\overline{\rmW},\overline{\rmZ})] \big)  -  \frac{\lambda^2 \sigma}{2}.
\end{multline}

Thus, we have: 
\begin{multline} \label{eq11attri}
    D(P_{\rmW|\rmZ} \otimes P_{\rmZ|\rmT=t}||P_{\rmW}\otimes P_{\rmZ|\rmT=t}) -  \lambda (\E_{P_{\rmW|\rmZ} \otimes P_{\rmZ|\rmT=t}}[ \ell(\rmW,\rmZ)] -  \E_{P_{\overline{\rmW}}\otimes P_{\overline{\rmZ}|\overline{\rmT}=t}} [\ell(\overline{\rmW},\overline{\rmZ})] )   \\ + \lambda^2 \sigma^2 \geq 0 , \forall \lambda  \in \sR.
\end{multline}
\eqref{eq11attri} is a non-negative parabola with respect to $\lambda$. Thus, its discriminant must be non-positive. This implies 
\begin{equation}
   |  \E_{P_{\rmW|\rmZ} \otimes P_{\rmZ|\rmT=t}}[ \ell(\rmW,\rmZ)] -  \E_{P_{\overline{\rmW}}\otimes P_{\overline{\rmZ}|\overline{\rmT}=t}} [\ell(\overline{\rmW},\overline{\rmZ})] | \leq \sqrt{2\sigma^2D(P_{\rmW|\rmZ} \otimes P_{\rmZ|\rmT=t}||P_{\rmW}\otimes P_{\rmZ|\rmT=t}) }.
\end{equation}
   This completes the proof.
\end{proof}

\end{document}